%% file: main_arxiv.tex
\documentclass[10pt,a4paper,onecolumn,teaser,showabstract]{styles_naverlabs/naverlabseurope}

\usepackage{tikz,pgfplots}
\usepackage{bm}
\usepackage{multirow}
\usepackage{float}
\usepackage[htt]{hyphenat}
\usepackage{makecell}
\usepackage{subcaption}
\setitemize{noitemsep,topsep=0pt,parsep=0pt,partopsep=0pt}
\usepackage{tcolorbox}
\tcbuselibrary{breakable}
\usepackage{ulem}
\usepackage{listings}
\input{iclr2026/math_commands.tex}

\newcommand{\paragraphshortstyle}[1]{\paragraph{#1}}
\input{macros}
\SetPaperVersion{arxiv}
\ShowPaperTodos %

\input{styles/upshot_box}

\defcitealias{torchtune}{TorchTune}

\newcommand{\citepAliasYear}[1]{(\citetalias{#1}, \citeyear{#1})}

\newcommand{\chip}[2]{\begingroup\setlength{\fboxsep}{1pt}\colorbox{#1}{#2}\endgroup}

\newcommand{\dchip}[2]{%
  \begingroup
  \pgfmathsetmacro{\d}{max(-80,min(80,#1))}%
  \pgfmathsetmacro{\a}{abs(\d)}%
  \pgfmathsetmacro{\p}{max(5,(\a/80)*70)}%
  \ifdim \d pt > 0pt
    \chip{green!\p}{#2}%
  \else\ifdim \d pt < 0pt
    \chip{red!\p}{#2}%
  \else
    \chip{white}{#2}%
  \fi\fi
  \endgroup
}

\RenewDocumentCommand{\thomas}{O{} +m}{\PaperTodo[#1]{Thomas}{cyan!20}{#2}}

\newcommand{\oursraw}{\textsc{ExpRAG}}
\newcommand{\ours}{\oursraw~}

\title{Retrieval-Augmented LLM Agents:\\Learning to Learn from Experience}
\titlerunning{\textit{Retrieval-Augmented LLM Agents: Learning to Learn from Experience}}
\date{March, 2026}

\correspondingauthor{[thomas.palmeira, stephane.clinchant]@naverlabs.com}
\authors{Thomas Palmeira Ferraz \quad Romain Deffayet \quad  Vassilina Nikoulina \quad Herv\'{e} D\'{e}jean \quad St\'{e}phane Clinchant}

\affiliations{NAVER LABS Europe, France}

\teaserfig{
    \vspace{5pt}
    \makebox[\linewidth][c]{%
        \includegraphics[width=1.04\linewidth]{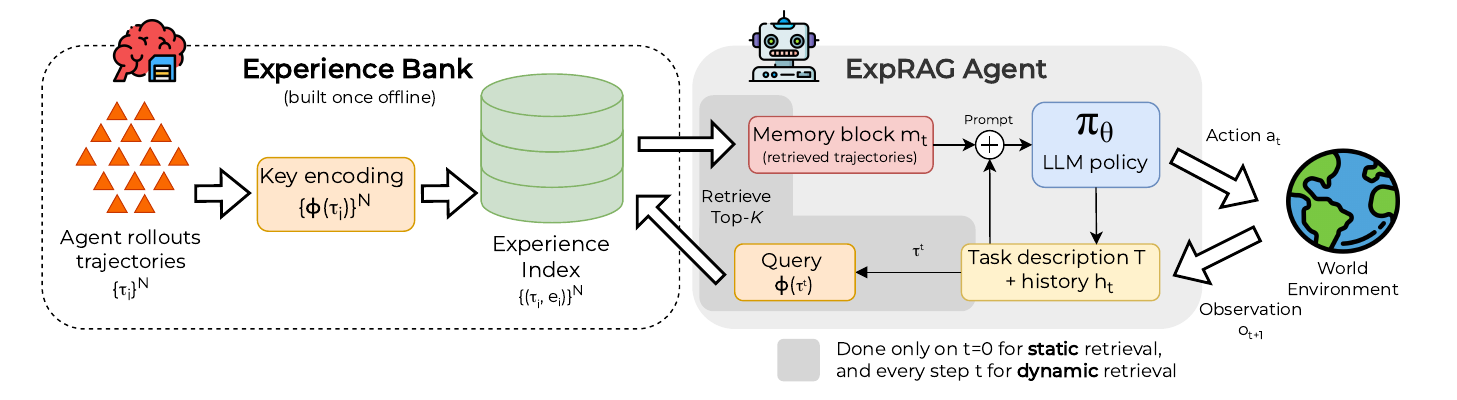}
    }
    \vspace{2pt}
}

\teasercaption{\textbf{\ours Agent overview.} \ours augments an LLM agent with retrieval over past experience. Offline, an experience bank is constructed once by collecting agent rollouts and encoding each trajectory \(\tau_i\) into a key embedding \(\phi(\tau_i)\), forming an index of trajectories paired with their representations. During inference, the current task description and interaction history \(h_t\) are encoded into a query, which is used to retrieve the top-\(K\) most relevant trajectories from the bank. The retrieved trajectories are then assembled into a memory block \(m_t\) and injected into the system prompt of the LLM policy \(\pi_\theta\), which outputs the next action \(a_t\). The environment returns a new observation \(o_{t+1}\), which is appended to the interaction history. Retrieval may be performed only once at \(t=0\) in the \textit{static} setting, or refreshed throughout the episode in the \textit{dynamic} setting. The loop is repeated, enabling continual experience-grounded decision making and improved generalization to unseen tasks.}

\input{Sections/0_abstract}

\begin{document}

\maketitle

\section{Introduction}
\input{tables/intro_table_files/intro_table}

\input{Sections/1_intro}

\input{Sections/2_related_work}

\input{Sections/3_method}

\input{Sections/4_experiments}

\input{Sections/5_zero_shot}
\input{Sections/6_finetuning}
\input{Sections/6.4_robustness_failures.tex}

\input{Sections/7_conclusion}
\subsection{Limitations and Future Work} 
\input{Sections/8_limitations}

\input{Sections/9_acknowledgements}

\bibliographystyle{styles_naverlabs/plainnat}
\bibliography{references}

\clearpage
\appendix
\tableofcontents
\clearpage

\input{appendix/additional_related_work}

\input{appendix/implementation_details}

\input{appendix/benchmark_details}
\clearpage
\input{appendix/zero_shot_continuation}

\input{appendix/longer_training}

\input{appendix/statistical_reliability}
\input{appendix/robustness}
\clearpage
\input{appendix/prompts}
\input{appendix/efficiency}

\end{document}

%% file: iclr2026/math_commands.tex
\usepackage{amsmath,amsfonts,bm}

\def\eqref#1{equation~\ref{#1}}

\def\1{\bm{1}}

\newcommand{\test}{\mathcal{D_{\mathrm{test}}}}

\DeclareMathAlphabet{\mathsfit}{\encodingdefault}{\sfdefault}{m}{sl}
\SetMathAlphabet{\mathsfit}{bold}{\encodingdefault}{\sfdefault}{bx}{n}

%% file: styles/upshot_box.tex
\tcbuselibrary{skins}

\definecolor{UpshotBlue}{HTML}{245B8F}
\definecolor{UpshotBlueFill}{HTML}{F3F7FB}
\definecolor{UpshotTeal}{HTML}{0B6E69}
\definecolor{UpshotTealFill}{HTML}{F1F8F7}
\definecolor{UpshotOchre}{HTML}{8A5A00}
\definecolor{UpshotOchreFill}{HTML}{FFF8E7}
\definecolor{UpshotGraphite}{HTML}{424B54}
\definecolor{UpshotGraphiteFill}{HTML}{F5F6F7}
\definecolor{UpshotInk}{HTML}{111111}

\tcbset{
  upshot/base/.style={
    enhanced,
    breakable,
    width=\linewidth,
    boxsep=0pt,
    before skip=5pt,
    after skip=5pt,
    left=5pt,
    right=5pt,
    coltext=UpshotInk,
    fontupper=\normalfont
  },
  upshot/palette/blue/.style={
    colback=UpshotBlueFill,
    colframe=UpshotBlue,
    colbacktitle=UpshotBlue
  },
  upshot/palette/teal/.style={
    colback=UpshotTealFill,
    colframe=UpshotTeal,
    colbacktitle=UpshotTeal
  },
  upshot/palette/ochre/.style={
    colback=UpshotOchreFill,
    colframe=UpshotOchre,
    colbacktitle=UpshotOchre
  },
  upshot/palette/graphite/.style={
    colback=UpshotGraphiteFill,
    colframe=UpshotGraphite,
    colbacktitle=UpshotGraphite
  },
  upshot/geometry/badge/.style={
    boxrule=0.7pt,
    leftrule=0.7pt,
    rightrule=0.7pt,
    toprule=0.7pt,
    bottomrule=0.7pt,
    arc=2pt,
    top=7.5pt,
    bottom=4.5pt,
    before upper={},
    title={Upshot},
    coltitle=white,
    fonttitle=\small\bfseries,
    attach boxed title to top left={xshift=4pt,yshift=-3pt},
    boxed title style={
      boxrule=0pt,
      arc=1.5pt,
      left=4pt,
      right=4pt,
      top=1.2pt,
      bottom=1.2pt
    }
  },
  upshot/geometry/rail/.style={
    boxrule=0pt,
    leftrule=1.6pt,
    rightrule=0pt,
    toprule=0pt,
    bottomrule=0pt,
    arc=0pt,
    left=6pt,
    right=3pt,
    top=3pt,
    bottom=3pt,
    before upper={\textbf{\textcolor{tcbcolframe}{Upshot.}}\enspace},
    title={}
  },
  upshot/geometry/outline/.style={
    boxrule=0.6pt,
    leftrule=0.6pt,
    rightrule=0.6pt,
    toprule=0.6pt,
    bottomrule=0.6pt,
    arc=0.5pt,
    top=6pt,
    bottom=4pt,
    before upper={},
    title={Upshot},
    colbacktitle=white,
    coltitle=UpshotInk,
    fonttitle=\footnotesize\bfseries,
    attach boxed title to top left={xshift=4pt,yshift=-3pt},
    boxed title style={
      boxrule=0pt,
      colback=white,
      left=2pt,
      right=2pt,
      top=0pt,
      bottom=0pt
    }
  },
  upshot/default palette/.style={upshot/palette/blue}
}

\ifaclversion
  \tcbset{upshot/default geometry/.style={upshot/geometry/rail}}
\else
  \tcbset{upshot/default geometry/.style={upshot/geometry/badge}}
\fi

\newtcolorbox{upshotbox}[1][]{
  upshot/base,
  upshot/default palette,
  upshot/default geometry,
  #1
}

\NewDocumentCommand{\upshot}{O{} +m}{%
  \begin{upshotbox}[#1]
    #2%
  \end{upshotbox}%
}

%% file: Sections/0_abstract.tex
\begin{abstract}
While large language models (LLMs) have advanced the development of general-purpose agents, robust generalization to unseen tasks remains challenging. Two common approaches are supervised fine-tuning and training-free memory-augmented generation using retrieved experience; yet both have limitations: fine-tuning often fails to extrapolate to new tasks, while experience retrieval often underperforms compared to supervised baselines. In this work, we combine these approaches and study how retrieval-augmented LLM agents can learn to use retrieved trajectories in-context. First, we establish a strong LoRA fine-tuning baseline that outperforms several state-of-the-art agent training pipelines. Second, we analyze key design choices for experience retrieval, including storage, querying, and trajectory selection. We then integrate experience retrieval directly into the fine-tuning process, finding that this substantially improves generalization to unseen tasks. Finally, we show that these gains often persist with imperfect experience and, even when agents reuse their own failed attempts without test-time parameter updates. Overall, our results establish simple episodic retrieval as a strong foundation for agent memory and \textbf{retrieval-aware fine-tuning as a practical and effective framework for building agents that learn to learn from experience}.
\end{abstract}

%% file: tables/intro_table_files/intro_table.tex
\ACLorOther{
\begin{table}[!h]
    \centering
    \newsavebox{\introtablebox}
    \sbox{\introtablebox}{%
    \begin{tabular}{lc}
    \toprule
    Method & ALFWorld \\
    \midrule
    \multicolumn{2}{l}{\quad\textbf{\textit{Prompting}}} \\
    Zero-shot & 29.9 \\
    ReAct (\citealp{react}) & 17.1\textsuperscript{a} \\
    ITP\textsubscript{I} (\citealp{imagine-then-plan2026}) & 35.7 \\
    \addlinespace[0.2em]
    \multicolumn{2}{l}{\quad\textbf{\textit{Training-Free Memory-Augmented}}} \\
    Mem0 (\citealp{Mem02025}) & 33.6\textsuperscript{b} \\
    A-MEM (\citealp{A-MEM2025}) & 34.7\textsuperscript{c} \\
    AgeMem-noRL (\citealp{AgenticMemory2026}) & 37.9 \\
    Memory Bank (\citealp{MemoryBank2024}) & 40.3\textsuperscript{d} \\
    Reflexion (\citealp{reflexion2023}) & 42.7\textsuperscript{e} \\
    \ours baseline (ours) & \textbf{83.6} \\
    \addlinespace[0.2em]
    \multicolumn{2}{l}{\quad\textbf{\textit{Supervised Fine-tuned}}} \\
    NAT (\citealp{NAT2024}) + ReAct & 66.4\textsuperscript{f} \\
    IWM (\citealp{EarlyExperience}) & 70.3 \\
    Self-Reflection (\citealp{EarlyExperience}) & 71.1 \\
    ETO (\citealp{song2024ETO}) + ReAct & 79.9\textsuperscript{f} \\
    SFT + ReAct (\citealp{react}) & 80.7\textsuperscript{f} \\
    SAND (\citealp{SAND2025}) & 85.0 \\
    ITP\textsubscript{R} (\citealp{imagine-then-plan2026}) & 85.1 \\
    Rule-based Expert & 89.6 \\
    LoRA baseline (ours) & \textbf{94.1} \\
    \bottomrule
    \end{tabular}
    }
    \ACLorOther
        {\resizebox{0.85\linewidth}{!}{\usebox{\introtablebox}}}
        {\usebox{\introtablebox}}
    \caption{\textbf{Complex solutions can underperform a strong baseline.} Success rate results for \textbf{Qwen 2.5-7B} on official held-out split of ALFWorld (valid-unseen) for training-free and supervised fine-tuned methods. \ACLorOther{This cross-paper comparison should be interpreted qualitatively.}{Despite using the same backbone model, this is a cross-paper comparison and should therefore be interpreted qualitatively, as different works may potentially use different experimental setups.} Unless noted, values come from original works; superscripts denote third-party reports used only when original papers do not report that model result: \textsuperscript{a}\citet{imagine-then-plan2026}, \textsuperscript{b}\citet{SkillRL2026}, \textsuperscript{c}\citet{AgenticMemory2026}, \textsuperscript{e}\citet{VerlAgentsLibrary2025}, \textsuperscript{d}\citet{GMemory2025}, \textsuperscript{f}\citet{embodiedtaskplanning2025}.}
    \OnlyACL{
        \vspace{-10pt}
    }
    \label{tab:our-baselines}
\end{table}
}{
    \begin{table*}[t]
        \centering
        \small
        \setlength{\tabcolsep}{4pt}
        \begin{subtable}[t]{0.48\textwidth}
            \centering
            \begin{tabular}{lc}
            \toprule
            Method & ALFWorld \\
            \midrule
            \multicolumn{2}{l}{\quad\textbf{\textit{Prompting}}} \\
            Zero-shot & 29.9 \\
            ReAct (\citealp{react}) & 17.1\textsuperscript{a} \\
            ITP\textsubscript{I} (\citealp{imagine-then-plan2026}) & 35.7 \\
            \addlinespace[0.2em]
            \multicolumn{2}{l}{\quad\textbf{\textit{Training-Free Memory-Augmented}}} \\
            Mem0 (\citealp{Mem02025}) & 33.6\textsuperscript{b} \\
            A-MEM (\citealp{A-MEM2025}) & 34.7\textsuperscript{c} \\
            AgeMem-noRL (\citealp{AgenticMemory2026}) & 37.9 \\
            Memory Bank (\citealp{MemoryBank2024}) & 40.3\textsuperscript{d} \\
            Reflexion (\citealp{reflexion2023}) & 42.7\textsuperscript{e} \\
            \ours baseline (ours) & \textbf{83.6} \\
            \bottomrule
            \end{tabular}
            \caption{\textbf{Training-free memory-augmented methods.}}
            \label{tab:intro-memory-baselines}
        \end{subtable}
        \hfill
        \begin{subtable}[t]{0.48\textwidth}
            \centering
            \begin{tabular}{lc}
            \toprule
            Method & ALFWorld \\
            \midrule
            \multicolumn{2}{l}{\quad\textbf{\textit{Prompting}}} \\
            Zero-shot & 29.9 \\
            ReAct (\citealp{react}) & 17.1\textsuperscript{a} \\
            ITP\textsubscript{I} (\citealp{imagine-then-plan2026}) & 35.7 \\
            \multicolumn{2}{l}{\quad\textbf{\textit{Supervised Fine-tuned}}} \\
            NAT (\citealp{NAT2024}) + ReAct & 66.4\textsuperscript{f} \\
            IWM (\citealp{EarlyExperience}) & 70.3 \\
            Self-Reflection (\citealp{EarlyExperience}) & 71.1 \\
            ETO (\citealp{song2024ETO}) + ReAct & 79.9\textsuperscript{f} \\
            SFT + ReAct (\citealp{react}) & 80.7\textsuperscript{f} \\
            SAND (\citealp{SAND2025}) & 85.0 \\
            ITP\textsubscript{R} (\citealp{imagine-then-plan2026}) & 85.1 \\
            Rule-based Expert & 89.6 \\
            LoRA baseline (ours) & \textbf{94.1} \\
            \bottomrule
            \end{tabular}
            \caption{\textbf{Supervised fine-tuned methods.}}
            \label{tab:intro-sft-baselines}
        \end{subtable}
        \caption{\textbf{Complex solutions can underperform a strong baseline.} Success rate results for \textbf{Qwen 2.5-7B} on official held-out split of ALFWorld (valid-unseen) for training-free and supervised fine-tuned methods. \ACLorOther{This cross-paper comparison should be interpreted qualitatively.}{Despite using the same backbone model, this is a cross-paper comparison and should therefore be interpreted qualitatively, as different works may potentially use different experimental setups.} Unless noted, values come from original works; superscripts denote third-party reports used only when original papers do not report that model result: \textsuperscript{a}\citet{imagine-then-plan2026}, \textsuperscript{b}\citet{SkillRL2026}, \textsuperscript{c}\citet{AgenticMemory2026}, \textsuperscript{e}\citet{VerlAgentsLibrary2025}, \textsuperscript{d}\citet{GMemory2025}, \textsuperscript{f}\citet{embodiedtaskplanning2025}.}
        \label{tab:our-baselines}
    \end{table*}    
}

%% file: Sections/1_intro.tex
Large language models (LLMs) have recently enabled general-purpose agents that can interact with textual environments, execute multi-step plans, and learn to solve families of tasks with minimal task-specific engineering~\citep{toolformer2023,Gorilla2024,voyager_minecraft,swebench2024,Sweagent2024,generativeagents2023}. In addition to sophisticated prompting, reflection and refinement pipelines~\citep{react,reflexion2023}, adaptation through parameter-efficient fine-tuning~\citep{lora,PEFTSurvey2024} is a natural step for building agents~\citep{fireact2023,autoact2024}. This approach often yields strong performance on tasks seen during training. However, we argue that \textbf{truly useful agents should be able to perform a broad range of tasks, including previously unseen ones}, within the environment in which they were trained to operate. Yet standard held-out benchmarks often evaluate tasks that remain very similar to those seen during training.

The in-context learning capabilities of modern LLMs make retrieval-augmented generation (RAG) an attractive inference-time adaptation mechanism, by providing relevant external context~\citep{izacard2022atlas,incontextRAGRAM2023}. For agents, a natural source of external context is prior experience, whether drawn from the agent's own rollouts, other agents, or expert demonstrations. Retrieving relevant experience and feeding it in-context to the model has been proposed in prior work such as RAP \citep{rap_2024}, and contemporaneous work by~\citet{wei2025EVOmemory} coined the term \oursraw, which we adopt in this paper.

This paper studies how to \emph{train} retrieval-augmented LLM agents that can \emph{learn from retrieved experience in-context}, enabling better \emph{generalization to unseen tasks} within the same environment. We focus on a practical setting: moderately-sized open models adapted with supervised LoRA fine-tuning~\citep{lora}, operating in text-based environments with multi-turn decision making.

\paragraphshort{Contribution \ding{182}: Strong fine-tuning and retrieval-augmented baselines.} Our first contribution is to provide effective recipes for both fine-tuning and retrieval-augmented inference. We noticed that the performance of fine-tuned models varies greatly in existing literature, and therefore propose our own LoRA baseline. In Table~\ACLorOther{\ref{tab:our-baselines}}{\ref{tab:intro-sft-baselines}}, we show that on the widely used ALFWorld benchmark, our LoRA baseline outperforms existing training pipelines, including more elaborate agentic methods. \ACLorOther{The same table further suggests}{Results in Table~\ref{tab:intro-memory-baselines} further suggest} that elaborate memory-augmented pipelines (without training) can provide limited improvement or even underperform compared with a simple experience-retrieval baseline (Best \oursraw)\ACLorOther{,}{.}%
\ACLorOther{
    consistent with \citeposs{wei2025EVOmemory} observation that simple episodic retrieval is still competitive with self-evolving memory systems.
}{
    Although this comparison should be interpreted qualitatively, as it aggregates results from prior works with potentially different experimental setups, its overall takeaway is consistent with the contemporaneous findings of \citet{wei2025EVOmemory}: simple episodic retrieval can be highly competitive with more complex self-evolving memory systems. 
}
Both results call for \textbf{benchmarking against strong simple baselines before attributing gains to complex memory architectures}.

\paragraphshort{Contribution \ding{183}: A careful examination of \oursraw.}%
Second, although RAG has mature recipes for document retrieval, yet no analogous recipe is established for retrieving and using agent trajectories. In Section~\ref{sec:experiments:zero_shot}, we systematically study the agent-specific design choices: how to store experience, how to build a retrieval query, how much and when to retrieve, and how the source of indexed trajectories and the backbone model affect performance. These results provide a practical recipe for robust and effective adaptation through \oursraw.

\paragraphshort{Contribution \ding{184}: Training agents to use retrieved experience (\oursraw-LoRA).}%
We propose to combine LoRA fine-tuning with \oursraw, training the policy to use retrieved experience rather than introducing it only at inference time. In Sec.~\ref{sec:experiments:finetuning}, \oursraw-LoRA substantially improves generalization to unseen task groups while retaining strong performance on seen tasks, showing that memory gains depend not only on retrieving useful experience, but also on learning how to use it.

\paragraphshort{Contribution \ding{185}: Learning from imperfect and self-generated experience.} Finally, we stress-test \oursraw-LoRA beyond matched expert demonstrations using sparse, mismatched, stronger-LLM-generated, and self-generated experience at inference time. It often remains robust to imperfect but relevant memory and, in some cases, can even benefit from its own failed attempts without parameter updates.

Taken together, these results establish simple episodic retrieval as a strong foundation for agent memory, while showing that effective memory depends not only on \emph{what} experience is stored and retrieved, but also on whether the agent has learned \emph{how to use it}, a capability that can transfer from expert experience during training to unseen tasks and imperfect experience at inference.

%% file: Sections/2_related_work.tex
\section{Background and Related Work}
\label{sec:related_work}

LLM agents have been studied in interactive settings such as text environments, web navigation, and tool use. We position our work at the intersection of agent training and episodic retrieval: the agent retrieves past trajectories, rather than documents, and we study how that retrieval should be used both at inference time and during fine-tuning.

\paragraphshort{LLM-based agents.}
A common strategy for LLM agents is to define a fixed interaction protocol and adapt an instruction-tuned model on expert trajectories via supervised fine-tuning (SFT) / behavior cloning~\citep{fireact2023,AgentFLAN24}. This usually improves in-domain scores and output-format reliability, but generalization to unseen tasks often remains limited due to distribution shift and compounding errors in sequential decision making~\citep{song2024ETO}. To improve robustness, one line of work initializes from SFT and further learns through trial-and-error, often with reinforcement learning or preference-based updates\ACLorOther{~\citep{song2024ETO,EarlyExperience,VerlAgentsLibrary2025}.}{~\citep{song2024ETO,EarlyExperience,VerlAgentsLibrary2025,GLIDER}.} 
However, we find that simple SFT with LoRA can already outperform more complex methods in our settings, so we instead explore an orthogonal direction: experience retrieval.

\paragraphshort{Retrieval-augmented generation (RAG).}
Classical RAG augments a parametric generator with retrieved passages from an external corpus~\citep{lewis2020retrieval,guu2020realm,incontextRAGRAM2023}, typically for tasks such as open-domain QA and knowledge-intensive generation. Some systems integrate retrieval more tightly into the architecture, training the model to process retrieved context (e.g., RETRO, \citealt{borgeaud2022RETRO}; Atlas, \citealt{izacard2022atlas}; and RAFT, \citealt{raft2024}). 
\NotACL{From a memory perspective, RAG can be viewed as endowing LLMs with an \textit{external long-term semantic memory} (the corpus and its index), whose retrieved items are loaded into the model's \textit{working memory} via in-context augmentation to ground and steer generation \citep{wu2025human}. }
We borrow this retrieval-as-context view, but replace \textit{semantic} memory with \textit{episodic} memory: instead of retrieving documents to ground factual generation, the agent retrieves prior interaction trajectories to guide action selection in a new situation.

\paragraphshort{Memory for agents.}
Broader agent-memory systems extend retrieval with explicit writing, compression, reflection, and control mechanisms. Recent surveys distinguish classical RAG from \textit{agent memory}, which typically involves (i) what gets written (episodic traces, feedback, reflections, skills), (ii) how memory is structured or compressed (summaries, abstractions, procedural artifacts), and (iii) control over when and how memory is retrieved and updated (e.g., step-wise loops, learned or heuristic read/write policies)~\citep{memoryageaiagents_survey2026,memory_mechanism_survey2025,wu2025human,AgenticMemory2026}. 
Representative examples include systems that manage long-term context or learn read/write policies (e.g., MemGPT, \citealt{MEMGPT}), and methods that store reflective summaries or compact experience units rather than full trajectories, such as Reflexion~\citep{reflexion2023}, A-MEM~\citep{A-MEM2025}, and Memento~\citep{memento2025}. Unlike these approaches, we do not propose a new memory controller, online memory update rule, or reflection mechanism. Our memory is intentionally simpler: a fixed, read-only bank of full trajectories retrieved into context. Closest to our setting, contemporaneous work by \citet{wei2025EVOmemory} compares inference-only episodic retrieval with self-evolving read--write memory, using strong proprietary models. By contrast, we study weaker open-source models under both inference-time and fine-tuning retrieval augmentation pipelines.

\paragraphshort{Learning from experience.}
Beyond storing past interactions, a complementary line of work studies how agents improve from them. Classical meta-learning trains models to adapt rapidly from new examples or trajectories \citep{metalearningsurvey,MAML}, while recent LLM-agent methods reuse past trials either through external memory or parameter updates. Reflexion and ExpeL transform prior interactions into reusable verbal feedback or distilled experience \citep{ExpeL,reflexion2023}, whereas methods such as ETO, NAT, and Early Experience use successful or failed trajectories as supervision for further training \citep{song2024ETO,NAT2025,EarlyExperience}. Our setting is complementary: we keep experience as raw retrieved trajectories and train the policy itself to use them in context, then test whether this ability transfers to sparse, teacher-generated, and self-generated experience without test-time parameter updates.

\paragraphshort{Position of this work.}
Our setting differs from prior memory-based agents along four experiment-relevant axes: we retrieve full trajectories rather than summaries or reflections, keep memory fixed rather than updating it online, study retrieval during training as well as at inference time, and distinguish gains from retrieval and from parameter updates. Accordingly, our goal is not to propose a rich read--write memory architecture, but to establish strong foundations for fine-tuning augmented with episodic retrieval: (i) we propose strong baselines and identify their main drivers of success, (ii) we investigate whether retrieval-augmented fine-tuning teaches agents a reusable ability to learn from retrieved experience on unseen tasks, including under imperfect experience, and (iii) we characterize when this adaptation succeeds or fails as the availability, source, and quality of experience change.

%% file: Sections/3_method.tex
\section{\oursraw: Experience Retrieval-Augmented Generation}
\label{sec:method}

Purely parametric learning (e.g.\ LLM fine-tuning with LoRA) can fit the training-tasks distribution well, but may fail to generalize to new tasks. Retrieval adds an adaptive, inference-time mechanism: the agent can condition its generation on relevant past experience and apply it in-context without updating parameters. In that case, our goal is to \textbf{build agents that learn to reliably use retrieved experience when it is available, rather than simply memorize cases seen during training}. In this section we motivate the use of retrieval for in-context learning and formalize our approach for experience RAG (which we refer to as \oursraw).

\paragraphshort{LLM-based Agents.}
We consider an agent interacting with a textual environment over at most \(T\) decision steps for a task description \(\mathcal{T}\) provided as text. At each step \(t\), the agent observes \(o_t\), outputs an action \(a_t\), and receives the next observation \(o_{t+1}\) from the environment or a terminal flag when the task is completed. We denote the full trajectory by
$\tau=(\mathcal{T},o_1,a_1,\ldots,o_{T-1},a_{T-1},o_T)$,
and the trajectory history available at step \(t\) by
$h_t=(\mathcal{T},o_1,a_1,\ldots,o_t)$.
An LLM policy \(\pi_\theta\) defines a conditional distribution over actions given history, i.e., \(a_t \sim \pi_\theta(\cdot \mid h_t)\). 
Our goal is to learn \(\pi_\theta\) that generalizes to \emph{unseen} tasks within the same environment, where interaction dynamics are fixed but required action sequences may differ. This includes tasks requiring action types not observed during training (e.g., task is to heat an object but agent was trained to clean an object).

\paragraphshort{Encoding Agent Trajectories as Multi-turn LLM Chats.} Given a trajectory \(\tau\), we serialize the interaction as a multi-turn chat \(chat(\tau)\) with the base model's native chat template, mapping observations to user turns and actions to assistant turns. At step \(t\), the policy conditions on \(chat(h_t)\) and generates an action as a natural-language token sequence \(a_t=(a_t^1,\ldots,a_t^{n_t})\), autoregressively \(a_t^i \sim \pi_\theta(\cdot \mid chat(h_t), a_t^{<i})\). This serialization follows multi-turn agent formulations~\citep{PractitionersGuideAgents25,RAGEN2025,Search-R1-25}, with supervised next-token prediction on assistant tokens as objective. It contrasts with more recently employed \emph{stepwise} encodings~\citep{EarlyExperience,wei2025EVOmemory,VerlAgentsLibrary2025}, which serialize \((h_t,a_t)\) as independent samples and re-encode history at every step. In our early explorations, we observe similar task performance but substantially faster training for multi-turn chat due to KV-cache reuse across turns.

\paragraphshort{Indexing and Retrieval.}
We maintain an index $I=\{(\tau_i,e_i)\}_{i=1}^N$ of textual trajectories and their key embeddings, where \(e_i=\phi(\tau_i)\) is computed by a trajectory encoder \(\phi(\cdot)\). Trajectories are stored as raw chat-formatted data without further filtering or aggregation. At decision step \(t\), we build a textual query \(q_t\) from the current agent context and retrieve the top-$K$ trajectories by nearest-neighbor search:
\[
(\tau^1,\ldots,\tau^K)=\arg\underset{1\le i\le N}{\mathrm{topK}} \ \phi(q_t)\cdot e_i .
\]

\OnlyACL{
    \vspace{-6pt}
}

\paragraphshort{Experience-Conditioned Generation.} We then format retrieved trajectories into a memory block \(m_t = \mathrm{system}(\tau^1, \allowbreak  \ldots,\tau^K)\), inserted in the system prompt. When available, we distinguish successful and unsuccessful trajectories in the memory block prompt. The action \(a_t^i\) is generated autoregressively conditioned on both memory \(m_t\) and dialogue history \(chat(h_t)\):
\ACLorArxivorOther{
\[
    \begin{aligned}
    a_t^i &\sim \pi_\theta\!\left(
    \cdot \mid m_t,\; chat(h_t),\; a_t^{<i}
    \right).
    \end{aligned}
\]
}{
\[
    \begin{aligned}
        a_t^i &\sim \pi_\theta\!\left(
        \cdot \mid m_t,\; chat(h_t),\; a_t^{<i}
        \right).
    \end{aligned}
\]
}{
\[
    \begin{aligned}
        a_t^i &\sim \pi_\theta\!\left(
        \cdot \mid m_t,\; chat(h_t),\; a_t^{<i}
        \right).
    \end{aligned}
\]
}

For \oursraw, we assume that, for out-of-distribution tasks unseen during training, a small set of in-domain reference trajectories can be collected and used as in-context examples. These trajectories may come either from expert demonstrations (e.g., rule-based or stronger models) or from previously recorded agent rollouts generated with a strong prompt. We provide further details on index construction and retrieval in Appendix~\ref{app:implementation-details}.

%% file: Sections/4_experiments.tex
\section{Experiments}
\label{sec:experimental_setup}

We evaluate retrieval-augmented agents on text-based interactive benchmarks with multi-step decision making. Our goals are to measure gains on \emph{unseen} tasks and understand which design choices matter most.%

\subsection{Dataset and Benchmarks}
\label{sec:experiments:dataset-and-benchmarks}

\paragraphshort{Environments.}
We evaluate on two commonly used text-based embodied environments: \textbf{ALFWorld}~\citep{ALFWorld20} and \textbf{ScienceWorld}~\citep{wang-etal-2022-scienceworld}. ALFWorld involves household manipulation tasks with binary episode outcome (success/failure). ScienceWorld contains science-oriented tasks aligned with an elementary-school science curriculum and provides a dense episode score in \([-1,1]\), with success at 1. For consistency, we report \textbf{success rate}. For ScienceWorld, we convert the episode outcome to a binary success/failure score. %

\paragraphshort{Benchmarks primarily measure within-group generalization.}
Both benchmarks associated with the environments provide official held-out splits (ALFWorld: \textit{valid-unseen}; ScienceWorld: \textit{test}), where tasks are held out within predefined task groups (ALFWorld contains 6 ``\textit{task-types}" while ScienceWorld contains 10 ``\textit{topics}"). However, sampling splits within each group makes the test set still close to the training distribution and thus less realistic, as agents are evaluated on tasks very similar to the ones they were trained on. This is demonstrated by the performance obtained in preliminary runs, reported in Table~\ref{tab:our-baselines}, where standard LoRA fine-tuning already achieves very strong performance on ALFWorld, surpassing rule-based expert and nearly saturating the benchmark.

\paragraphshort{Building Out-of-Distribution Benchmarks.}
To evaluate stronger out-of-distribution generalization and the ability of the model to adapt at test time from past experiences with \oursraw, we partition the task groups into \textbf{easy} (used for training trajectory collection and fine-tuning) and \textbf{hard} (held out for evaluation). In our experiments, we report results separately on easy and hard tasks to isolate generalization. Extra details on these splits are available in Appendix~\ref{app:implementation-details:data-statistics}.

\subsection{Implementation Details}
\label{sec:experiments:implementation-details}

\paragraphshort{Data and Indexing.}
Across all experiments, we instantiate experience as scripted expert trajectories generated by the environment-provided policy. This controlled choice allows us to study retrieval-augmented adaptation under a reliable memory source, avoiding an additional confound from noisy trajectories given that zero-shot backbones perform poorly. Importantly, these trajectories still contain failures and suboptimal segments, requiring the model to learn to identify which retrieved actions are relevant. We serialize trajectories using the chat template described in Section~\ref{sec:method}\OnlyArxiv{. Each trajectory contains an environment-specific system prompt, user turns for the task description and subsequent observations, and assistant turns for actions} (full templates in Appendix~\ref{app:prompts}). Unless otherwise stated, the retrieval index is built from the training split and includes both successful and unsuccessful episodes. For fine-tuning, we use only successful trajectories from the training split. We use successful trajectories from the corresponding validation split for model selection and report results on the corresponding held-out split; each split is further partitioned into Easy and Hard subsets.

\OnlyArxiv{
    \paragraphshort{Models.}
    We use only instruction-tuned base LLMs as backbones: Ministral 3-8B, Gemma 3-4B, Qwen 2.5-7B, and Qwen 2.5-7B-1M. We implement training with TorchTune~\citepAliasYear{torchtune} and run inference with Hugging Face Transformers~\citep{HFTransformers}, which is more efficient in our setup. Unless otherwise stated, decoding is greedy (temperature \(=0\)). Additional model details are provided in Appendix~\ref{app:implementation-details:models}.
}

\paragraphshort{System Prompt.}
For each environment, we use a minimal system prompt that specifies (i) the task setting, (ii) the action interface, and (iii) the response format (one action per turn). Concretely, at the beginning of each episode, we provide a static list of action \emph{templates} valid in the entire environment (e.g., \textit{go to [receptacle]}, \textit{use [object]}), rather than step-specific instantiated valid actions. In contrast, some prior work provides per-step valid-action candidates during rollout (e.g.,~\citealp{VerlAgentsLibrary2025,EarlyExperience}); for example, if only \textit{table 1} and \textit{cabinet 4} are currently relevant, candidates such as \textit{go to table 1}, \textit{open cabinet 4}, (\ldots) are shown. We avoid this additional guidance to reduce action-space leakage and evaluate stronger decision-making from context alone. %
Full prompt templates are provided in Appendix~\ref{app:prompts}.

\paragraphshort{Training Setup.}
In Section~\ref{sec:experiments:finetuning}, we study adaptation with low-rank adapters (LoRA)~\citep{lora}. We supervise models on agent trajectories formatted as multi-turn chats, as described in Sec.~\ref{sec:method}. We compute cross-entropy loss on assistant tokens only.\NotACL{ All models are trained on a single 80~GB NVIDIA A100 GPU.} Key hyperparameters and implementation details are provided in Appendix~\ref{app:implementation-details}.

%% file: Sections/5_zero_shot.tex
\subsection{Retrieval-Augmented Inference without Training}
\label{sec:experiments:zero_shot}
We first evaluate \ours \emph{only at inference time}: the policy remains a frozen instruction-tuned model, and retrieval is enabled by prepending retrieved trajectories as in-context memory. Thus, in this section, all conditions use the same instruction-tuned checkpoint and differ only in which retrieved trajectories are added to the prompt.

\paragraphshort{Ablations.}
\ACLorArxivorOther{
    We vary: (i) top-$k$ (\(K \in \{1,2,4\}\)), (ii) index composition (\textbf{all}/\textbf{easy}/\textbf{hard} task-types available for retrieval), and (iii) retrieval mode. \textbf{Static retrieval} selects trajectories once at episode start. \textbf{Dynamic retrieval} re-queries at every step using the partial interaction history, which requires clearing the KV cache and re-encoding the prompt with updated retrieved examples up to the current observation.
}{
    In this experiment, we investigate the impact of three parts of the retrieval setup:
    \begin{enumerate}
        \item \textbf{Top-$k$}: the number of retrieved trajectories added to the prompt, with \(K \in \{1,2,4\}\).
        \item \textbf{Retrieval mode}: whether retrieval is performed once (\textbf{static} retrieval) or refreshed during interaction (\textbf{dynamic} retrieval).
        \item \textbf{Index composition}: whether the retrieval index contains trajectories from train/\textbf{all}, only train/\textbf{easy}, or only train/\textbf{hard} task types.
    \end{enumerate}
    \textbf{Static retrieval} selects trajectories once at episode start. \textbf{Dynamic retrieval} re-queries at every step using the partial interaction history, which requires clearing the KV cache and re-encoding the prompt with updated retrieved examples up to the current observation.
}{
    We vary: (i) index composition (\textbf{all}/\textbf{easy}/\textbf{hard} task-types available for retrieval), (ii) top-$k$ (\(K \in \{1,2,4\}\)), and (iii) retrieval mode. \textbf{Static retrieval} selects trajectories once at episode start. \textbf{Dynamic retrieval} re-queries at every step using the partial interaction history, which requires clearing the KV cache and re-encoding the prompt with updated retrieved examples up to the current observation.
}

For this experiment, static retrieval uses task descriptions as queries/keys; dynamic retrieval uses partial trajectories as queries and full trajectories as keys (JSON). This serialization affects retrieval embeddings only, not policy-side display. We investigate the impact of this choice in Appendix~\ref{subapp:trajectory_formatting}.

\ACLorOther{
    Table~\ref{tab:zero-shot} shows inference-only \ours results for Ministral 3-8B on ALFWorld and ScienceWorld; other backbones are in Appendix~\ref{subapp:extra_models}.
}{
    Table~\ref{tab:zero-shot} reports inference-only \ours results for Ministral 3-8B on ALFWorld and ScienceWorld; results for other backbones are in Appendix~\ref{subapp:extra_models}.
}

\input{tables/zero_shot_ministral}

\subsubsection{Discussion}

\paragraphshort{Inference-only Experience Retrieval improves in all scenarios.}
\ACLorOther{
    Adding retrieved trajectories at inference time strongly improves over the \textsc{No RAG} baseline on both benchmarks. The best all-task score rises from \(4.48\%\) to \(64.18\%\) on ALFWorld and from \(10.40\%\) to \(35.24\%\) on ScienceWorld. Gains appear in all scenarios, including when the retrieval index is mismatched with the evaluation split, indicating that retrieved trajectories help not only by recalling near-matching solutions, but also by providing reusable action patterns and subgoal strategies.
}{
    Enabling retrieval at inference time substantially outperforms the strictly zero-shot \textsc{No RAG} baseline across both benchmarks. On ALFWorld, the best all-task result rises from \(4.48\%\) to \(64.18\%\) with inference-only \oursraw (static retrieval, index=\texttt{all}, \(K{=}4\)); on ScienceWorld, it rises from \(10.40\%\) to \(35.24\%\) (dynamic retrieval, index=\texttt{all}, \(K{=}4\)). More importantly, retrieval helps across all evaluated settings, including those where the retrieval index is mismatched with the evaluation split. This suggests that retrieved trajectories are useful not only as directly matched solutions, but also as in-context guidance for action formatting, subgoal decomposition, and partial strategy transfer, enabling strong training-free gains from a frozen policy alone.
}

\paragraphshort{Choosing top-$K$: more experiences vs. context rot.}
\ACLorOther{
    Increasing $K$ generally improves performance, but the gains are much larger on ALFWorld than on ScienceWorld. The best all-task score on ALFWorld rises from \(41.04\%\) at \(K{=}1\) to \(64.18\%\) at \(K{=}4\), whereas ScienceWorld improves at \(K{=}2\) (\(26.62\%\rightarrow34.67\%\)) and only marginally to \(35.24\%\) at \(K{=}4\), suggesting earlier saturation. We also observed more sensitivity in mismatched index settings. These results suggest that larger $k$ helps when they add complementary evidence, but can hurt when they introduce noisy or weakly transferable context. Since inference cost grows roughly linearly with \(K\), we use \(K{=}2\) in later experiments as a better cost-performance trade-off.
    We report inference cost in Appendix~\ref{app:efficiency}.
}{
    Increasing the number of retrieved trajectories generally improves performance, but the gains are much more consistent on ALFWorld than on ScienceWorld. On ALFWorld, the best all-task score rises from \(41.04\%\) at \(K{=}1\) to \(64.18\%\) at \(K{=}4\), and larger \(K\) also usually helps on hard tasks when the retrieval index is well aligned (e.g., \texttt{all} or \texttt{hard}). On ScienceWorld, by contrast, the best all-task score increases from \(26.62\%\) to \(34.67\%\) and then only marginally to \(35.24\%\), suggesting earlier saturation. On hard tasks, the effect of larger \(K\) is also more sensitive to mismatched index, and performance can even decline in some settings (e.g., static retrieval with the \texttt{hard} index: \(17.97\%\rightarrow15.36\%\rightarrow13.02\%\)). Overall, additional retrieved trajectories help when they provide complementary, task-relevant evidence, but they can hurt when they are noisy, conflicting, or weakly transferable, or when the model cannot effectively use the longer context. This effect is weaker for long-context models: Qwen~2.5~7B~1M benefits more from larger \(K\) (Appendix~\ref{subapp:extra_models}), consistent with the view that context-handling capacity limits how much additional in-context experience the policy can exploit. Because inference cost grows roughly linearly with \(K\), we use \(K{=}2\) as the default in later experiments: it captures most of the zero-shot gain while avoiding much of the prompt-length and latency cost of \(K{=}4\). We report and provide more details for inference cost in Appendix~\ref{app:efficiency}.
}

\paragraphshort{Refreshed context relevance vs. instability in dynamic retrieval.}
\ACLorOther{
    Dynamic retrieval helps mainly when the retrieval index is aligned with the evaluation setting, but is less stable than static retrieval under index mismatch. The largest failures occur when hard tasks retrieve from the \texttt{easy} index, suggesting that these trajectories may provide local action cues but often lack the task-specific strategy needed for later subgoals. A likely reason is that dynamic retrieval replaces retrieved context at every step, creating context churn and making later decisions more sensitive to retrieval quality (similar to effects reported by~\citet{LossyProvenanceMemory2026,RAGRobustnessSpuriousFeatures2025}). Overall, re-retrieval helps only when the refreshed context remains sufficiently relevant.
}{
    Dynamic retrieval is helpful primarily when the retrieval index is matched to the evaluation setting, improving performance in nearly all such cases. However, it is less stable than static retrieval when the index is mismatched. On hard tasks with the \texttt{all} index, gains saturate quickly and can even reverse at larger \(K\), (e.g., on ALFWorld at \(K{=}4\), \(65.03\%\rightarrow55.74\%\)), while on ScienceWorld the gains remain small. The largest failures occur when retrieving from the \texttt{easy} index on hard tasks, suggesting that while these trajectories may still help with action formatting or local interaction patterns, they often fail to provide the task-specific strategy needed for later subgoals. This may be explained by the fact that dynamic retrieval replaces the retrieved context at every step. This can create \textit{context churn}: past actions remain in the prompt, but the trajectories that originally supported them are removed, leaving only a lossy trace of the earlier reasoning evidence and making later decisions more sensitive to retrieval quality and newly introduced context, similar to effects reported by~\citet{LossyProvenanceMemory2026} and~\citet{RAGRobustnessSpuriousFeatures2025}. Overall, re-retrieval helps only when the relevance of the refreshed memory block is high enough to outweigh the instability introduced by repeated prompt updates; otherwise, those updates can amplify noisy or weakly transferable trajectories.
}

\paragraphshort{Index composition controls which tasks benefit.}
\ACLorOther{
    Performance is generally weaker when the retrieval index is mismatched with the evaluation setting, especially when hard tasks retrieve from the \texttt{easy} index, although some cross-split gains remain. This suggests that easy-task trajectories can still provide reusable sub-skills or partial reasoning cues. At larger \(K\), ALFWorld often benefits more from the \texttt{hard} index, whereas ScienceWorld more often benefits from the \texttt{easy} index. A plausible explanation is that ALFWorld hard tasks largely recombine easy sub-tasks, so hard trajectories provide reusable multi-step plans. In ScienceWorld, however, hard trajectories may contain more failed, longer, or less reusable traces that an inference-only model cannot reliably filter.
}{
    We observe weaker performance when the retrieval index is mismatched with the evaluation setting, especially when hard tasks retrieve from the \texttt{easy} index. Still, some cross-split gains remain. This is plausible: in ALFWorld, hard tasks often compose sub-tasks that already appear in easy tasks, while in ScienceWorld, easy trajectories can still provide partial reasoning cues within the same topic. We also observe a benchmark asymmetry at larger \(K\): on ALFWorld, all-task performance often benefits more from the \texttt{hard} index than from the \texttt{easy} index, whereas ScienceWorld often shows the opposite. One possible explanation is that ALFWorld hard trajectories condense longer multi-step solutions that remain useful at inference time (due to the subtasks composition), while ScienceWorld hard trajectories may contain more failed, longer, or less reusable traces that an inference-only model cannot reliably filter.
}

\upshot{Retrieval-augmented inference is a strong training-free mechanism for adapting to unseen tasks; assuming sufficiently relevant retrieval, more experience generally helps up to the model’s effective context capacity, so a good practice is to use the highest top-\(K\) before performance saturates or degrades.}

%% file: tables/zero_shot_ministral.tex
\begin{table*}[t]
\centering
\small
\setlength{\tabcolsep}{5pt}
\newsavebox{\zeroshotministralbox}
\sbox{\zeroshotministralbox}{%
\begin{tabular}{l c c c c c c c c}
\toprule
\multirow{2}{*}{\makecell{\oursraw\\type}} & \multirow{2}{*}{Top-$K$} & \multirow{2}{*}{Index} & \multicolumn{3}{c}{ALFWorld} & \multicolumn{3}{c}{ScienceWorld} \\
\cmidrule(lr){4-6} \cmidrule(lr){7-9}
& & & Easy tasks & Hard tasks & All tasks & Easy tasks & Hard tasks & All tasks \\
\midrule
No RAG & 0 & -- & 8.22 & 0.00 & 4.48 & 14.41 & 5.08 & 10.40 \\
\midrule
\multirow{3}{*}{static} & \multirow[c]{6}{*}{1} & all & 43.84 & 33.88 & 39.30 & 33.33 & 17.71 & 26.62 \\
 &  & easy & 42.93 & 5.47 & 25.87 & 28.83 & 10.16 & 20.81 \\
 &  & hard & 35.62 & 33.88 & 34.83 & 22.65 & \textbf{17.97} & 20.64 \\
\addlinespace[0.4em]
\multirow{3}{*}{dynamic} &  & all & 46.92{\scriptsize \textcolor{green!50!black}{\,$\uparrow 3.1$}} & 34.02{\scriptsize \textcolor{green!50!black}{\,$\uparrow 0.1$}} & 41.04{\scriptsize \textcolor{green!50!black}{\,$\uparrow 1.7$}} & 29.81{\scriptsize \textcolor{red!70!black}{\,$\downarrow 3.5$}} & 17.71{\scriptsize \,0.0} & 24.61{\scriptsize \textcolor{red!70!black}{\,$\downarrow 2.0$}} \\
 &  & easy & 44.75{\scriptsize \textcolor{green!50!black}{\,$\uparrow 1.8$}} & 4.37{\scriptsize \textcolor{red!70!black}{\,$\downarrow 1.1$}} & 26.37{\scriptsize \textcolor{green!50!black}{\,$\uparrow 0.5$}} & 32.65{\scriptsize \textcolor{green!50!black}{\,$\uparrow 3.8$}} & 7.03{\scriptsize \textcolor{red!70!black}{\,$\downarrow 3.1$}} & 21.65{\scriptsize \textcolor{green!50!black}{\,$\uparrow 0.8$}} \\
 &  & hard & 38.82{\scriptsize \textcolor{green!50!black}{\,$\uparrow 3.2$}} & 39.89{\scriptsize \textcolor{green!50!black}{\,$\uparrow 6.0$}} & 39.30{\scriptsize \textcolor{green!50!black}{\,$\uparrow 4.5$}} & \textbf{26.18}{\scriptsize \textcolor{green!50!black}{\,$\uparrow 3.5$}} & 19.53{\scriptsize \textcolor{green!50!black}{\,$\uparrow 1.6$}} & 23.32{\scriptsize \textcolor{green!50!black}{\,$\uparrow 2.7$}} \\
\midrule
\multirow{3}{*}{static} & \multirow[c]{6}{*}{2} & all & 53.42 & 48.09 & 50.99 & \textbf{44.31} & 21.88 & \textbf{34.67} \\
 &  & easy & 47.27 & \textbf{11.48} & 30.97 & \textbf{43.53} & 12.24 & \textbf{30.09} \\
 &  & hard & 39.73 & 46.99 & 43.04 & 24.12 & 15.36 & 20.36 \\
\addlinespace[0.4em]
\multirow{3}{*}{dynamic} &  & all & 57.53{\scriptsize \textcolor{green!50!black}{\,$\uparrow 4.1$}} & 48.36{\scriptsize \textcolor{green!50!black}{\,$\uparrow 0.3$}} & 53.36{\scriptsize \textcolor{green!50!black}{\,$\uparrow 2.4$}} & 43.53{\scriptsize \textcolor{red!70!black}{\,$\downarrow 0.8$}} & 21.88{\scriptsize \,0.0} & 34.23{\scriptsize \textcolor{red!70!black}{\,$\downarrow 0.4$}} \\
 &  & easy & 57.53{\scriptsize \textcolor{green!50!black}{\,$\uparrow 10.3$}} & 6.56{\scriptsize \textcolor{red!70!black}{\,$\downarrow 4.9$}} & 34.33{\scriptsize \textcolor{green!50!black}{\,$\uparrow 3.4$}} & 43.73{\scriptsize \textcolor{green!50!black}{\,$\uparrow 0.2$}} & 9.64{\scriptsize \textcolor{red!70!black}{\,$\downarrow 2.6$}} & 29.08{\scriptsize \textcolor{red!70!black}{\,$\downarrow 1.0$}} \\
 &  & hard & 39.27{\scriptsize \textcolor{red!70!black}{\,$\downarrow 0.5$}} & 48.09{\scriptsize \textcolor{green!50!black}{\,$\uparrow 1.1$}} & 43.28{\scriptsize \textcolor{green!50!black}{\,$\uparrow 0.2$}} & 25.10{\scriptsize \textcolor{green!50!black}{\,$\uparrow 1.0$}} & 22.92{\scriptsize \textcolor{green!50!black}{\,$\uparrow 7.6$}} & \textbf{24.16}{\scriptsize \textcolor{green!50!black}{\,$\uparrow 3.8$}} \\
\midrule
\multirow{3}{*}{static} & \multirow[c]{6}{*}{4} & all & \textbf{63.47} & \textbf{65.03} & \textbf{64.18} & 42.06 & \textbf{22.27} & 33.56 \\
 &  & easy & \textbf{63.01} & 7.11 & \textbf{37.56} & 20.39 & \textbf{17.71} & 19.24 \\
 &  & hard & \textbf{40.19} & \textbf{62.84} & \textbf{50.50} & \textbf{41.57} & 13.02 & \textbf{29.31} \\
\addlinespace[0.4em]
\multirow{3}{*}{dynamic} &  & all & \textbf{70.55}{\scriptsize \textcolor{green!50!black}{\,$\uparrow 7.1$}} & \textbf{55.74}{\scriptsize \textcolor{red!70!black}{\,$\downarrow 9.3$}} & \textbf{63.81}{\scriptsize \textcolor{red!70!black}{\,$\downarrow 0.4$}} & \textbf{44.71}{\scriptsize \textcolor{green!50!black}{\,$\uparrow 2.7$}} & \textbf{22.66}{\scriptsize \textcolor{green!50!black}{\,$\uparrow 0.4$}} & \textbf{35.24}{\scriptsize \textcolor{green!50!black}{\,$\uparrow 1.7$}} \\
 &  & easy & \textbf{71.69}{\scriptsize \textcolor{green!50!black}{\,$\uparrow 8.7$}} & \textbf{9.29}{\scriptsize \textcolor{green!50!black}{\,$\uparrow 2.2$}} & \textbf{43.28}{\scriptsize \textcolor{green!50!black}{\,$\uparrow 5.7$}} & \textbf{46.67}{\scriptsize \textcolor{green!50!black}{\,$\uparrow 26.3$}} & \textbf{10.94}{\scriptsize \textcolor{red!70!black}{\,$\downarrow 6.8$}} & \textbf{31.32}{\scriptsize \textcolor{green!50!black}{\,$\uparrow 12.1$}} \\
 &  & hard & \textbf{42.93}{\scriptsize \textcolor{green!50!black}{\,$\uparrow 2.7$}} & \textbf{65.57}{\scriptsize \textcolor{green!50!black}{\,$\uparrow 2.7$}} & \textbf{53.24}{\scriptsize \textcolor{green!50!black}{\,$\uparrow 2.7$}} & 17.25{\scriptsize \textcolor{red!70!black}{\,$\downarrow 24.3$}} & \textbf{23.96}{\scriptsize \textcolor{green!50!black}{\,$\uparrow 10.9$}} & 20.14{\scriptsize \textcolor{red!70!black}{\,$\downarrow 9.2$}} \\
\bottomrule
\end{tabular}%
}
\ACLorArxivorOther{
    \resizebox{0.87\textwidth}{!}{\usebox{\zeroshotministralbox}}
}{
    \usebox{\zeroshotministralbox}
}{
    \resizebox{0.95\textwidth}{!}{\usebox{\zeroshotministralbox}}
}
\caption{\textbf{Results for \ours inference without training across ALFWorld and ScienceWorld for Ministral 3-8B.} Values are mean success rates over \(3\) seeds; full mean \(\pm\) std for every entry is reported in Appendix~\ref{subapp:random_seeds}. Across entries, the median/max std is \(1.58/6.20\) points on ALFWorld and \(1.83/4.45\) on ScienceWorld. The \textbf{bold} values mark the best top-$k$ setting for each (\ours type, index) within a given dataset split. For dynamic retrieval rows, the superscript arrows report the per-cell difference against the corresponding static row with the same index and top-$k$: \textcolor{green!50!black}{$\uparrow$} indicates improvement, \textcolor{red!70!black}{$\downarrow$} indicates a decline, and black values denote ties.}
\OnlyACL{
    \vspace{-10pt}
}
\label{tab:zero-shot}
\end{table*}

%% file: Sections/6_finetuning.tex
\subsection{Retrieval-Augmented Fine-Tuning}
\label{sec:experiments:finetuning}

Based on the results obtained with inference-only \oursraw, we study fine-tuning with \oursraw. Our central question is whether retrieval should be used only at inference time or also during fine-tuning on retrieval-augmented data. Recall that our aim to improve performance on unseen tasks within the same environment without degrading seen-task performance.

\subsubsection{Experimental Setup}
\label{sec:experiments:finetuning-setup}

Based on the findings from Section~\ref{sec:experiments:zero_shot}, we adopt a practical shared \ours setting for the fine-tuning experiments: (i) trajectories stored in JSON format, (ii) static retrieval, and (iii) $K=2$. In both environments, we fine-tune four backbone models (Ministral 3-8B, Gemma 3-4B, Qwen 2.5-7B, Qwen 2.5-7B-1M) on the train split of \textit{easy} tasks. We then report inference results on the held-out split of both \textit{easy} and \textit{hard} tasks, in order to assess both in-distribution and out-of-distribution performance. We consider two methods:%

\ACLorOther{
\noindent\textbf{\textcircled{\small A} LoRA}: we employ SFT as described in Sec.~\ref{sec:experimental_setup}, and (optionally) apply \ours during inference;\\
\noindent\textbf{\textcircled{\small B} \oursraw-LoRA}: retrieval-augmented fine-tuning where the \ours memory block is added to each training context using the inference-time retrieval pipeline (Sec.~\ref{sec:method}). We hypothesize this encourages model to learn to use retrieved trajectories instead of memorizing training targets, increasing its reliance on retrieved context at inference.
}{
\begin{itemize}
    \item \textbf{LoRA}: we conduct supervised fine-tuning as described in Section~\ref{sec:experimental_setup}, and (optionally) apply \ours during inference;
    \item \textbf{\oursraw-LoRA}: retrieval-augmented fine-tuning in which the \ours memory block is added to each training context via the same retrieval pipeline used at inference time (Section~\ref{sec:method}). We hypothesize that this encourages the model to learn to solve tasks by leveraging retrieved trajectories rather than memorizing training targets, thereby increasing its reliance on retrieved context at inference time.
\end{itemize}
}

\paragraphshort{Generalization Dynamics and Epoch Selection.}
For LoRA and \oursraw-LoRA, we observed rollout success often continued improving after validation loss started increasing, particularly on held-out task groups. We therefore avoid validation-loss-based early stopping and evaluate longer runs. We report this preliminary findings on LLM-agents training dynamics in Appendix~\ref{app:longer-training}. Considering this, for our comparisons, we use fixed training budgets of 10 epochs for ALFWorld and 30 for ScienceWorld, chosen to balance performance, training cost, and robustness across setups. We encourage practitioners to consider even longer runs when maximizing out-of-distribution performance is the primary goal.

\subsubsection{Retrieval-Augmented Fine-tuning Enables Robust Task Generalization}
\label{subsec:finetuning:results}

\input{tables/fine_tuning_results}

\ACLorOther{
    We report fine-tuning results in Table~\ref{tab:ft_results}. We assess hard-task gains with paired task-level confidence analyses; the testing protocol and full results are reported in Appendix~\ref{app:statistical-reliability}.
}{
    In Table~\ref{tab:ft_results}, we report fine-tuning results for four backbone models: Ministral 3-8B, Gemma 3-4B, Qwen 2.5-7B, Qwen 2.5-7B-1M. We assess hard-task gains with paired task-level confidence analyses; the testing protocol and full results are reported in Appendix~\ref{app:statistical-reliability}.
}

\paragraphshort{In-distribution tasks.} On easy tasks seen during training, all fine-tuning methods outperform training-free \ours across models on both ALFWorld and ScienceWorld. Retrieval-augmented training (\oursraw-LoRA) consistently improves performance on ScienceWorld, but sometimes yields a minor decrease on ALFWorld. Note that for in-distribution tasks, \oursraw-LoRA does not require additional held-out trajectories, since we use the training set as experience index.

\paragraphshort{Out-of-distribution task generalization.} 
On hard tasks, which the agent has never encountered during training, the performance of LoRA fine-tuned models collapses, falling below training-free \ours in most cases. These sharp easy-to-hard drops are consistent with SFT over-specializing the policy to the easy-task training distribution, such that introducing relevant experience only at inference may be insufficient. Applying \ours at inference time on the LoRA-trained model often improves over bare LoRA, but the paired evidence is weaker in some settings. In contrast, \oursraw-LoRA is reliably better than bare LoRA on ALFWorld hard for all backbones and on ScienceWorld hard for all backbones except Gemma 3-4B.

\upshot{Bringing retrieval augmentation already into training enables the model to natively handle retrieved trajectories in-context and generalize better to new out-of-distribution tasks. This comes with higher training cost due to the longer context, but preserves in-distribution performance in the same high-success range. These findings are consistent across benchmarks and models.}%

%% file: tables/fine_tuning_results.tex
\providecommand{\fttrend}{\textsuperscript{\(\dagger\)}}
\providecommand{\ftone}{\textsuperscript{*}}
\providecommand{\fttwo}{\textsuperscript{**}}
\providecommand{\ftthree}{\textsuperscript{***}}

\newcommand{\FTLlamaThreeOneRowA}{\ours & 35.6 & 34.4 & 23.5 & 20.3 \\}
\newcommand{\FTLlamaThreeOneRowB}{LoRA (no \oursraw) & 97.3 & 21.3 & 55.3 & 7.8 \\}
\newcommand{\FTLlamaThreeOneRowC}{LoRA (with \oursraw) & 84.9 & 44.3 & 41.2 & 21.9 \\}
\newcommand{\FTLlamaThreeOneRowD}{\oursraw-LoRA & 94.5 & 80.3 & 55.3 & 21.9 \\}

\newcommand{\FTLlamaThreeTwoRowA}{\ours & 0.0 & 0.0 & 0.0 & 0.0 \\}
\newcommand{\FTLlamaThreeTwoRowB}{LoRA (no \oursraw) & 98.6 & 0.0 & 48.3 & 4.7 \\}
\newcommand{\FTLlamaThreeTwoRowC}{LoRA (with \oursraw) & 43.8 & 8.2 & 21.2 & 4.7 \\}
\newcommand{\FTLlamaThreeTwoRowD}{\oursraw-LoRA & 76.7 & 34.4 & 52.3 & 10.9 \\}

\newcommand{\FTMinistralRowA}{\ours & 54.8 & 47.5\fttrend & 43.5 & 18.8 \\}
\newcommand{\FTMinistralRowB}{LoRA (no \oursraw) & 98.6 & 34.4 & 38.8 & 15.6 \\}
\newcommand{\FTMinistralRowC}{LoRA (with \oursraw) & 97.3 & \textbf{67.2}\ftthree & 54.1 & 28.1\fttrend \\}
\newcommand{\FTMinistralRowD}{\oursraw-LoRA & 97.3 & \textbf{88.5}\ftthree & 58.8 & \textbf{42.2}\fttwo \\}

\newcommand{\FTGemmaRowA}{\ours & 20.6 & 4.9 & 10.6 & 6.3 \\}
\newcommand{\FTGemmaRowB}{LoRA (no \oursraw) & 61.6 & 1.6 & 8.2 & 6.3 \\}
\newcommand{\FTGemmaRowC}{LoRA (with \oursraw) & 57.5 & 3.3 & 15.3 & 7.8 \\}
\newcommand{\FTGemmaRowD}{\oursraw-LoRA & 86.3 & \textbf{73.8}\ftthree & 31.8 & 6.2 \\}

\newcommand{\FTQwenRowA}{\ours & 81.6 & \textbf{70.5}\ftthree & 16.5 & 6.2 \\}
\newcommand{\FTQwenRowB}{LoRA (no \oursraw) & 86.3 & 21.3 & 24.7 & 7.8 \\}
\newcommand{\FTQwenRowC}{LoRA (with \oursraw) & 89.0 & \textbf{70.5}\ftthree & 25.9 & \textbf{23.4}\ftone \\}
\newcommand{\FTQwenRowD}{\oursraw-LoRA & 84.9 & \textbf{90.2}\ftthree & 38.8 & \textbf{29.7}\fttwo \\}

\newcommand{\FTQwenLongRowA}{\ours & 67.1 & \textbf{54.1}\ftthree & 20.0 & 15.6 \\}
\newcommand{\FTQwenLongRowB}{LoRA (no \oursraw) & 98.6 & 23.0 & 43.5 & 14.1 \\}
\newcommand{\FTQwenLongRowC}{LoRA (with \oursraw) & 82.2 & \textbf{68.9}\ftthree & 34.1 & 12.5 \\}
\newcommand{\FTQwenLongRowD}{\oursraw-LoRA & 97.3 & \textbf{91.8}\ftthree & 50.6 & \textbf{29.7}\ftone \\}

\newcommand{\FTBlockWide}[5]{%
    \multirow{4}{*}{#1} & #2
    & #3
    & #4
    & #5
}

\newcommand{\FTBlockACL}[5]{%
    \multicolumn{5}{l}{\textbf{\textit{#1}}} \\
    #2
    #3
    #4
    #5
}

\ACLorOther{
\begin{table}[t]
    \centering
    \newsavebox{\ftresultsaclbox}
    \sbox{\ftresultsaclbox}{%
    \begin{tabular}{lcccc}
    \toprule
    \multirow{4}{*}{Method} &
    \multicolumn{2}{c}{ALFWorld} &
    \multicolumn{2}{c}{ScienceWorld} \\
    \cmidrule(lr){2-3}\cmidrule(lr){4-5}
    & Easy & Hard & Easy & Hard \\
    & tasks & tasks & tasks & tasks \\
    & \makecell{(in-d)} & \makecell{(oo-d)} & \makecell{(in-d)} & \makecell{(oo-d)} \\
    \midrule
    \FTBlockACL{\quad Ministral 3-8B}{\FTMinistralRowA}{\FTMinistralRowB}{\FTMinistralRowC}{\FTMinistralRowD}
    \midrule
    \FTBlockACL{\quad Gemma 3-4B}{\FTGemmaRowA}{\FTGemmaRowB}{\FTGemmaRowC}{\FTGemmaRowD}
    \midrule
    \FTBlockACL{\quad Qwen 2.5-7B}{\FTQwenRowA}{\FTQwenRowB}{\FTQwenRowC}{\FTQwenRowD}
    \midrule
    \FTBlockACL{\quad Qwen 2.5-7B-1M}{\FTQwenLongRowA}{\FTQwenLongRowB}{\FTQwenLongRowC}{\FTQwenLongRowD}
    \bottomrule
    \end{tabular}%
    }
    \resizebox{0.95\linewidth}{!}{\usebox{\ftresultsaclbox}}
    \caption{\textbf{Task generalization through retrieval-augmented training for different model backbones.} We report success rates of ALFWorld and ScienceWorld after fine-tuning on a subset of tasks (easy tasks). For hard-task columns, bold marks improvements over the corresponding LoRA baseline that are significant under a paired McNemar test (\(p<0.05\)); superscripts denote \(^{\dagger}p<0.1\), \(^{*}p<0.05\), \(^{**}p<0.01\), and \(^{***}p<0.001\). Full paired bootstrap confidence intervals and McNemar tests are reported in Appendix~\ref{app:statistical-reliability}. All retrieval settings use matched index.
    }
    \OnlyACL{
        \vspace{-10pt}
    }
    \label{tab:ft_results}
\end{table}
}{
\begin{table*}[t]
    \centering
    \small
    \setlength{\tabcolsep}{6pt}
    \begin{tabular}{llcccc}
    \toprule
    \multirow{3}{*}{Backbone} &
    \multirow{3}{*}{Method} &
    \multicolumn{2}{c}{ALFWorld} &
    \multicolumn{2}{c}{ScienceWorld} \\
    \cmidrule(lr){3-4}\cmidrule(lr){5-6}
    & & Easy tasks & Hard tasks & Easy tasks & Hard tasks \\
    & & (in-d) & (oo-d) & (in-d) & (oo-d) \\
    \midrule
    \FTBlockWide{Ministral 3-8B}{\FTMinistralRowA}{\FTMinistralRowB}{\FTMinistralRowC}{\FTMinistralRowD}
    \midrule
    \FTBlockWide{Gemma 3-4B}{\FTGemmaRowA}{\FTGemmaRowB}{\FTGemmaRowC}{\FTGemmaRowD}
    \midrule
    \FTBlockWide{Qwen 2.5-7B}{\FTQwenRowA}{\FTQwenRowB}{\FTQwenRowC}{\FTQwenRowD}
    \midrule
    \FTBlockWide{Qwen 2.5-7B-1M}{\FTQwenLongRowA}{\FTQwenLongRowB}{\FTQwenLongRowC}{\FTQwenLongRowD}
    \bottomrule
    \end{tabular}%
    \caption{\textbf{Task generalization through retrieval-augmented training for different model backbones.} We report success rates of ALFWorld and ScienceWorld after fine-tuning on a subset of tasks (easy tasks). For hard-task columns, bold marks improvements over the corresponding LoRA baseline that are significant under a paired McNemar test (\(p<0.05\)); superscripts denote \(^{\dagger}p<0.1\), \(^{*}p<0.05\), \(^{**}p<0.01\), and \(^{***}p<0.001\). Full paired bootstrap confidence intervals and McNemar tests are reported in Appendix~\ref{app:statistical-reliability}. All retrieval settings use matched index.}
    \label{tab:ft_results}
\end{table*}
}

%% file: Sections/6.4_robustness_failures.tex
\subsubsection{Robustness to Imperfect Experience and Learning from Failures}
\label{sec:experiments:imperfect-experience}

\input{figures/robustness/robustness}

Our experiments so far use a matched index of scripted expert trajectories, a controlled but strong assumption. We therefore stress-test whether the gains of fine-tuned \oursraw-LoRA persist when available experience is sparse, mismatched, absent, or model-generated, and whether agents can reuse their own failed attempts. We evaluate these conditions on the same \ours variants from Sec.~\ref{subsec:finetuning:results}.

We use \textbf{\underline{Matched Expert}}, the full hard-task expert index, as our reference condition. We test more realistic variations: \textbf{\underline{Empty}} removes retrieval, \textbf{\underline{Mismatched Expert}} keeps only the easy-task expert index, and \textbf{\underline{Low-Coverage Expert}} contains only five trajectories per hard task type. To test whether expert experience is necessary, \textbf{\underline{Matched Strong-LLM}} replaces it with matched trajectories generated by stronger Qwen2.5-70B, generally performing worse than expert and containing more suboptimal solutions, representing a setting where a stronger teacher seeds experience for a smaller one. Finally, in \textbf{\underline{Self-Experience}}, we go one step further and use the agent's own first attempt (successful or failed)\footnote{For Ministral-3-8B, Self-Experience index contains 0/61 successful trajectories on ALFWorld and 2/64 on ScienceWorld; full index statistics are reported in Appendix~\ref{app:imperfect-experience:index}.} of No RAG as index for a second attempt, testing whether it can improve from its own experience without parameter updates. We test this condition both alone and together with each non-empty experience source. Results for Ministral 3-8B are shown in Figure~\ref{fig:robustness_stress_tests_unified}; full index-construction details and results for other models are provided in Appendix~\ref{app:imperfect-experience}.

\paragraphshort{Retrieval-aware fine-tuned model learns to use imperfect experience.}  \oursraw-LoRA is generally the strongest variant when experience is imperfect but still useful. It remains robust with \textbf{Low-Coverage Expert} and \textbf{Matched Strong-LLM} indices, despite the latter containing more suboptimal trajectories than the matched expert bank. On ALFWorld, the same holds for \textbf{Self-Experience} and \textbf{Mismatched Expert}; only with \textbf{Empty} index does it fall on par with LoRA + \oursraw. Together, these results are consistent with retrieval-aware fine-tuning helping the model identify and combine useful information from imperfect experience, rather than merely imitating retrieved demonstrations: for \oursraw-LoRA, relevance appears more important than having a large or near-perfect expert bank. The ScienceWorld exceptions may reflect factors observed earlier: a larger easy-to-hard gap with less reusable behaviors, harder retrieval among similar trajectories, and lower-quality trajectories overall. 

\paragraphshort{Towards Self-Improving Agents from Past Experience.} \oursraw-LoRA substantially benefits on ALFWorld when \textbf{Self-Experience} is the sole source of experience. This is particularly striking as all first attempts fail, suggesting that retrieval-aware fine-tuning can help the model revisit its mistakes and turn unsuccessful rollouts into useful experience. This ability also persists when the agent's own experience is combined with \textbf{Matched Expert} or \textbf{Matched Strong-LLM} memories. The same behavior does not emerge on ScienceWorld, where \textbf{Self-Experience} hurts performance, which may reflect the lower diversity of available training experience in ScienceWorld, along with the issues discussed above. Learning from failures therefore also appears to depend on the meta-learning setting, particularly on the diversity of cases seen during training.

\upshot{Retrieval-aware fine-tuning enables agents to use experience beyond large, scripted expert banks: useful memory can be sparse, model-generated, or even come from the agent's own attempts. Its effectiveness still depends on the experience containing relevant, reusable information, but these results point toward in-context self-improvement without parameter updates.}

%% file: figures/robustness/robustness.tex
\ACLorArxivorOther{
\begin{figure*}[t]
    \includegraphics[width=1.01\linewidth]{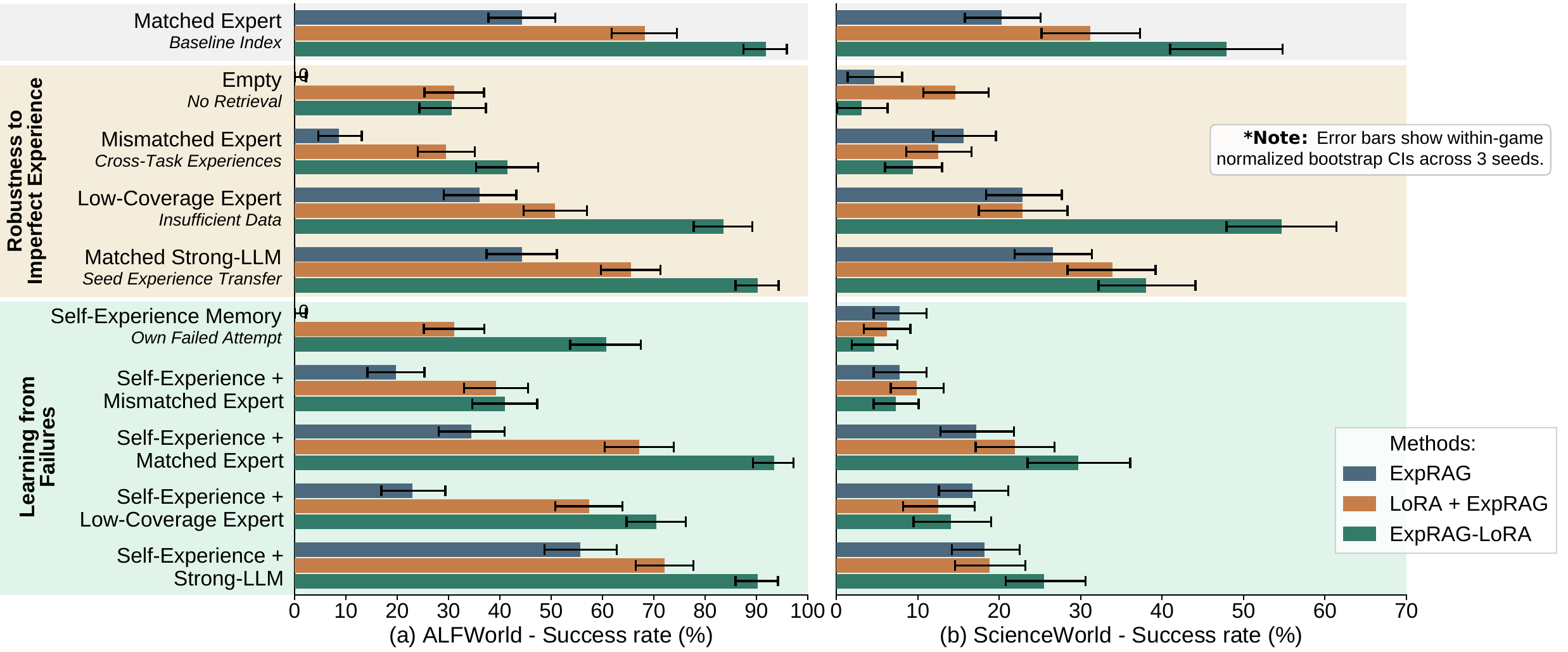}
    \vspace{-0.65cm}
    \caption{\textbf{Experience Robustness and Learning from failures.} Results on unseen \emph{hard} tasks with Ministral 3-8B. Error bars show 95\% within-game normalized bootstrap CIs across three seeds; full statistics are in Appendix~\ref{app:imperfect-experience}.\vspace{-0.25cm}}
    \label{fig:robustness_stress_tests_unified}
\end{figure*}
}{
\begin{figure*}[t]
    \includegraphics[width=1.0\linewidth]{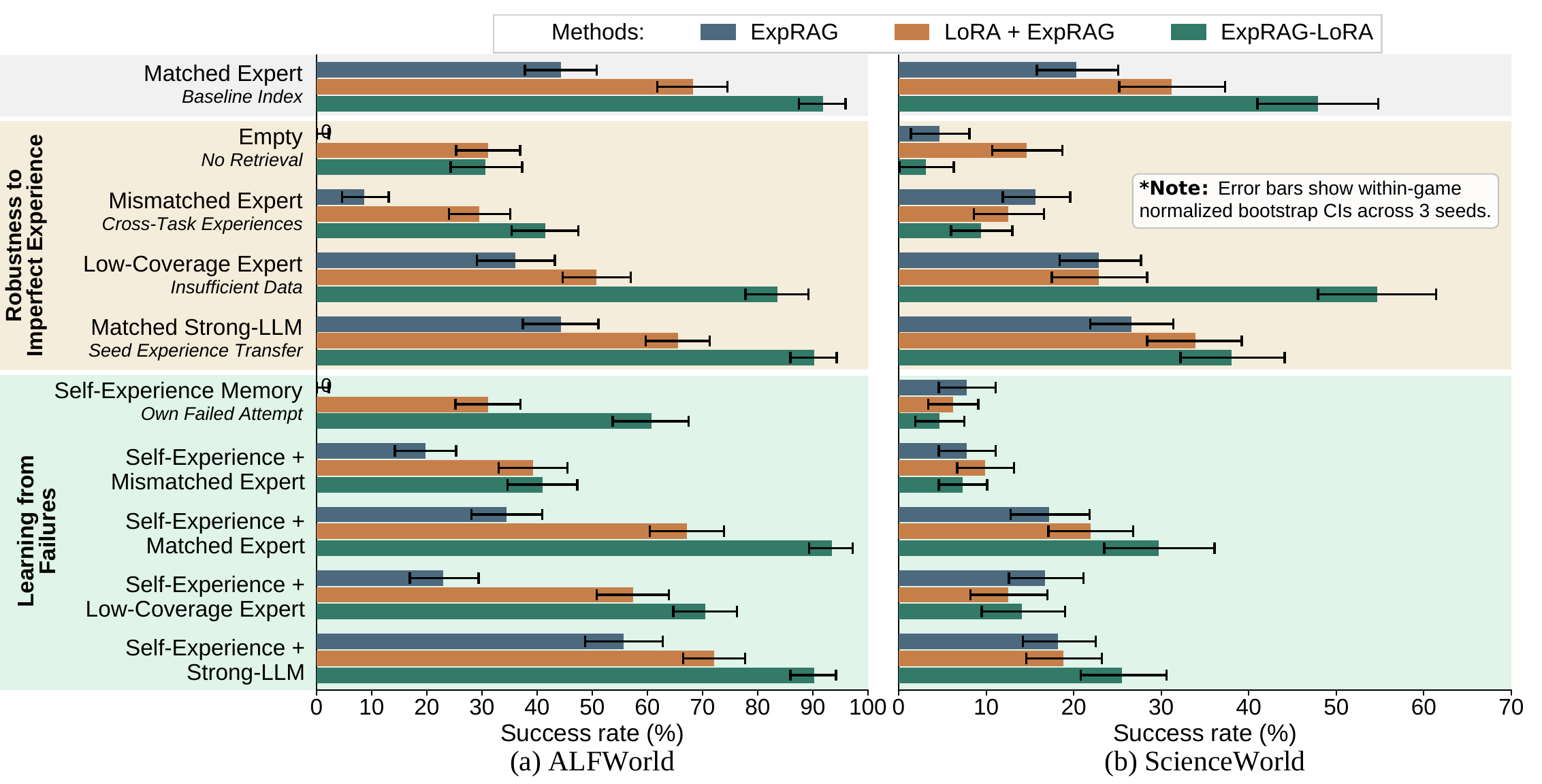}
    \caption{\textbf{Robustness to imperfect experience and learning from failures.}Success rates on unseen hard tasks with Ministral 3-8B across three seeds. The upper block stress-tests experience availability, relevance, coverage, and source relative to the \textbf{Matched Expert} index; the lower block tests reuse of the agent's own first attempt as experience for a second attempt, alone or combined with other memory sources and without parameter updates. Error bars show 95\% within-game normalized bootstrap CIs over aligned game--seed instances. Full details and additional results are reported in Appendix~\ref{app:imperfect-experience}.
    }
    \label{fig:robustness_stress_tests_unified}
\end{figure*}
}{
}

%% file: Sections/7_conclusion.tex
\section{Conclusion}

We studied episodic retrieval for LLM agents, from training-free baselines to retrieval-aware fine-tuning. Across ALFWorld and ScienceWorld, \ours yields large performance gains at inference-time, and our ablations establish a practical trajectory-retrieval recipe. Because standard held-out splits can remain close to the training distribution, we evaluate easy-to-hard task transfer: LoRA often collapses on unseen \emph{hard} tasks, whereas \oursraw-LoRA generally provides the strongest hard-task generalization while preserving strong seen-task performance. This advantage often persists with imperfect but useful experience; on ALFWorld, the agent can even reuse a failed attempt to improve a second rollout without test-time parameter updates. Overall, simple read-only episodic retrieval provides a strong foundation for agent memory. We argue that new, more complex memory architectures should be compared against strong fine-tuning and retrieval baselines under task-level distribution shifts and memory-quality stress tests before attributing gains to their added complexity.

%% file: Sections/8_limitations.tex
Our study has some limitations. 
First, our main experiments use scripted expert trajectories as a controlled memory source. Section~\ref{sec:experiments:imperfect-experience} stress-test this assumption with sparse, mismatched, stronger-LLM-generated, and self-generated experience, but does not study long-term accumulation of heterogeneous experience: Self-Experience reuses only the immediately preceding attempt. Future work should evaluate continually evolving memories over longer interaction horizons.
Second, performance remains dependent on the relevance and coverage of available experience: missing or poorly matched trajectories can degrade generalization, and learning from self-generated failures does not emerge consistently across environments. This suggests the need for more principled strategies for experience collection, selection, and retrieval.
Finally, our approach uses a fixed, read-only episodic memory. This isolates the policy-side ability to use retrieved experience, but leaves open how these findings extend to read--write memories with consolidation mechanisms (e.g., summarization, abstraction, and selective retention) and longer-horizon adaptation.

%% file: Sections/9_acknowledgements.tex
\section*{Acknowledgements}

We thank our colleagues Nadezhda Chirkova and Anastasiia Ivanova for their helpful discussions and valuable feedback on this work and manuscript. Thomas' research is partially supported by the French Ministry of Higher Education, Research and Innovation under CIFRE PhD Convention \textit{``Reconcile Efficiency and Interpretability in Structured Planning and Reasoning''} (no. \texttt{2025/0487}).

\paragraph{Use of AI Assistants.} AI tools were used during this research for manuscript language editing and polishing, assistance with editing software code, and creating and modifying scientific figures. All AI-generated outputs were reviewed and verified by the authors, who take full responsibility for the scientific content, experimental results, interpretations, and conclusions.

%% file: appendix/additional_related_work.tex
\section{Additional Related Work}
\label{app:additional-related-work}

Section~\ref{sec:related_work} focuses on the work most directly comparable to our experiments: agent fine-tuning, episodic memory, and retrieval of prior trajectories. Here, we place our setting in the broader literature on systems that improve from interaction. Meta-learning, online and continual learning, self-correction, memory-based adaptation, and self-improvement emerged as partly separate paradigms, but increasingly converge in LLM agents, where the same interaction trace may serve as immediate feedback, persistent context, training data, or a signal for modifying skills, tools, and the agent itself. We organize these connections by how experience changes the system: through contextual adaptation, parameter updates, memory evolution, or changes to the surrounding agent architecture.

\paragraphshort{Meta-learning and continual adaptation.}
Meta-learning optimizes models for rapid adaptation from new examples or trajectories, either through a learned initialization~\citep{MAML} or a policy that internalizes a learning procedure~\citep{wang2016learning}. Online and continual learning instead study adaptation over sequential experience, with continual learning emphasizing the retention of previously acquired capabilities under distribution change~\citep{,continual-learning-llm-survey}. Work combining these paradigms formalizes systems that both improve how they learn and adapt over time~\citep{metalearningmeetsonline24}. Recent LLM-agent roadmaps broaden this view beyond parameter updates to adaptation mediated by memory, context, and interaction~\citep{lifelonglearningllmssurvey26,self-evolving-survey}. Our setting is narrower: retrieval-aware training occurs offline, while adaptation to a new task occurs only through retrieved context.

\paragraphshort{From self-correction to evolving memory.}
A parallel line of work improves model behavior by using feedback on previous outputs. Self-correction methods iteratively critique and revise a current solution, but do not generally retain the resulting lessons across inputs \citep{selfrefine2023,decrim20204,CRITIC2024}. Other works connect self-correction and memory by retaining reflections, feedback, or distilled insights that can guide later trials \citep{ExpeL,reflexion2023}. More recent systems make this persistence central: Dynamic Cheatsheet \citep{suzgun-etal-2026-dynamic} accumulates reusable strategies at test time, ACE~\citep{ACE2026} treats context as an evolving playbook, and Evo-Memory~\citep{wei2025EVOmemory} evaluates memories that update across task sequences. This progression blurs the boundary between self-correction, episodic memory, and test-time learning: feedback is no longer used only to repair one output, but becomes reusable experience for future behavior.

\paragraphshort{Experience quality, provenance, and transfer.}
The usefulness of experience depends not only on its relevance, but also on its outcome, origin, and compatibility with the learner. As summarized in Section~\ref{sec:related_work}, methods such as ETO, NAT, and Early Experience turn failed or suboptimal interactions into supervision for further policy optimization~\citep{song2024ETO,NAT2025,EarlyExperience}. A complementary teacher--student line trains smaller agents on trajectories produced by stronger models~\citep{fireact2023,smartad26}, while~\citet{kim2026agentmemorydistillationempowering} instead converts successful teacher trajectories into structured external memory. Our stress tests isolate a different question: keeping the trained policy and inference pipeline fixed, can a retrieval-trained agent directly exploit raw trajectories from sparse expert banks, stronger teachers, or its own previous attempts, without additional test-time optimization or memory synthesis?

\paragraphshort{Toward self-evolving agents.}
Broader self-improvement work expands what components are allowed to change from experience. At the model level, SPIN iteratively updates an LLM using responses generated by previous versions of itself~\citep{SPIN24}. Agent-oriented systems further connect interaction to policy and memory evolution: RAGEN studies self-evolution through multi-turn reinforcement learning, while SkillRL jointly evolves a policy and a retrieved skill library~\citep{RAGEN2025,SkillRL2026}. Mem$^2$Evolve~\citep{mem2evolve2026} extends this process to reusable tools and expert assets, using accumulated experience to guide capability expansion and newly acquired capabilities to generate further experience. At the system level, the Darwin G\"odel Machine iteratively modifies and evaluates its own agent code, allowing the mechanisms supporting future improvement to change as well~\citep{zhang2026darwin,zhang2026hyperagents}. These approaches illustrate a progression from adapting behavior through context, to evolving memory and skills, and ultimately to modifying model parameters or the agent architecture itself. Our work isolates a prerequisite for these richer loops: whether a policy can first be trained to reliably interpret and reuse raw episodic experience.

\subsection{Terminology clarification: Retrieval-Augmented Agents vs.\ Agentic RAG}

The similarity in terminology can obscure an important conceptual distinction. Our setting is not an instance of \textit{Agentic RAG}. Following \citet{memoryageaiagents_survey2026}, we distinguish \textit{retrieval-augmented agents}, which retrieve \textit{experience} to improve decisions in an environment, from \textit{Agentic RAG}~\citep{agenticRAG2025}, where an agent orchestrates multi-step retrieval over external knowledge sources. Our work belongs to the former: the goal is policy generalization and robustness through episodic context reuse,  whereas the latter targets knowledge acquisition and synthesis for question answering.

%% file: appendix/implementation_details.tex
\section{Implementation details}
\label{app:implementation-details}

This appendix centralizes the implementation choices that are shared across experiments.

\subsection{Indexing and Retrieval experimental settings}
\label{app:implementation-details:retrieval}

In all conditions with retrieval ($K > 0$), we build a fixed retrieval index from expert trajectories in the environment training split (including failures unless stated otherwise). We embed item queries and index keys (in formatted text) using \texttt{Qwen/Qwen3-Embedding-0.6B} model\footnote{Available at:\url{https://huggingface.co/Qwen/Qwen3-Embedding-0.6B}}~\citep{qwen3embedding} and retrieve the top-$K$ nearest trajectories by dot-product similarity in embedding space (Section~\ref{sec:method}). We do not experiment with embedding models other than this one, as we found the retrieved trajectories with this model to be satisfactory, given the simplicity of the retrieval task in this scenario, and therefore keep it fixed throughout the paper. We use the \textit{Sentence Transformers} library\footnote{More information:~\url{https://sbert.net/}}~\citep{sentencetransformers} to compute the embeddings.

Trajectories are stored as raw chat-formatted data. Unless otherwise stated, when retrieved trajectories are inserted into the policy prompt, we format them as \textit{chat JSON}; this is the default used in the main-paper zero-shot and fine-tuning results. Appendix~\ref{subapp:trajectory_formatting} compares this default against \textit{agentic JSON}, \textit{compact JSON}, and \textit{textual} alternatives.

\subsection{Benchmark data statistics}
\label{app:implementation-details:data-statistics}

In Section~\ref{sec:experiments:dataset-and-benchmarks} we introduced our benchmark datasets and how we split them into hard/easy tasks. Here we provide more details on the data statistics. Table \ref{tab:hard_easy_splits} reports how we split training and validation datasets into hard/easy tasks, and what tasks belong to each category. To evaluate models faster and avoid task imbalance, we subsample the ScienceWorld test set, and keep only 5 variations per task. We provide the list of variations in Table~\ref{tab:task_variations}. We use this subset in all tables except Table~\ref{tab:our-baselines}, where we compare with related work.
\input{tables/appendix/data_splits_details}
\input{tables/appendix/scienceworld_subset}

\subsection{Memory block formatting}
\label{app:implementation-details:memory-format}

When inserting retrieved trajectories into the agent context, we prepend a memory block that separates successful from unsuccessful demonstrations. Concretely, we use the following template:
\begin{quote}
\ttfamily\small
These are examples of successful trajectories: \{\dots\}. These are examples of unsuccessful trajectories: \{\dots\}.
\end{quote}

Within this memory block, the retrieved trajectories themselves are represented in \textit{chat JSON} format by default (Figure~\ref{fig:traj_format}), unless a formatting ablation explicitly states otherwise. %

\subsection{Fine-tuning implementation}
\label{app:implementation-details:finetuning}

We implement supervised LoRA fine-tuning with TorchTune~\citepAliasYear{torchtune}. Training trajectories are formatted as multi-turn chats using each backbone's default chat template, and we compute the cross-entropy loss on assistant tokens only. We train all models on a single 80~GB NVIDIA A100 GPU.

Despite using TorchTune for training, we run inference with Hugging Face Transformers~\citep{HFTransformers}, which we found to be more efficient.

\subsection{Model Backbones}
\label{app:implementation-details:models}

\paragraphshort{Models.}
In all experiments, we use only instruction-tuned base LLMs as agent policies. In the main text, we refer to each model by a short name. The corresponding official checkpoint names on the Hugging Face Hub\footnote{Base URL: \texttt{https://huggingface.co/<model\_name>} ; replace \texttt{<model\_name>} with the checkpoint official identifier to obtain the model page.} are:

\begin{itemize}
    \item \textbf{Ministral 3-8B}~\citep{ministral3}: \texttt{mistralai/Ministral-3-8B-Instruct-2512-BF16}
    \item \textbf{Gemma 3-4B}~\citep{gemma3report}: \texttt{google/gemma-3-4b-it}
    \item \textbf{Qwen 2.5-7B-1M}~\citep{qwen25-1mcontext}: \texttt{Qwen/Qwen2.5-7B-Instruct-1M}
    \item \textbf{Qwen 2.5-7B}~\citep{qwen25report}: \texttt{Qwen/Qwen2.5-7B-Instruct}
    \item \textbf{Qwen 2.5-3B}~\citep{qwen25report}: \texttt{Qwen/Qwen2.5-3B-Instruct}
    \item \textbf{Llama 3.1-8B}~\citep{llama3herdmodels2024}: \texttt{meta-llama/Llama-3.1-8B-Instruct}
    \item \textbf{Llama 3.2-3B}~\citep{meta_llama32}: \texttt{meta-llama/Llama-3.2-3B-Instruct}
    \item \textbf{Llama 3.2-1B}~\citep{meta_llama32}: \texttt{meta-llama/Llama-3.2-1B-Instruct}
\end{itemize}

We include both Qwen 2.5-7B and Qwen 2.5-7B-1M at matched parameter scales. Qwen2.5-7B-1M is not merely a longer-context setting of Qwen 2.5-7B: it is obtained via additional long-context adaptation (continued training with progressive context-length expansion and associated positional configuration changes), and is post-trained for long-input instruction following~\citep{qwen25-1mcontext}. Consequently, comparisons between Qwen 2.5-7B and Qwen 2.5-7B-1M reflect both increased context capacity and the effects of long-context-specific training and alignment.

\subsection{Hyperparameters}
\label{app:implementation-details:hyperparameters}

Table~\ref{tab:hyperparameters} reports the hyperparameters used for our method. The values below are the defaults used in all experiments unless stated otherwise in the main text or in the experimental setup; when a setting is specified elsewhere (e.g., for a particular benchmark or ablation), that specification overrides the corresponding entry in this table.

\input{tables/appendix/hyperparameters}

In addition to the shared optimization and decoding settings above, we set the maximum number of actions that the agent is allowed to perform, using \texttt{max\_steps\_per\_task}. For ALFWorld, we follow the environment default (\texttt{max\_steps\_per\_task = 50}). For ScienceWorld, we found that some tasks are not solvable within 50 steps. On the other hand, setting a uniformly high budget would make evaluation inefficient for easy tasks that the model cannot solve, where the agent may get stuck in loops. Based on the number of steps taken by the rule-based expert to solve each task for each task group, we therefore define a different \texttt{max\_steps\_per\_task} value for each ScienceWorld task category. We report these values in Table~\ref{tab:scienceworld-max-steps}. If the model fails to solve the task within the budget, we consider the episode as a failure.

\input{tables/appendix/scienceworld_max_steps}

%% file: tables/appendix/data_splits_details.tex
\begin{table*}[!h]
\small
\centering
\setlength{\tabcolsep}{4pt}
\renewcommand{\arraystretch}{1.1}
\resizebox{\textwidth}{!}{
\begin{tabular}{l p{0.11\linewidth} p{0.29\linewidth} | p{0.11\linewidth} p{0.35\linewidth}}
\toprule
& \multicolumn{2}{c}{ALFWorld} & \multicolumn{2}{c}{ScienceWorld} \\
\cmidrule(lr){2-3}\cmidrule(l){4-5}
Split & Samples & Tasks (\textit{task-types}) & Samples & Tasks \\
\midrule
easy
& train=1748, test=73
& \makecell[tl]{look\_at\_obj\_in\_light\\pick\_clean\_then\_place\_in\_recep\\pick\_and\_place\_simple}
& train=2335, test=1183
& \makecell[tl]{find-plant\\freeze\\inclined-plane-friction-unnamed-surfaces\\lifespan-longest-lived\\lifespan-longest-lived-then-shortest-lived\\inclined-plane-friction-named-surfaces\\boil\\change-the-state-of-matter-of\\inclined-plane-determine-angle\\measure-melting-point-known-substance\\measure-melting-point-\\unknown-substance\\use-thermometer\\find-non-living-thing\\melt\\find-animal\\lifespan-shortest-lived\\find-living-thing} \\
\midrule
hard
& train=1805, test=61
& \makecell[tl]{pick\_cool\_then\_place\_in\_recep\\pick\_heat\_then\_place\_in\_recep\\pick\_two\_obj\_and\_place}
& train=1254, test=636
& \makecell[tl]{chemistry-mix-paint-secondary-color\\test-conductivity\\power-component-renewable-vs-\\nonrenewable-energy\\chemistry-mix-paint-tertiary-color\\identify-life-stages-1\\identify-life-stages-2\\test-conductivity-of-unknown-substances\\grow-fruit\\mendelian-genetics-known-plant\\power-component\\grow-plant\\mendelian-genetics-unknown-plant\\chemistry-mix} \\
\bottomrule
\end{tabular}%
}
\caption{Details about the hard/easy task splits: number of training and test samples, and task names assigned to each category. Samples from \textit{training tasks} are used to build the index for the corresponding split (hard/easy), while easy training tasks are used for \ours fine-tuning.}
\label{tab:hard_easy_splits}
\end{table*}

%% file: tables/appendix/scienceworld_subset.tex
\begin{table*}[t]
\centering
\small
\setlength{\tabcolsep}{4pt}
\renewcommand{\arraystretch}{1.1}
\rowcolors{2}{gray!8}{white}
\begin{tabular}{@{}p{0.5\linewidth}p{0.32\linewidth}@{}}
\toprule
Task & Task variations \\
\midrule
boil & 29, 27, 25, 23, 21 \\
change-the-state-of-matter-of & 29, 23, 28, 27, 26 \\
chemistry-mix & 26, 31, 24, 29, 25 \\
chemistry-mix-paint-secondary-color & 34, 27, 31, 33, 30 \\
chemistry-mix-paint-tertiary-color & 30, 33, 35, 28, 29 \\
find-animal & 248, 260, 294, 271, 299 \\
find-living-thing & 257, 256, 277, 266, 268 \\
find-non-living-thing & 280, 261, 233, 299, 253 \\
find-plant & 251, 231, 227, 254, 236 \\
freeze & 26, 23, 28, 21, 29 \\
grow-fruit & 123, 114, 108, 105, 96 \\
grow-plant & 95, 94, 100, 112, 116 \\
identify-life-stages-1 & 12, 9, 11, 13, 10 \\
identify-life-stages-2 & 6, 7, 8, 9 \\
inclined-plane-determine-angle & 151, 147, 150, 152, 148 \\
inclined-plane-friction-named-surfaces & 1356, 1327, 1322, 1366, 1360 \\
inclined-plane-friction-unnamed-surfaces & 146, 158, 145, 157, 156 \\
lifespan-longest-lived & 112, 106, 107, 111, 93 \\
lifespan-longest-lived-then-shortest-lived & 110, 114, 99, 101, 115 \\
lifespan-shortest-lived & 124, 94, 95, 121, 112 \\
measure-melting-point-known-substance & 355, 400, 422, 413, 403 \\
measure-melting-point-unknown-substance & 266, 263, 243, 281, 265 \\
melt & 27, 28, 29, 26, 23 \\
mendelian-genetics-known-plant & 94, 105, 100, 93, 91 \\
mendelian-genetics-unknown-plant & 413, 429, 406, 467, 442 \\
power-component & 19, 17, 18, 15, 16 \\
power-component-renewable-vs-nonrenewable-energy & 19, 16, 17, 15, 18 \\
test-conductivity & 858, 726, 839, 878, 844 \\
test-conductivity-of-unknown-substances & 538, 475, 494, 464, 532 \\
use-thermometer & 534, 476, 450, 473, 510 \\
\bottomrule
\end{tabular}
\rowcolors{2}{}{}
\caption{Task variations used in our subsampled ScienceWorld test set.}
\label{tab:task_variations}
\end{table*}

%% file: tables/appendix/hyperparameters.tex
\begin{table}[!htbp]
    \centering
    \small
    \setlength{\tabcolsep}{4pt}
    \renewcommand{\arraystretch}{1.1}
    \begin{tabular}{l c}
    \toprule
    \multicolumn{2}{c}{Hyperparameters} \\
    \midrule
    optimizer & PagedAdamW8bit \\
    \addlinespace
    learning rate & 5e-5 \\
    \addlinespace
    weight decay & 0.0 \\
    \addlinespace
    lr scheduler & constant \\
    \addlinespace
    LoRA target modules & \makecell[c]{\texttt{q\_proj}, \texttt{v\_proj},\\ \texttt{k\_proj}, \texttt{output\_proj}} \\
    \addlinespace
    LoRA rank & 8 \\
    \addlinespace
    LoRA $\alpha$ & 16 \\
    \addlinespace
    LoRA dropout & 0.1 \\
    \addlinespace
    dtype & bf16 \\
    \addlinespace
    decoding temperature & 0.0 (greedy) \\
    seed & 2025 \\
    \bottomrule
    \end{tabular}
    \caption{All the hyperparameters used for our method. Values are shared across models unless specified otherwise in the text.}
    \label{tab:hyperparameters}
\end{table}

%% file: tables/appendix/scienceworld_max_steps.tex
\begin{table}[!htbp]
\centering
\small
\setlength{\tabcolsep}{4pt}
\renewcommand{\arraystretch}{1.1}
\begin{tabular}{>{\centering\arraybackslash}m{0.10\linewidth} p{0.80\linewidth}}
\toprule
Max Steps & ScienceWorld tasks \\
\midrule
\rowcolor{gray!8}
150 & mendelian-genetics-known-plant; mendelian-genetics-unknown-plant \\
120 & boil \\
\rowcolor{gray!8}
90 & freeze \\
80 & grow-fruit \\
\multirow{2}{*}{\cellcolor{gray!8}70} & \cellcolor{gray!8}change-the-state-of-matter-of; inclined-plane-determine-angle; \\
& \cellcolor{gray!8}inclined-plane-friction-unnamed-surfaces; melt \\
\rowcolor{gray!8}
\multirow{8}{*}{50} & chemistry-mix; chemistry-mix-paint-secondary-color; chemistry-mix-paint-tertiary-color; \\
& find-animal; find-living-thing; find-non-living-thing; find-plant; grow-plant; \\
& identify-life-stages-1; identify-life-stages-2; inclined-plane-friction-named-surfaces; \\
& lifespan-longest-lived; lifespan-longest-lived-then-shortest-lived; lifespan-shortest-lived; \\
& measure-melting-point-known-substance; measure-melting-point-unknown-substance; \\
& power-component; power-component-renewable-vs-nonrenewable-energy; \\
& test-conductivity; test-conductivity-of-unknown-substances; \\
& use-thermometer \\
\bottomrule
\end{tabular}
\caption{ScienceWorld task-dependent rollout budgets used in evaluation (\texttt{max\_steps\_per\_task}).}
\label{tab:scienceworld-max-steps}
\end{table}

%% file: appendix/zero_shot_continuation.tex
\section{Extra experiments on \ours without Training}
\label{app:zero-shot-continuation}

\subsection{Main Results Stability and Reliability Across Seeds}
\label{subapp:random_seeds}

Tables~\ref{tab:zero-shot-ministral-std} and \ref{tab:zero-shot-ministral-std-summary} show the variability of the main results for Ministral~3-8B across three seeds (\texttt{2025}, \texttt{2026}, \texttt{2027}). These are the same results as in Table~\ref{tab:zero-shot}, with the standard deviations reported. Although decoding is greedy, the seed still affects mainly: (i) sampled environment instance (e.g., object placement), (ii) retrieval tie-breaking\footnote{Ties happen when multiple trajectories are retrieved with same score, which is recurrent, particularly in the static setting with task description as query.}, and therefore (iii) the subsequent interaction history seen by dynamic retrieval. Overall, the results are reasonably stable: across all configurations, the median/max standard deviation is \(1.58/6.20\) points on ALFWorld and \(1.83/4.45\) on ScienceWorld. We do not observe a single universally unstable regime. Instead, variability concentrates in harder or more retrieval-sensitive settings, while all-task averages are more stable (median std \(1.56\) on ALFWorld and \(1.69\) on ScienceWorld). Dynamic retrieval is only slightly less stable than static retrieval on ScienceWorld (median/max \(2.00/4.45\) vs.\ \(1.80/3.93\)), whereas on ALFWorld both have the same median variability (\(1.58\)). Importantly, the main best-performing configurations from Section~\ref{sec:experiments:zero_shot} are not unusually unstable: static retrieval with the \texttt{all} index at \(K{=}4\) on ALFWorld reaches \(64.18 \pm 1.97\), and dynamic retrieval with the \texttt{all} index at \(K{=}4\) on ScienceWorld reaches \(35.24 \pm 1.69\). Thus, the seed analysis supports the reliability of the main conclusions: the qualitative ranking of retrieval settings is robust, and the larger deviations mostly appear in specialized split/index combinations rather than in the central all-task findings.

\input{tables/zero_shot_appendix/zero_shot_ministral_std}
\input{tables/zero_shot_appendix/zeroshot_std_extraanalysis.tex}

\subsection{Validating Main Results across Different Models}
\label{subapp:extra_models}
Tables~\ref{tab:zero-shot-llama}, \ref{tab:zero-shot-llama32-3b}, \ref{tab:zero-shot-gemma3-4b}, \ref{tab:zero-shot-qwen25-3b}, \ref{tab:zero-shot-qwen25-7b}, and~\ref{tab:zero-shot-qwen7b1m} report inference-only \ours results for additional backbones. Overall, they support the same main conclusion as Table~\ref{tab:zero-shot} for Ministral~3-8B: retrieval is already highly useful without any training, although the magnitude of the gain depends strongly on the backbone.

\paragraphshort{Retrieval consistently improves over zero-shot.}
Retrieval consistently improves over \textsc{No RAG} at the best setting for every model. In Section~\ref{sec:experiments:zero_shot}, Ministral~3-8B improves from \(4.48/10.40\) to \(64.18/35.24\) ALFWorld/ScienceWorld all-task success. In the models reported here in this appendix, the best retrieval setting improves from \(7.46/7.38\) to \(47.76/27.52\) for Llama~3.1~8B, from \(0.75/2.01\) to \(12.69/19.46\) for Llama~3.2~3B, from \(0.75/2.01\) to \(20.90/11.41\) for Gemma~3~4B, from \(5.22/2.68\) to \(25.37/8.05\) for Qwen~2.5~3B, from \(29.85/2.01\) to \(83.58/12.75\) for Qwen~2.5~7B, and from \(5.22/3.36\) to \(81.34/27.52\) for Qwen~2.5~7B~1M. This matches the main-paper pattern: retrieved trajectories provide strong training-free gains from a frozen policy alone. Paired confidence intervals and McNemar tests for the hard-task comparisons against \textsc{No RAG} are reported in Table~\ref{tab:zero-shot-statistical-reliability} in Appendix~\ref{subapp:zero-shot-statistical-reliability}.

\paragraphshort{Larger top-\(k\) is useful, but more backbone-dependent.}
Larger top-\(k\) remains useful overall, but the pattern is more backbone-dependent. In Section~\ref{sec:experiments:zero_shot}, for Ministral~3-8B, ScienceWorld largely saturates after \(K{=}2\), whereas the appendix backbones usually achieve their best all-task scores at \(K{=}4\), especially on ALFWorld. At the same time, the gains are not monotonic in every split or index setting: smaller backbones such as Gemma~3~4B and Qwen~2.5~3B still show noticeable regressions when additional retrieved trajectories are weakly matched or noisy. This remains consistent with the main-text interpretation that larger memory helps when it adds complementary evidence, but can hurt when it introduces distractors.

\paragraphshort{Dynamic retrieval is the least stable factor across backbones.}
Dynamic retrieval can help substantially in selected settings, for example Qwen~2.5~3B with the \texttt{all} index at \(K{=}2\) on ALFWorld all tasks (\(20.90\%\rightarrow24.63\%\)) or Qwen~2.5~7B~1M with the \texttt{all} index at \(K{=}4\) on ScienceWorld all tasks (\(23.49\%\rightarrow27.52\%\)). However, it also produces some of the largest drops, such as Qwen~2.5~7B with the \texttt{all} index at \(K{=}2\) on ALFWorld all tasks (\(82.84\%\rightarrow73.13\%\)). This mirrors the Ministral~3-8B results: re-retrieval can increase relevance, but it is much less predictable because it repeatedly refreshes the prompt with partial-trajectory matches.

\paragraphshort{Index specialization remains broadly coherent.}
Easy-index rows tend to help easy splits more, hard-index rows tend to help hard splits more, and the mixed \texttt{all} index is often the strongest overall compromise. The same benchmark asymmetry as in the main paper also appears here: ALFWorld benefits more reliably from stronger retrieval and larger memory, whereas ScienceWorld remains more sensitive to index mismatch and retrieval noise, especially for smaller models.

\paragraph{Model size changes the magnitude of these effects.} Among the smaller backbones, Llama~3.2~3B (Table~\ref{tab:zero-shot-llama32-3b}), Gemma~3~4B (Table~\ref{tab:zero-shot-gemma3-4b}), and Qwen~2.5~3B (Table~\ref{tab:zero-shot-qwen25-3b}) all benefit from retrieval but remain substantially weaker and noisier than Ministral~3-8B, Llama~3.1~8B, and Qwen~2.5~7B. Within the Qwen family, scaling from 3B to 7B sharply improves both baseline competence and retrieval gains, especially on ALFWorld, where the best all-task score rises from \(25.37\%\) to \(83.58\%\).

\input{tables/zero_shot_appendix/zeroshot_gemma3_4b.tex}
\input{tables/zero_shot_appendix/zeroshot_qwen25_3b.tex}
\input{tables/zero_shot_appendix/zeroshot_qwen25_7b.tex}
\input{tables/zero_shot_appendix/zeroshot_qwen7b1m.tex}
\input{tables/zero_shot_appendix/compare_qwens_7b.tex}
\input{tables/zero_shot_appendix/zeroshot_llama31_8b.tex}
\input{tables/zero_shot_appendix/zeroshot_llama32_3b.tex}

\subsection{Long-context Qwen Supports Larger Top-\(K\) Retrieval.}
\label{subapp:long_context_qwen}

Tables~\ref{tab:zero-shot-qwen25-7b}, \ref{tab:zero-shot-qwen7b1m}, and \ref{tab:compare-qwens-7b} together provide a more nuanced picture than a simple ``longer context is better'' summary. The two Qwen variants share similar scale and architecture, but their behavior differs enough that the interpretation depends strongly on the benchmark.

The most striking difference between Qwen~2.5~7B (Table~\ref{tab:zero-shot-qwen25-7b}) and Qwen~2.5~7B~1M (Table~\ref{tab:zero-shot-qwen7b1m}) appears already in the \textsc{No RAG} baseline on ALFWorld: \(29.85\%\) all-task success for the standard 7B model versus only \(5.22\%\) for the 1M variant, with similarly large gaps on easy and hard tasks. This makes the ALFWorld comparison intrinsically confounded. Given the unusually strong zero-shot performance of Qwen~2.5~7B on this benchmark, we suspect that the standard 7B model may have been exposed during training to ALFWorld tasks or very close variants. We cannot verify this directly, but the two models, despite similar size and architecture, need not share the same pre-training or post-training data. For that reason, we do not treat Qwen~2.5~7B as a main backbone in the paper, and we interpret its ALFWorld advantage with caution.

This caution is visible in Table~\ref{tab:compare-qwens-7b}. On ALFWorld, the left half of the table is dominated by red values because the standard 7B model starts from a much stronger prior. Those absolute differences are therefore not directly comparable to the ScienceWorld side. What remains informative on ALFWorld is mostly the direction of the trend: retrieval often reduces the large initial gap, and some cells even flip from red to green when \(K\) becomes large, especially in the \(K{=}4\) block. For example, static retrieval with the \texttt{all} index moves from a \(-24.63\) all-task gap in \textsc{No RAG} to \(-5.22\) at \(K{=}1\), \(-20.15\) at \(K{=}2\), and then \(-2.24\) at \(K{=}4\), while the hard-index \(K{=}4\) rows contain several green entries on easy and all tasks. We therefore read the ALFWorld comparison mainly as evidence that retrieval can partially close the initial gap, not as a clean estimate of which model is intrinsically better.

ScienceWorld is more comparable because both models are weak in the pure zero-shot setting (\(2.01\%\) for Qwen~2.5-7B and \(3.36\%\) for Qwen~2.5-7B-1M on all tasks). In that benchmark, Table~\ref{tab:compare-qwens-7b} becomes progressively greener as \(K\) increases, especially in the lower \(K{=}4\) block. This is consistent with the individual model tables: the bold best cells in Table~\ref{tab:zero-shot-qwen7b1m} are concentrated at \(K{=}4\), whereas Table~\ref{tab:zero-shot-qwen25-7b} often peaks earlier on ScienceWorld, with several bold entries already at \(K{=}1\) or \(K{=}2\). The all-task gap for the \texttt{all} index illustrates this pattern clearly: under static retrieval it moves from \(-0.67\) at \(K{=}1\) to \(+5.37\) at \(K{=}2\) and \(+12.08\) at \(K{=}4\); under dynamic retrieval it moves from \(-2.69\) to \(+7.39\) and then \(+14.77\). The same pattern appears on harder ScienceWorld settings, where many of the largest positive differences are bright green in the comparison table, such as \(+23.43\) for static \texttt{all} / hard tasks at \(K{=}4\) and \(+28.12\) for dynamic \texttt{easy} / hard tasks at \(K{=}4\).

\paragraphshort{Long-context support is one important retrieval bottleneck, but not the only one.}
Taken together, these results are consistent with the view that context-handling capacity is an important bottleneck for scaling retrieval to larger top-\(K\), especially when the model must integrate several retrieved trajectories on ScienceWorld. However, we would avoid claiming that this experiment proves it is \emph{the} main bottleneck. The comparison is not fully controlled: the ALFWorld discrepancy strongly suggests that the two Qwen variants may differ not only in context length but also in training data or post-training recipe. A more defensible conclusion is therefore that long-context support is a plausible and practically important factor behind the improved large-\(K\) behavior of Qwen~2.5~7B~1M, while other factors remain entangled in this particular model pair.

Also, surprisingly, the 1M variant demonstrated weaker performance for dynamic retrieval compared to static retrieval on mismatched scenarios. One may expect that the Sparse Attention Mechanism from Qwen~2.5~7B~1M would help to reduce the instability of dynamic retrieval, but it does not seem to be the case.

\subsection{Impact of trajectory formatting on retrieval performance}
\label{subapp:trajectory_formatting}
All the trajectories are stored as raw chat-formatted data in standard OpenAI message format (see example in Figure~\ref{json-example}). However, when retrieved, we can prepend the trajectory examples in a different format to make them shorter, and hopefully easier for the model to process in-context. Figure~\ref{fig:traj_format} provides examples of different trajectory formatting. Please note that once the retrieved trajectory is formatted, it is appended to the system prompt and treated as plain text by the model; it is therefore different from the chat template used by the model. 

\input{figures/appendix/traj_format}

Table~\ref{tab:prompt_formatting} compares the impact of trajectory formatting on final \ours performance. Results indicate that the effect of trajectory formatting is strongly backbone-dependent. 
JSON-based formats are clearly preferable for the Qwen models, where textual formatting can degrade sharply at larger top-\(K\), but this pattern does not hold uniformly for Ministral~3-8B and Gemma~3~4B, for which textual formatting is often competitive and sometimes best. Among the JSON variants, chat JSON is a strong and relatively robust default, especially at larger \(K\), but it is not a universal optimum: compact and agentic JSON each outperform it in some settings. We therefore use chat JSON in the main experiments as a consistent default rather than because it dominates every model-format combination.

\input{tables/zero_shot_appendix/prompt_formatting}

%% file: tables/zero_shot_appendix/zero_shot_ministral_std.tex
\begin{table*}[t]
    \centering
    \small
    \setlength{\tabcolsep}{5pt}
    \newsavebox{\zeroshotministralstdbox}
    \sbox{\zeroshotministralstdbox}{%
    \begin{tabular}{l c c c c c c c c}
    \toprule
    \multirow{2}{*}{\makecell{\oursraw\\type}} & \multirow{2}{*}{Top-$K$} & \multirow{2}{*}{Index} & \multicolumn{3}{c}{ALFWorld} & \multicolumn{3}{c}{ScienceWorld} \\
    \cmidrule(lr){4-6} \cmidrule(lr){7-9}
    & & & Easy tasks & Hard tasks & All tasks & Easy tasks & Hard tasks & All tasks \\
    \midrule
    No RAG & 0 & -- & 8.22{\scriptsize $\pm 2.37$} & 0.00{\scriptsize $\pm 0.00$} & 4.48{\scriptsize $\pm 1.29$} & 14.41{\scriptsize $\pm 3.24$} & 5.08{\scriptsize $\pm 1.50$} & 10.40{\scriptsize $\pm 1.78$} \\
    \midrule
    \multirow{3}{*}{static} & \multirow[c]{6}{*}{1} & all & 43.84{\scriptsize $\pm 4.94$} & 33.88{\scriptsize $\pm 6.20$} & 39.30{\scriptsize $\pm 5.40$} & 33.33{\scriptsize $\pm 1.80$} & 17.71{\scriptsize $\pm 2.39$} & 26.62{\scriptsize $\pm 2.05$} \\
     &  & easy & 42.93{\scriptsize $\pm 0.79$} & 5.47{\scriptsize $\pm 0.95$} & 25.87{\scriptsize $\pm 0.87$} & 28.83{\scriptsize $\pm 0.68$} & 10.16{\scriptsize $\pm 3.25$} & 20.81{\scriptsize $\pm 1.45$} \\
     &  & hard & 35.62{\scriptsize $\pm 4.94$} & 33.88{\scriptsize $\pm 4.73$} & 34.83{\scriptsize $\pm 0.86$} & 22.65{\scriptsize $\pm 1.48$} & \textbf{17.97}{\scriptsize $\pm 2.02$} & 20.64{\scriptsize $\pm 1.27$} \\
    \addlinespace[0.4em]
    \multirow{3}{*}{dynamic} &  & all & 46.92{\scriptsize $\pm 0.69$} & 34.02{\scriptsize $\pm 0.82$} & 41.04{\scriptsize $\pm 0.61$} & 29.81{\scriptsize $\pm 2.71$} & 17.71{\scriptsize $\pm 2.39$} & 24.61{\scriptsize $\pm 1.69$} \\
     &  & easy & 44.75{\scriptsize $\pm 3.45$} & 4.37{\scriptsize $\pm 0.95$} & 26.37{\scriptsize $\pm 2.28$} & 32.65{\scriptsize $\pm 1.13$} & 7.03{\scriptsize $\pm 0.90$} & 21.65{\scriptsize $\pm 0.64$} \\
     &  & hard & 38.82{\scriptsize $\pm 5.19$} & 39.89{\scriptsize $\pm 0.95$} & 39.30{\scriptsize $\pm 3.11$} & \textbf{26.18}{\scriptsize $\pm 2.01$} & 19.53{\scriptsize $\pm 3.72$} & 23.32{\scriptsize $\pm 2.00$} \\
    \midrule
    \multirow{3}{*}{static} & \multirow[c]{6}{*}{2} & all & 53.42{\scriptsize $\pm 2.37$} & 48.09{\scriptsize $\pm 3.79$} & 50.99{\scriptsize $\pm 1.56$} & \textbf{44.31}{\scriptsize $\pm 3.40$} & 21.88{\scriptsize $\pm 3.13$} & \textbf{34.67}{\scriptsize $\pm 2.36$} \\
     &  & easy & 47.27{\scriptsize $\pm 0.97$} & \textbf{11.48}{\scriptsize $\pm 2.31$} & 30.97{\scriptsize $\pm 1.58$} & \textbf{43.53}{\scriptsize $\pm 1.29$} & 12.24{\scriptsize $\pm 2.30$} & \textbf{30.09}{\scriptsize $\pm 1.43$} \\
     &  & hard & 39.73{\scriptsize $\pm 1.37$} & 46.99{\scriptsize $\pm 1.89$} & 43.04{\scriptsize $\pm 1.56$} & 24.12{\scriptsize $\pm 1.93$} & 15.36{\scriptsize $\pm 1.18$} & 20.36{\scriptsize $\pm 1.39$} \\
    \addlinespace[0.4em]
    \multirow{3}{*}{dynamic} &  & all & 57.53{\scriptsize $\pm 1.94$} & 48.36{\scriptsize $\pm 0.95$} & 53.36{\scriptsize $\pm 0.96$} & 43.53{\scriptsize $\pm 0.00$} & 21.88{\scriptsize $\pm 2.21$} & 34.23{\scriptsize $\pm 0.95$} \\
     &  & easy & 57.53{\scriptsize $\pm 3.87$} & 6.56{\scriptsize $\pm 0.00$} & 34.33{\scriptsize $\pm 2.11$} & 43.73{\scriptsize $\pm 2.03$} & 9.64{\scriptsize $\pm 2.08$} & 29.08{\scriptsize $\pm 1.83$} \\
     &  & hard & 39.27{\scriptsize $\pm 2.09$} & 48.09{\scriptsize $\pm 1.89$} & 43.28{\scriptsize $\pm 1.98$} & 25.10{\scriptsize $\pm 2.31$} & 22.92{\scriptsize $\pm 3.07$} & \textbf{24.16}{\scriptsize $\pm 2.36$} \\
    \midrule
    \multirow{3}{*}{static} & \multirow[c]{6}{*}{4} & all & \textbf{63.47}{\scriptsize $\pm 1.58$} & \textbf{65.03}{\scriptsize $\pm 5.75$} & \textbf{64.18}{\scriptsize $\pm 1.97$} & 42.06{\scriptsize $\pm 2.61$} & \textbf{22.27}{\scriptsize $\pm 1.50$} & 33.56{\scriptsize $\pm 2.12$} \\
     &  & easy & \textbf{63.01}{\scriptsize $\pm 1.37$} & 7.11{\scriptsize $\pm 0.95$} & \textbf{37.56}{\scriptsize $\pm 0.86$} & 20.39{\scriptsize $\pm 1.80$} & \textbf{17.71}{\scriptsize $\pm 0.90$} & 19.24{\scriptsize $\pm 1.40$} \\
     &  & hard & \textbf{40.19}{\scriptsize $\pm 0.79$} & \textbf{62.84}{\scriptsize $\pm 3.41$} & \textbf{50.50}{\scriptsize $\pm 1.88$} & \textbf{41.57}{\scriptsize $\pm 2.45$} & 13.02{\scriptsize $\pm 3.93$} & \textbf{29.31}{\scriptsize $\pm 0.39$} \\
    \addlinespace[0.4em]
    \multirow{3}{*}{dynamic} &  & all & \textbf{70.55}{\scriptsize $\pm 3.26$} & \textbf{55.74}{\scriptsize $\pm 1.89$} & \textbf{63.81}{\scriptsize $\pm 1.29$} & \textbf{44.71}{\scriptsize $\pm 1.66$} & \textbf{22.66}{\scriptsize $\pm 2.70$} & \textbf{35.24}{\scriptsize $\pm 1.69$} \\
     &  & easy & \textbf{71.69}{\scriptsize $\pm 0.79$} & \textbf{9.29}{\scriptsize $\pm 0.95$} & \textbf{43.28}{\scriptsize $\pm 0.75$} & \textbf{46.67}{\scriptsize $\pm 4.45$} & \textbf{10.94}{\scriptsize $\pm 1.56$} & \textbf{31.32}{\scriptsize $\pm 2.71$} \\
     &  & hard & \textbf{42.93}{\scriptsize $\pm 1.58$} & \textbf{65.57}{\scriptsize $\pm 1.64$} & \textbf{53.24}{\scriptsize $\pm 0.43$} & 17.25{\scriptsize $\pm 1.80$} & \textbf{23.96}{\scriptsize $\pm 0.90$} & 20.14{\scriptsize $\pm 1.17$} \\
    \bottomrule
    \end{tabular}%
    }
    \ACLorArxivorOther{
        \usebox{\zeroshotministralstdbox}
    }{
        \usebox{\zeroshotministralstdbox}
    }{
        \resizebox{\textwidth}{!}{\usebox{\zeroshotministralstdbox}}
    }
    \caption{\textbf{Mean and standard deviation of success-rates for \ours inference without training across ALFWorld and ScienceWorld for Ministral 3-8B.} Values are reported as mean \(\pm\) std over \(3\) seeds. Across entries, the median/max std is \(1.58/6.20\) points on ALFWorld and \(1.83/4.45\) on ScienceWorld. The \textbf{bold} mean values mark the best top-$k$ setting for each (\ours type, index) within a given dataset split.}
    \label{tab:zero-shot-ministral-std}
    \end{table*}

%% file: tables/zero_shot_appendix/zeroshot_std_extraanalysis.tex
\begin{table*}[t]
    \centering
    \small
    \setlength{\tabcolsep}{5pt}
    \newsavebox{\zeroshotstdgroupbox}
    \newsavebox{\zeroshotstdgroupotherbox}
    \newsavebox{\zeroshotstdsplitbox}
    \newsavebox{\zeroshotstdsplitotherbox}
    \newsavebox{\zeroshotstdindexbox}
    \begin{subtable}[t]{0.47\textwidth}
        \centering
        \sbox{\zeroshotstdgroupbox}{%
            \begin{tabular*}{\linewidth}{@{\extracolsep{\fill}}lcccc}
            \toprule
            Grouping & \multicolumn{2}{c}{ALFWorld} & \multicolumn{2}{c}{ScienceWorld} \\
            \cmidrule(lr){2-3} \cmidrule(lr){4-5}
             & Median std & Max std & Median std & Max std \\
            \midrule
            No RAG & 1.29 & 2.37 & 1.78 & 3.24 \\
            static & 1.58 & 6.20 & 1.80 & 3.93 \\
            dynamic & 1.58 & 5.19 & 2.00 & 4.45 \\
            \midrule
            Top-$k$ = 0 & 1.29 & 2.37 & 1.78 & 3.24 \\
            Top-$k$ = 1 & 1.62 & 6.20 & 1.90 & 3.72 \\
            Top-$k$ = 2 & 1.89 & 3.87 & 2.06 & 3.40 \\
            Top-$k$ = 4 & 1.48 & 5.75 & 1.75 & 4.45 \\
            \bottomrule
            \end{tabular*}
        }
        \sbox{\zeroshotstdgroupotherbox}{%
            \begin{tabular}{lcccc}
            \toprule
            Grouping & \multicolumn{2}{c}{ALFWorld} & \multicolumn{2}{c}{ScienceWorld} \\
            \cmidrule(lr){2-3} \cmidrule(lr){4-5}
             & Median std & Max std & Median std & Max std \\
            \midrule
            No RAG & 1.29 & 2.37 & 1.78 & 3.24 \\
            static & 1.58 & 6.20 & 1.80 & 3.93 \\
            dynamic & 1.58 & 5.19 & 2.00 & 4.45 \\
            \midrule
            Top-$k$ = 0 & 1.29 & 2.37 & 1.78 & 3.24 \\
            Top-$k$ = 1 & 1.62 & 6.20 & 1.90 & 3.72 \\
            Top-$k$ = 2 & 1.89 & 3.87 & 2.06 & 3.40 \\
            Top-$k$ = 4 & 1.48 & 5.75 & 1.75 & 4.45 \\
            \bottomrule
            \end{tabular}
        }
        \ACLorArxivorOther{
            \usebox{\zeroshotstdgroupbox}
        }{
            \usebox{\zeroshotstdgroupbox}
        }{
            \resizebox{\linewidth}{!}{\usebox{\zeroshotstdgroupotherbox}}
        }
        \caption{\textbf{Grouped by retrieval setting.} The first block aggregates by retrieval type; the second aggregates by top-$k$.}
        \label{tab:zero-shot-std-extraanalysis}
    \end{subtable}
    \hfill
    \begin{subtable}[t]{0.47\textwidth}
        \centering
        \sbox{\zeroshotstdsplitbox}{%
            \begin{tabular*}{\linewidth}{@{\extracolsep{\fill}}lcccc}
            \toprule
            Split & \multicolumn{2}{c}{ALFWorld} & \multicolumn{2}{c}{ScienceWorld} \\
            \cmidrule(lr){2-3}\cmidrule(lr){4-5}
            & Median std & Max std & Median std & Max std \\
            \midrule
            Easy Tasks & 1.94 & 5.19 & 1.93 & 4.45 \\
            Hard Tasks & 1.64 & 6.20 & 2.21 & 3.93 \\
            All Tasks & 1.56 & 5.40 & 1.69 & 2.71 \\
            \bottomrule
            \end{tabular*}
        }
        \sbox{\zeroshotstdsplitotherbox}{%
            \begin{tabular}{lcccc}
            \toprule
            Split & \multicolumn{2}{c}{ALFWorld} & \multicolumn{2}{c}{ScienceWorld} \\
            \cmidrule(lr){2-3}\cmidrule(lr){4-5}
            & Median std & Max std & Median std & Max std \\
            \midrule
            Easy Tasks & 1.94 & 5.19 & 1.93 & 4.45 \\
            Hard Tasks & 1.64 & 6.20 & 2.21 & 3.93 \\
            All Tasks & 1.56 & 5.40 & 1.69 & 2.71 \\
            \bottomrule
            \end{tabular}
        }
        \ACLorArxivorOther{
            \usebox{\zeroshotstdsplitbox}
        }{
            \usebox{\zeroshotstdsplitbox}
        }{
            \resizebox{\linewidth}{!}{\usebox{\zeroshotstdsplitotherbox}}
        }
        \caption{\textbf{Grouped by evaluation split.} Each row pools all configurations evaluated on that split.}
        \label{tab:zero-shot-ministral-std-split-summary}
    \end{subtable}
    
    \vspace{0.8em}
    
    \begin{subtable}[t]{\textwidth}
        \centering
        \sbox{\zeroshotstdindexbox}{%
            \begin{tabular}{lcccccc}
            \toprule
            Index & \multicolumn{3}{c}{ALFWorld} & \multicolumn{3}{c}{ScienceWorld} \\
            \cmidrule(lr){2-4}\cmidrule(lr){5-7}
             & Easy Tasks & Hard Tasks & All Tasks & Easy Tasks & Hard Tasks & All Tasks \\
            \midrule
            all & 2.16 / 4.94 & 2.84 / 6.20 & 1.43 / 5.40 & 2.21 / 3.40 & 2.39 / 3.13 & 1.87 / 2.36 \\
            easy & 1.17 / 3.87 & 0.95 / 2.31 & 1.23 / 2.28 & 1.54 / 4.45 & 1.82 / 3.25 & 1.44 / 2.71 \\
            hard & 1.83 / 5.19 & 1.89 / 4.73 & 1.72 / 3.11 & 1.97 / 2.45 & 2.54 / 3.93 & 1.33 / 2.36 \\
            \bottomrule
            \end{tabular}
        }
        \ACLorArxivorOther{
            \usebox{\zeroshotstdindexbox}
        }{
            \usebox{\zeroshotstdindexbox}
        }{
            \resizebox{0.9\linewidth}{!}{\usebox{\zeroshotstdindexbox}}
        }
        \caption{\textbf{Grouped by retrieval index.} Each cell reports median/max std over all configurations with that index and benchmark split.}
        \label{tab:zero-shot-ministral-std-index-summary}
    \end{subtable}
    \caption{\textbf{Summary statistics for standard deviations in zero-shot Ministral 3-8B results.} The top row summarizes variability by retrieval setting and by evaluation split. The bottom table breaks the same standard deviations down jointly by retrieval index and benchmark split.}
    \label{tab:zero-shot-ministral-std-summary}
\end{table*}

%% file: tables/zero_shot_appendix/zeroshot_gemma3_4b.tex
\begin{table*}[!h]
\centering
\small
\setlength{\tabcolsep}{5pt}
\newsavebox{\zeroshotgemmabox}
\sbox{\zeroshotgemmabox}{%
\begin{tabular}{l c c c c c c c c}
\toprule
\multirow{2}{*}{\makecell{\oursraw\\type}} & \multirow{2}{*}{Top-$K$} & \multirow{2}{*}{Index} & \multicolumn{3}{c}{ALFWorld} & \multicolumn{3}{c}{ScienceWorld} \\
\cmidrule(lr){4-6} \cmidrule(lr){7-9}
& & & Easy tasks & Hard tasks & All tasks & Easy tasks & Hard tasks & All tasks \\
\midrule
No RAG & 0 & -- & 1.37 & 0.00 & 0.75 & 0.00 & 4.69 & 2.01 \\
\midrule
\multirow{3}{*}{static} & \multirow[c]{6}{*}{1} & all & 20.55 & 3.28 & 12.69 & 4.71 & 1.56 & 3.36 \\
 &  & easy & 17.81 & 0.00 & 9.70 & 7.06 & \textbf{4.69} & 6.04 \\
 &  & hard & \textbf{8.22} & 3.28 & \textbf{5.97} & 2.35 & 0.00 & 1.34 \\
\addlinespace[0.4em]
\multirow{3}{*}{dynamic} &  & all & 19.18{\scriptsize \textcolor{red!70!black}{\,$\downarrow 1.4$}} & 3.28{\scriptsize \,0.0} & 11.94{\scriptsize \textcolor{red!70!black}{\,$\downarrow 0.8$}} & 7.06{\scriptsize \textcolor{green!50!black}{\,$\uparrow 2.4$}} & 3.12{\scriptsize \textcolor{green!50!black}{\,$\uparrow 1.6$}} & 5.37{\scriptsize \textcolor{green!50!black}{\,$\uparrow 2.0$}} \\
 &  & easy & 17.81{\scriptsize \,0.0} & 0.00{\scriptsize \,0.0} & 9.70{\scriptsize \,0.0} & 7.06{\scriptsize \,0.0} & 1.56{\scriptsize \textcolor{red!70!black}{\,$\downarrow 3.1$}} & 4.70{\scriptsize \textcolor{red!70!black}{\,$\downarrow 1.3$}} \\
 &  & hard & \textbf{9.59}{\scriptsize \textcolor{green!50!black}{\,$\uparrow 1.4$}} & 6.56{\scriptsize \textcolor{green!50!black}{\,$\uparrow 3.3$}} & 8.21{\scriptsize \textcolor{green!50!black}{\,$\uparrow 2.2$}} & 4.71{\scriptsize \textcolor{green!50!black}{\,$\uparrow 2.4$}} & 3.12{\scriptsize \textcolor{green!50!black}{\,$\uparrow 3.1$}} & 4.03{\scriptsize \textcolor{green!50!black}{\,$\uparrow 2.7$}} \\
\midrule
\multirow{3}{*}{static} & \multirow[c]{6}{*}{2} & all & 16.44 & 6.56 & 11.94 & 6.67 & 3.12 & 5.15 \\
 &  & easy & 21.92 & \textbf{4.92} & 14.18 & \textbf{10.59} & 1.56 & 6.71 \\
 &  & hard & 4.11 & 4.92 & 4.48 & 4.71 & \textbf{4.69} & 4.70 \\
\addlinespace[0.4em]
\multirow{3}{*}{dynamic} &  & all & \textbf{23.29}{\scriptsize \textcolor{green!50!black}{\,$\uparrow 6.8$}} & 4.10{\scriptsize \textcolor{red!70!black}{\,$\downarrow 2.5$}} & \textbf{14.55}{\scriptsize \textcolor{green!50!black}{\,$\uparrow 2.6$}} & 9.41{\scriptsize \textcolor{green!50!black}{\,$\uparrow 2.7$}} & 3.12{\scriptsize \,0.0} & 6.71{\scriptsize \textcolor{green!50!black}{\,$\uparrow 1.6$}} \\
 &  & easy & 26.03{\scriptsize \textcolor{green!50!black}{\,$\uparrow 4.1$}} & \textbf{3.28}{\scriptsize \textcolor{red!70!black}{\,$\downarrow 1.6$}} & 15.67{\scriptsize \textcolor{green!50!black}{\,$\uparrow 1.5$}} & 9.41{\scriptsize \textcolor{red!70!black}{\,$\downarrow 1.2$}} & 1.56{\scriptsize \,0.0} & 6.04{\scriptsize \textcolor{red!70!black}{\,$\downarrow 0.7$}} \\
 &  & hard & \textbf{9.59}{\scriptsize \textcolor{green!50!black}{\,$\uparrow 5.5$}} & \textbf{8.20}{\scriptsize \textcolor{green!50!black}{\,$\uparrow 3.3$}} & \textbf{8.96}{\scriptsize \textcolor{green!50!black}{\,$\uparrow 4.5$}} & 3.53{\scriptsize \textcolor{red!70!black}{\,$\downarrow 1.2$}} & 3.12{\scriptsize \textcolor{red!70!black}{\,$\downarrow 1.6$}} & 3.36{\scriptsize \textcolor{red!70!black}{\,$\downarrow 1.3$}} \\
\midrule
\multirow{3}{*}{static} & \multirow[c]{6}{*}{4} & all & \textbf{21.92} & \textbf{19.67} & \textbf{20.90} & \textbf{8.24} & \textbf{7.81} & \textbf{8.05} \\
 &  & easy & \textbf{24.66} & 3.28 & \textbf{14.93} & 9.41 & \textbf{4.69} & \textbf{7.38} \\
 &  & hard & 4.11 & \textbf{6.56} & 5.22 & \textbf{7.06} & 3.12 & \textbf{5.37} \\
\addlinespace[0.4em]
\multirow{3}{*}{dynamic} &  & all & 17.81{\scriptsize \textcolor{red!70!black}{\,$\downarrow 4.1$}} & \textbf{6.56}{\scriptsize \textcolor{red!70!black}{\,$\downarrow 13.1$}} & 12.69{\scriptsize \textcolor{red!70!black}{\,$\downarrow 8.2$}} & \textbf{14.12}{\scriptsize \textcolor{green!50!black}{\,$\uparrow 5.9$}} & \textbf{7.81}{\scriptsize \,0.0} & \textbf{11.41}{\scriptsize \textcolor{green!50!black}{\,$\uparrow 3.4$}} \\
 &  & easy & \textbf{27.40}{\scriptsize \textcolor{green!50!black}{\,$\uparrow 2.7$}} & \textbf{3.28}{\scriptsize \,0.0} & \textbf{16.42}{\scriptsize \textcolor{green!50!black}{\,$\uparrow 1.5$}} & \textbf{14.12}{\scriptsize \textcolor{green!50!black}{\,$\uparrow 4.7$}} & \textbf{3.12}{\scriptsize \textcolor{red!70!black}{\,$\downarrow 1.6$}} & \textbf{9.40}{\scriptsize \textcolor{green!50!black}{\,$\uparrow 2.0$}} \\
 &  & hard & \textbf{9.59}{\scriptsize \textcolor{green!50!black}{\,$\uparrow 5.5$}} & 6.56{\scriptsize \,0.0} & 8.21{\scriptsize \textcolor{green!50!black}{\,$\uparrow 3.0$}} & \textbf{5.88}{\scriptsize \textcolor{red!70!black}{\,$\downarrow 1.2$}} & \textbf{12.50}{\scriptsize \textcolor{green!50!black}{\,$\uparrow 9.4$}} & \textbf{8.72}{\scriptsize \textcolor{green!50!black}{\,$\uparrow 3.4$}} \\
\bottomrule
\end{tabular}
}
\ACLorArxivorOther{
    \usebox{\zeroshotgemmabox}
}{
    \usebox{\zeroshotgemmabox}
}{
    \resizebox{\linewidth}{!}{\usebox{\zeroshotgemmabox}}
}
\caption{\textbf{Results for \ours inference without training across ALFWorld and ScienceWorld for Gemma~3~4B.} The \textbf{bold} values mark the best top-$k$ setting for each (\ours type, index) within a given dataset split. For dynamic retrieval rows, the superscript arrows report the per-cell difference against the corresponding static row with the same index and top-$k$: \textcolor{green!50!black}{$\uparrow$} indicates improvement, \textcolor{red!70!black}{$\downarrow$} indicates a decline, and black values denote ties.}
\label{tab:zero-shot-gemma3-4b}
\end{table*}

%% file: tables/zero_shot_appendix/zeroshot_qwen25_3b.tex
\begin{table*}[!h]
\centering
\small
\setlength{\tabcolsep}{5pt}
\newsavebox{\zeroshotqwentwentyfivethreebbox}
\sbox{\zeroshotqwentwentyfivethreebbox}{%
\begin{tabular}{l c c c c c c c c}
\toprule
\multirow{2}{*}{\makecell{\oursraw\\type}} & \multirow{2}{*}{Top-$K$} & \multirow{2}{*}{Index} & \multicolumn{3}{c}{ALFWorld} & \multicolumn{3}{c}{ScienceWorld} \\
\cmidrule(lr){4-6} \cmidrule(lr){7-9}
& & & Easy tasks & Hard tasks & All tasks & Easy tasks & Hard tasks & All tasks \\
\midrule
No RAG & 0 & -- & 5.48 & 4.92 & 5.22 & 1.18 & 4.69 & 2.68 \\
\midrule
\multirow{3}{*}{static} & \multirow[c]{6}{*}{1} & all & 20.55 & 11.48 & 16.42 & 5.88 & 3.12 & 4.70 \\
 &  & easy & \textbf{24.66} & \textbf{3.28} & \textbf{14.93} & 7.06 & \textbf{1.56} & 4.70 \\
 &  & hard & 8.22 & 18.03 & 12.69 & 2.35 & 3.12 & 2.68 \\
\addlinespace[0.4em]
\multirow{3}{*}{dynamic} &  & all & 24.66{\scriptsize \textcolor{green!50!black}{\,$\uparrow 4.1$}} & \textbf{19.67}{\scriptsize \textcolor{green!50!black}{\,$\uparrow 8.2$}} & 22.39{\scriptsize \textcolor{green!50!black}{\,$\uparrow 6.0$}} & -- & -- & -- \\
 &  & easy & 23.29{\scriptsize \textcolor{red!70!black}{\,$\downarrow 1.4$}} & 1.64{\scriptsize \textcolor{red!70!black}{\,$\downarrow 1.6$}} & 13.43{\scriptsize \textcolor{red!70!black}{\,$\downarrow 1.5$}} & 7.06{\scriptsize \,0.0} & \textbf{0.00}{\scriptsize \textcolor{red!70!black}{\,$\downarrow 1.6$}} & 4.03{\scriptsize \textcolor{red!70!black}{\,$\downarrow 0.7$}} \\
 &  & hard & \textbf{9.59}{\scriptsize \textcolor{green!50!black}{\,$\uparrow 1.4$}} & \textbf{22.95}{\scriptsize \textcolor{green!50!black}{\,$\uparrow 4.9$}} & \textbf{15.67}{\scriptsize \textcolor{green!50!black}{\,$\uparrow 3.0$}} & \textbf{2.35}{\scriptsize \,0.0} & 0.00{\scriptsize \textcolor{red!70!black}{\,$\downarrow 3.1$}} & 1.34{\scriptsize \textcolor{red!70!black}{\,$\downarrow 1.3$}} \\
\midrule
\multirow{3}{*}{static} & \multirow[c]{4}{*}{2} & all & 26.03 & 14.75 & 20.90 & 4.71 & \textbf{4.69} & 4.70 \\
 &  & easy & 23.29 & 1.64 & 13.43 & \textbf{12.94} & 0.00 & \textbf{7.38} \\
 &  & hard & \textbf{15.07} & \textbf{21.31} & \textbf{17.91} & \textbf{3.53} & \textbf{4.69} & \textbf{4.03} \\
\addlinespace[0.4em]
\multirow{1}{*}{dynamic} &  & all & \textbf{30.14}{\scriptsize \textcolor{green!50!black}{\,$\uparrow 4.1$}} & 18.03{\scriptsize \textcolor{green!50!black}{\,$\uparrow 3.3$}} & \textbf{24.63}{\scriptsize \textcolor{green!50!black}{\,$\uparrow 3.7$}} & 5.88{\scriptsize \textcolor{green!50!black}{\,$\uparrow 1.2$}} & \textbf{6.25}{\scriptsize \textcolor{green!50!black}{\,$\uparrow 1.6$}} & 6.04{\scriptsize \textcolor{green!50!black}{\,$\uparrow 1.3$}} \\
\midrule
\multirow{3}{*}{static} & \multirow[c]{6}{*}{4} & all & \textbf{31.51} & \textbf{18.03} & \textbf{25.37} & \textbf{7.06} & 3.12 & \textbf{5.37} \\
 &  & easy & \textbf{24.66} & \textbf{3.28} & \textbf{14.93} & 4.71 & 0.00 & 2.68 \\
 &  & hard & 10.96 & \textbf{21.31} & 15.67 & \textbf{3.53} & \textbf{4.69} & \textbf{4.03} \\
\addlinespace[0.4em]
\multirow{3}{*}{dynamic} &  & all & \textbf{30.14}{\scriptsize \textcolor{red!70!black}{\,$\downarrow 1.4$}} & 8.20{\scriptsize \textcolor{red!70!black}{\,$\downarrow 9.8$}} & 20.15{\scriptsize \textcolor{red!70!black}{\,$\downarrow 5.2$}} & \textbf{11.76}{\scriptsize \textcolor{green!50!black}{\,$\uparrow 4.7$}} & 0.00{\scriptsize \textcolor{red!70!black}{\,$\downarrow 3.1$}} & \textbf{6.71}{\scriptsize \textcolor{green!50!black}{\,$\uparrow 1.3$}} \\
 &  & easy & \textbf{24.66}{\scriptsize \,0.0} & \textbf{6.56}{\scriptsize \textcolor{green!50!black}{\,$\uparrow 3.3$}} & \textbf{16.42}{\scriptsize \textcolor{green!50!black}{\,$\uparrow 1.5$}} & \textbf{14.12}{\scriptsize \textcolor{green!50!black}{\,$\uparrow 9.4$}} & \textbf{0.00}{\scriptsize \,0.0} & \textbf{8.05}{\scriptsize \textcolor{green!50!black}{\,$\uparrow 5.4$}} \\
 &  & hard & \textbf{9.59}{\scriptsize \textcolor{red!70!black}{\,$\downarrow 1.4$}} & 14.75{\scriptsize \textcolor{red!70!black}{\,$\downarrow 6.6$}} & 11.94{\scriptsize \textcolor{red!70!black}{\,$\downarrow 3.7$}} & 1.18{\scriptsize \textcolor{red!70!black}{\,$\downarrow 2.4$}} & \textbf{6.25}{\scriptsize \textcolor{green!50!black}{\,$\uparrow 1.6$}} & \textbf{3.36}{\scriptsize \textcolor{red!70!black}{\,$\downarrow 0.7$}} \\
\bottomrule
\end{tabular}
}
\ACLorArxivorOther{
    \usebox{\zeroshotqwentwentyfivethreebbox}
}{
    \usebox{\zeroshotqwentwentyfivethreebbox}
}{
    \resizebox{\linewidth}{!}{\usebox{\zeroshotqwentwentyfivethreebbox}}
}
\caption{\textbf{Results for \ours inference without training across ALFWorld and ScienceWorld for Qwen~2.5~3B.} The \textbf{bold} values mark the best top-$k$ setting for each (\ours type, index) within a given dataset split. For dynamic retrieval rows, the superscript arrows report the per-cell difference against the corresponding static row with the same index and top-$k$: \textcolor{green!50!black}{$\uparrow$} indicates improvement, \textcolor{red!70!black}{$\downarrow$} indicates a decline, and black values denote ties.}
\label{tab:zero-shot-qwen25-3b}
\end{table*}

%% file: tables/zero_shot_appendix/zeroshot_qwen25_7b.tex
\begin{table*}[!h]
\centering
\small
\setlength{\tabcolsep}{5pt}
\newsavebox{\zeroshotqwentwentyfivesevenbbox}
\sbox{\zeroshotqwentwentyfivesevenbbox}{%
\begin{tabular}{l c c c c c c c c}
\toprule
\multirow{2}{*}{\makecell{\oursraw\\type}} & \multirow{2}{*}{Top-$K$} & \multirow{2}{*}{Index} & \multicolumn{3}{c}{ALFWorld} & \multicolumn{3}{c}{ScienceWorld} \\
\cmidrule(lr){4-6} \cmidrule(lr){7-9}
& & & Easy tasks & Hard tasks & All tasks & Easy tasks & Hard tasks & All tasks \\
\midrule
No RAG & 0 & -- & 35.62 & 22.95 & 29.85 & 1.18 & 3.12 & 2.01 \\
\midrule
\multirow{3}{*}{static} & \multirow[c]{6}{*}{1} & all & 73.97 & 59.02 & 67.16 & \textbf{17.65} & 4.69 & \textbf{12.08} \\
 &  & easy & 71.23 & 31.15 & 52.99 & \textbf{16.47} & \textbf{3.12} & \textbf{10.74} \\
 &  & hard & 64.38 & 62.30 & 63.43 & \textbf{7.06} & \textbf{7.81} & \textbf{7.38} \\
\addlinespace[0.4em]
\multirow{3}{*}{dynamic} &  & all & 72.60{\scriptsize \textcolor{red!70!black}{\,$\downarrow 1.4$}} & 65.57{\scriptsize \textcolor{green!50!black}{\,$\uparrow 6.6$}} & 69.40{\scriptsize \textcolor{green!50!black}{\,$\uparrow 2.2$}} & 11.76{\scriptsize \textcolor{red!70!black}{\,$\downarrow 5.9$}} & 7.81{\scriptsize \textcolor{green!50!black}{\,$\uparrow 3.1$}} & 10.07{\scriptsize \textcolor{red!70!black}{\,$\downarrow 2.0$}} \\
 &  & easy & 73.97{\scriptsize \textcolor{green!50!black}{\,$\uparrow 2.7$}} & 34.43{\scriptsize \textcolor{green!50!black}{\,$\uparrow 3.3$}} & 55.97{\scriptsize \textcolor{green!50!black}{\,$\uparrow 3.0$}} & \textbf{12.94}{\scriptsize \textcolor{red!70!black}{\,$\downarrow 3.5$}} & \textbf{3.12}{\scriptsize \,0.0} & \textbf{8.72}{\scriptsize \textcolor{red!70!black}{\,$\downarrow 2.0$}} \\
 &  & hard & \textbf{64.38}{\scriptsize \,0.0} & 62.30{\scriptsize \,0.0} & \textbf{63.43}{\scriptsize \,0.0} & \textbf{3.53}{\scriptsize \textcolor{red!70!black}{\,$\downarrow 3.5$}} & 7.81{\scriptsize \,0.0} & \textbf{5.37}{\scriptsize \textcolor{red!70!black}{\,$\downarrow 2.0$}} \\
\midrule
\multirow{3}{*}{static} & \multirow[c]{4}{*}{2} & all & \textbf{83.56} & 81.97 & 82.84 & 12.94 & \textbf{10.94} & \textbf{12.08} \\
 &  & easy & 71.23 & \textbf{54.10} & \textbf{63.43} & \textbf{16.47} & 0.00 & 9.40 \\
 &  & hard & \textbf{73.97} & 68.85 & \textbf{71.64} & 4.71 & 6.25 & 5.37 \\
\addlinespace[0.4em]
\multirow{1}{*}{dynamic} &  & all & \textbf{82.19}{\scriptsize \textcolor{red!70!black}{\,$\downarrow 1.4$}} & 62.30{\scriptsize \textcolor{red!70!black}{\,$\downarrow 19.7$}} & 73.13{\scriptsize \textcolor{red!70!black}{\,$\downarrow 9.7$}} & 8.24{\scriptsize \textcolor{red!70!black}{\,$\downarrow 4.7$}} & 7.81{\scriptsize \textcolor{red!70!black}{\,$\downarrow 3.1$}} & 8.05{\scriptsize \textcolor{red!70!black}{\,$\downarrow 4.0$}} \\
\midrule
\multirow{3}{*}{static} & \multirow[c]{6}{*}{4} & all & 79.45 & \textbf{88.52} & \textbf{83.58} & 12.94 & 9.38 & 11.41 \\
 &  & easy & \textbf{78.08} & 42.62 & 61.94 & 9.41 & 0.00 & 5.37 \\
 &  & hard & 47.95 & \textbf{80.33} & 62.69 & 1.18 & \textbf{7.81} & 4.03 \\
\addlinespace[0.4em]
\multirow{3}{*}{dynamic} &  & all & 76.71{\scriptsize \textcolor{red!70!black}{\,$\downarrow 2.7$}} & \textbf{77.05}{\scriptsize \textcolor{red!70!black}{\,$\downarrow 11.5$}} & \textbf{76.87}{\scriptsize \textcolor{red!70!black}{\,$\downarrow 6.7$}} & \textbf{14.12}{\scriptsize \textcolor{green!50!black}{\,$\uparrow 1.2$}} & \textbf{10.94}{\scriptsize \textcolor{green!50!black}{\,$\uparrow 1.6$}} & \textbf{12.75}{\scriptsize \textcolor{green!50!black}{\,$\uparrow 1.3$}} \\
 &  & easy & \textbf{87.67}{\scriptsize \textcolor{green!50!black}{\,$\uparrow 9.6$}} & \textbf{44.26}{\scriptsize \textcolor{green!50!black}{\,$\uparrow 1.6$}} & \textbf{67.91}{\scriptsize \textcolor{green!50!black}{\,$\uparrow 6.0$}} & \textbf{12.94}{\scriptsize \textcolor{green!50!black}{\,$\uparrow 3.5$}} & 0.00{\scriptsize \,0.0} & 7.38{\scriptsize \textcolor{green!50!black}{\,$\uparrow 2.0$}} \\
 &  & hard & 56.16{\scriptsize \textcolor{green!50!black}{\,$\uparrow 8.2$}} & \textbf{67.21}{\scriptsize \textcolor{red!70!black}{\,$\downarrow 13.1$}} & 61.19{\scriptsize \textcolor{red!70!black}{\,$\downarrow 1.5$}} & 2.35{\scriptsize \textcolor{green!50!black}{\,$\uparrow 1.2$}} & \textbf{9.38}{\scriptsize \textcolor{green!50!black}{\,$\uparrow 1.6$}} & \textbf{5.37}{\scriptsize \textcolor{green!50!black}{\,$\uparrow 1.3$}} \\
\bottomrule
\end{tabular}
}
\ACLorArxivorOther{
    \usebox{\zeroshotqwentwentyfivesevenbbox}
}{
    \usebox{\zeroshotqwentwentyfivesevenbbox}
}{
    \resizebox{\linewidth}{!}{\usebox{\zeroshotqwentwentyfivesevenbbox}}
}
\caption{\textbf{Results for \ours inference without training across ALFWorld and ScienceWorld for Qwen~2.5~7B.} The \textbf{bold} values mark the best top-$k$ setting for each (\ours type, index) within a given dataset split. For dynamic retrieval rows, the superscript arrows report the per-cell difference against the corresponding static row with the same index and top-$k$: \textcolor{green!50!black}{$\uparrow$} indicates improvement, \textcolor{red!70!black}{$\downarrow$} indicates a decline, and black values denote ties.}
\label{tab:zero-shot-qwen25-7b}
\end{table*}

%% file: tables/zero_shot_appendix/zeroshot_qwen7b1m.tex
\begin{table*}[!h]
\centering
\small
\setlength{\tabcolsep}{5pt}
\newsavebox{\zeroshotqwensevenbonembox}
\sbox{\zeroshotqwensevenbonembox}{%
\begin{tabular}{l c c c c c c c c}
\toprule
\multirow{2}{*}{\makecell{\oursraw\\type}} & \multirow{2}{*}{Top-$K$} & \multirow{2}{*}{Index} & \multicolumn{3}{c}{ALFWorld} & \multicolumn{3}{c}{ScienceWorld} \\
\cmidrule(lr){4-6} \cmidrule(lr){7-9}
& & & Easy tasks & Hard tasks & All tasks & Easy tasks & Hard tasks & All tasks \\
\midrule
No RAG & 0 & -- & 6.85 & 3.28 & 5.22 & 3.53 & 3.12 & 3.36 \\
\midrule
\multirow{3}{*}{static} & \multirow[c]{6}{*}{1} & all & 64.38 & 59.02 & 61.94 & 15.29 & 6.25 & 11.41 \\
 &  & easy & 56.16 & 14.75 & 37.31 & 17.65 & \textbf{0.00} & 10.07 \\
 &  & hard & 39.73 & 59.02 & 48.51 & 7.06 & 6.25 & 6.71 \\
\addlinespace[0.4em]
\multirow{3}{*}{dynamic} &  & all & 63.01{\scriptsize \textcolor{red!70!black}{\,$\downarrow 1.4$}} & 57.38{\scriptsize \textcolor{red!70!black}{\,$\downarrow 1.6$}} & 60.45{\scriptsize \textcolor{red!70!black}{\,$\downarrow 1.5$}} & 12.94{\scriptsize \textcolor{red!70!black}{\,$\downarrow 2.4$}} & 0.00{\scriptsize \textcolor{red!70!black}{\,$\downarrow 6.2$}} & 7.38{\scriptsize \textcolor{red!70!black}{\,$\downarrow 4.0$}} \\
 &  & easy & 57.53{\scriptsize \textcolor{green!50!black}{\,$\uparrow 1.4$}} & 16.39{\scriptsize \textcolor{green!50!black}{\,$\uparrow 1.6$}} & 38.81{\scriptsize \textcolor{green!50!black}{\,$\uparrow 1.5$}} & \textbf{12.94}{\scriptsize \textcolor{red!70!black}{\,$\downarrow 4.7$}} & 0.00{\scriptsize \,0.0} & 7.38{\scriptsize \textcolor{red!70!black}{\,$\downarrow 2.7$}} \\
 &  & hard & 45.21{\scriptsize \textcolor{green!50!black}{\,$\uparrow 5.5$}} & 54.10{\scriptsize \textcolor{red!70!black}{\,$\downarrow 4.9$}} & 49.25{\scriptsize \textcolor{green!50!black}{\,$\uparrow 0.7$}} & 12.94{\scriptsize \textcolor{green!50!black}{\,$\uparrow 5.9$}} & \textbf{6.25}{\scriptsize \,0.0} & 10.07{\scriptsize \textcolor{green!50!black}{\,$\uparrow 3.4$}} \\
\midrule
\multirow{3}{*}{static} & \multirow[c]{4}{*}{2} & all & 71.23 & 52.46 & 62.69 & \textbf{20.00} & 14.06 & 17.45 \\
 &  & easy & 67.12 & 29.51 & 50.00 & \textbf{20.00} & \textbf{0.00} & \textbf{11.41} \\
 &  & hard & 47.95 & 54.10 & 50.75 & 8.24 & 7.81 & 8.05 \\
\addlinespace[0.4em]
\multirow{1}{*}{dynamic} &  & all & 63.01{\scriptsize \textcolor{red!70!black}{\,$\downarrow 8.2$}} & 49.18{\scriptsize \textcolor{red!70!black}{\,$\downarrow 3.3$}} & 56.72{\scriptsize \textcolor{red!70!black}{\,$\downarrow 6.0$}} & 16.47{\scriptsize \textcolor{red!70!black}{\,$\downarrow 3.5$}} & 14.06{\scriptsize \,0.0} & 15.44{\scriptsize \textcolor{red!70!black}{\,$\downarrow 2.0$}} \\
\midrule
\multirow{3}{*}{static} & \multirow[c]{6}{*}{4} & all & \textbf{80.82} & \textbf{81.97} & \textbf{81.34} & 16.47 & \textbf{32.81} & \textbf{23.49} \\
 &  & easy & \textbf{69.86} & \textbf{34.43} & \textbf{53.73} & 17.65 & \textbf{0.00} & 10.07 \\
 &  & hard & \textbf{63.01} & \textbf{72.13} & \textbf{67.16} & \textbf{11.76} & \textbf{25.00} & \textbf{17.45} \\
\addlinespace[0.4em]
\multirow{3}{*}{dynamic} &  & all & \textbf{75.34}{\scriptsize \textcolor{red!70!black}{\,$\downarrow 5.5$}} & \textbf{60.66}{\scriptsize \textcolor{red!70!black}{\,$\downarrow 21.3$}} & \textbf{68.66}{\scriptsize \textcolor{red!70!black}{\,$\downarrow 12.7$}} & \textbf{23.53}{\scriptsize \textcolor{green!50!black}{\,$\uparrow 7.1$}} & \textbf{32.81}{\scriptsize \,0.0} & \textbf{27.52}{\scriptsize \textcolor{green!50!black}{\,$\uparrow 4.0$}} \\
 &  & easy & \textbf{79.45}{\scriptsize \textcolor{green!50!black}{\,$\uparrow 9.6$}} & \textbf{21.31}{\scriptsize \textcolor{red!70!black}{\,$\downarrow 13.1$}} & \textbf{52.99}{\scriptsize \textcolor{red!70!black}{\,$\downarrow 0.7$}} & 9.41{\scriptsize \textcolor{red!70!black}{\,$\downarrow 8.2$}} & \textbf{28.12}{\scriptsize \textcolor{green!50!black}{\,$\uparrow 28.1$}} & \textbf{17.45}{\scriptsize \textcolor{green!50!black}{\,$\uparrow 7.4$}} \\
 &  & hard & \textbf{61.64}{\scriptsize \textcolor{red!70!black}{\,$\downarrow 1.4$}} & \textbf{62.30}{\scriptsize \textcolor{red!70!black}{\,$\downarrow 9.8$}} & \textbf{61.94}{\scriptsize \textcolor{red!70!black}{\,$\downarrow 5.2$}} & \textbf{25.88}{\scriptsize \textcolor{green!50!black}{\,$\uparrow 14.1$}} & 0.00{\scriptsize \textcolor{red!70!black}{\,$\downarrow 25.0$}} & \textbf{14.77}{\scriptsize \textcolor{red!70!black}{\,$\downarrow 2.7$}} \\
\bottomrule
\end{tabular}%
}
\ACLorArxivorOther{
    \usebox{\zeroshotqwensevenbonembox}
}{
    \usebox{\zeroshotqwensevenbonembox}
}{
    \resizebox{\linewidth}{!}{\usebox{\zeroshotqwensevenbonembox}}
}
\caption{\textbf{Results for \ours inference without training across ALFWorld and ScienceWorld for Qwen~2.5~7B~1M.} The \textbf{bold} values mark the best top-$k$ setting for each (\ours type, index) within a given dataset split. For dynamic retrieval rows, the superscript arrows report the per-cell difference against the corresponding static row with the same index and top-$k$: \textcolor{green!50!black}{$\uparrow$} indicates improvement, \textcolor{red!70!black}{$\downarrow$} indicates a decline, and black values denote ties.}
\label{tab:zero-shot-qwen7b1m}
\end{table*}

%% file: tables/zero_shot_appendix/compare_qwens_7B.tex
\begin{table*}[t]
\centering
\small
\setlength{\tabcolsep}{5pt}
\begin{tabular}{l c c c c c c c c}
\toprule
\multirow{2}{*}{\makecell{\oursraw\\type}} & \multirow{2}{*}{Top-$K$} & \multirow{2}{*}{Index} & \multicolumn{3}{c}{ALFWorld} & \multicolumn{3}{c}{ScienceWorld} \\
\cmidrule(lr){4-6} \cmidrule(lr){7-9}
& & & Easy tasks & Hard tasks & All tasks & Easy tasks & Hard tasks & All tasks \\
\midrule
No RAG & 0 & -- & \textcolor{red!70!black}{-28.77} & \textcolor{red!70!black}{-19.67} & \textcolor{red!70!black}{-24.63} & \textcolor{green!50!black}{+2.35} & 0.00 & \textcolor{green!50!black}{+1.35} \\
\midrule
\multirow{3}{*}{static} & \multirow[c]{6}{*}{1} & all & \textcolor{red!70!black}{-9.59} & 0.00 & \textcolor{red!70!black}{-5.22} & \textcolor{red!70!black}{-2.36} & \textcolor{green!50!black}{+1.56} & \textcolor{red!70!black}{-0.67} \\
 &  & easy & \textcolor{red!70!black}{-15.07} & \textcolor{red!70!black}{-16.40} & \textcolor{red!70!black}{-15.68} & \textcolor{green!50!black}{+1.18} & \textcolor{red!70!black}{-3.12} & \textcolor{red!70!black}{-0.67} \\
 &  & hard & \textcolor{red!70!black}{-24.65} & \textcolor{red!70!black}{-3.28} & \textcolor{red!70!black}{-14.92} & 0.00 & \textcolor{red!70!black}{-1.56} & \textcolor{red!70!black}{-0.67} \\
\addlinespace[0.4em]
\multirow{3}{*}{dynamic} &  & all & \textcolor{red!70!black}{-9.59} & \textcolor{red!70!black}{-8.19} & \textcolor{red!70!black}{-8.95} & \textcolor{green!50!black}{+1.18} & \textcolor{red!70!black}{-7.81} & \textcolor{red!70!black}{-2.69} \\
 &  & easy & \textcolor{red!70!black}{-16.44} & \textcolor{red!70!black}{-18.04} & \textcolor{red!70!black}{-17.16} & 0.00 & \textcolor{red!70!black}{-3.12} & \textcolor{red!70!black}{-1.34} \\
 &  & hard & \textcolor{red!70!black}{-19.17} & \textcolor{red!70!black}{-8.20} & \textcolor{red!70!black}{-14.18} & \textcolor{green!50!black}{+9.41} & \textcolor{red!70!black}{-1.56} & \textcolor{green!50!black}{+4.70} \\
\midrule
\multirow{3}{*}{static} & \multirow[c]{4}{*}{2} & all & \textcolor{red!70!black}{-12.33} & \textcolor{red!70!black}{-29.51} & \textcolor{red!70!black}{-20.15} & \textcolor{green!50!black}{+7.06} & \textcolor{green!50!black}{+3.12} & \textcolor{green!50!black}{+5.37} \\
 &  & easy & \textcolor{red!70!black}{-4.11} & \textcolor{red!70!black}{-24.59} & \textcolor{red!70!black}{-13.43} & \textcolor{green!50!black}{+3.53} & 0.00 & \textcolor{green!50!black}{+2.01} \\
 &  & hard & \textcolor{red!70!black}{-26.02} & \textcolor{red!70!black}{-14.75} & \textcolor{red!70!black}{-20.89} & \textcolor{green!50!black}{+3.53} & \textcolor{green!50!black}{+1.56} & \textcolor{green!50!black}{+2.68} \\
\addlinespace[0.4em]
\multirow{1}{*}{dynamic} &  & all & \textcolor{red!70!black}{-19.18} & \textcolor{red!70!black}{-13.12} & \textcolor{red!70!black}{-16.41} & \textcolor{green!50!black}{+8.23} & \textcolor{green!50!black}{+6.25} & \textcolor{green!50!black}{+7.39} \\
\midrule
\multirow{3}{*}{static} & \multirow[c]{6}{*}{4} & all & \textcolor{green!50!black}{+1.37} & \textcolor{red!70!black}{-6.55} & \textcolor{red!70!black}{-2.24} & \textcolor{green!50!black}{+3.53} & \textcolor{green!50!black}{+23.43} & \textcolor{green!50!black}{+12.08} \\
 &  & easy & \textcolor{red!70!black}{-8.22} & \textcolor{red!70!black}{-8.19} & \textcolor{red!70!black}{-8.21} & \textcolor{green!50!black}{+8.24} & 0.00 & \textcolor{green!50!black}{+4.70} \\
 &  & hard & \textcolor{green!50!black}{+15.06} & \textcolor{red!70!black}{-8.20} & \textcolor{green!50!black}{+4.47} & \textcolor{green!50!black}{+10.58} & \textcolor{green!50!black}{+17.19} & \textcolor{green!50!black}{+13.42} \\
\addlinespace[0.4em]
\multirow{3}{*}{dynamic} &  & all & \textcolor{red!70!black}{-1.37} & \textcolor{red!70!black}{-16.39} & \textcolor{red!70!black}{-8.21} & \textcolor{green!50!black}{+9.41} & \textcolor{green!50!black}{+21.87} & \textcolor{green!50!black}{+14.77} \\
 &  & easy & \textcolor{red!70!black}{-8.22} & \textcolor{red!70!black}{-22.95} & \textcolor{red!70!black}{-14.92} & \textcolor{red!70!black}{-3.53} & \textcolor{green!50!black}{+28.12} & \textcolor{green!50!black}{+10.07} \\
 &  & hard & \textcolor{green!50!black}{+5.48} & \textcolor{red!70!black}{-4.91} & \textcolor{green!50!black}{+0.75} & \textcolor{green!50!black}{+23.53} & \textcolor{red!70!black}{-9.38} & \textcolor{green!50!black}{+9.40} \\
\bottomrule
\end{tabular}%
\caption{\textbf{Difference between Qwen~2.5~7B~1M and Qwen~2.5~7B in zero-shot performance.} Each cell reports \textbf{Qwen~2.5~7B~1M $-$ Qwen~2.5~7B} for the corresponding retrieval configuration and evaluation split. Positive values indicate that the 1M-context model performs better; negative values indicate that the standard-context 7B model performs better.}
\label{tab:compare-qwens-7b}
\end{table*}

%% file: tables/zero_shot_appendix/zeroshot_llama31_8b.tex
\begin{table*}[!h]
\centering
\small
\setlength{\tabcolsep}{5pt}
\newsavebox{\zeroshotllamathirtyonebbox}
\sbox{\zeroshotllamathirtyonebbox}{%
\begin{tabular}{l c c c c c c c c}
\toprule
\multirow{2}{*}{\makecell{\oursraw\\type}} & \multirow{2}{*}{Top-$K$} & \multirow{2}{*}{Index} & \multicolumn{3}{c}{ALFWorld} & \multicolumn{3}{c}{ScienceWorld} \\
\cmidrule(lr){4-6} \cmidrule(lr){7-9}
& & & Easy tasks & Hard tasks & All tasks & Easy tasks & Hard tasks & All tasks \\
\midrule
No RAG & 0 & -- & 10.96 & 3.28 & 7.46 & 5.88 & 9.38 & 7.38 \\
\midrule
\multirow{3}{*}{static} & \multirow[c]{6}{*}{1} & all & 31.51 & 13.11 & 23.13 & 23.53 & 10.94 & 18.12 \\
 &  & easy & 34.25 & \textbf{4.92} & 20.90 & 23.53 & 1.56 & 14.09 \\
 &  & hard & 15.07 & 24.59 & 19.40 & 10.59 & 14.06 & 12.08 \\
\addlinespace[0.4em]
\multirow{3}{*}{dynamic} &  & all & 32.88{\scriptsize \textcolor{green!50!black}{\,$\uparrow 1.4$}} & 31.15{\scriptsize \textcolor{green!50!black}{\,$\uparrow 18.0$}} & 32.09{\scriptsize \textcolor{green!50!black}{\,$\uparrow 9.0$}} & -- & -- & -- \\
 &  & easy & 38.36{\scriptsize \textcolor{green!50!black}{\,$\uparrow 4.1$}} & 3.28{\scriptsize \textcolor{red!70!black}{\,$\downarrow 1.6$}} & 22.39{\scriptsize \textcolor{green!50!black}{\,$\uparrow 1.5$}} & 21.18{\scriptsize \textcolor{red!70!black}{\,$\downarrow 2.4$}} & \textbf{3.12}{\scriptsize \textcolor{green!50!black}{\,$\uparrow 1.6$}} & 13.42{\scriptsize \textcolor{red!70!black}{\,$\downarrow 0.7$}} \\
 &  & hard & \textbf{17.81}{\scriptsize \textcolor{green!50!black}{\,$\uparrow 2.7$}} & 29.51{\scriptsize \textcolor{green!50!black}{\,$\uparrow 4.9$}} & 23.13{\scriptsize \textcolor{green!50!black}{\,$\uparrow 3.7$}} & 9.41{\scriptsize \textcolor{red!70!black}{\,$\downarrow 1.2$}} & 14.06{\scriptsize \,0.0} & 11.41{\scriptsize \textcolor{red!70!black}{\,$\downarrow 0.7$}} \\
\midrule
\multirow{3}{*}{static} & \multirow[c]{4}{*}{2} & all & 43.84 & 37.70 & 41.04 & 28.24 & \textbf{17.19} & 23.49 \\
 &  & easy & 35.62 & 1.64 & 20.15 & 23.53 & \textbf{4.69} & 15.44 \\
 &  & hard & \textbf{23.29} & 34.43 & 28.36 & \textbf{12.94} & \textbf{20.31} & \textbf{16.11} \\
\addlinespace[0.4em]
\multirow{1}{*}{dynamic} &  & all & 45.21{\scriptsize \textcolor{green!50!black}{\,$\uparrow 1.4$}} & 31.15{\scriptsize \textcolor{red!70!black}{\,$\downarrow 6.6$}} & 38.81{\scriptsize \textcolor{red!70!black}{\,$\downarrow 2.2$}} & 29.41{\scriptsize \textcolor{green!50!black}{\,$\uparrow 1.2$}} & 21.88{\scriptsize \textcolor{green!50!black}{\,$\uparrow 4.7$}} & 26.17{\scriptsize \textcolor{green!50!black}{\,$\uparrow 2.7$}} \\
\midrule
\multirow{3}{*}{static} & \multirow[c]{6}{*}{4} & all & \textbf{54.79} & \textbf{39.34} & \textbf{47.76} & \textbf{35.29} & 14.06 & \textbf{26.17} \\
 &  & easy & \textbf{54.79} & 3.28 & \textbf{31.34} & \textbf{29.41} & 0.00 & \textbf{16.78} \\
 &  & hard & 20.55 & \textbf{40.98} & \textbf{29.85} & 10.59 & 17.19 & 13.42 \\
\addlinespace[0.4em]
\multirow{3}{*}{dynamic} &  & all & \textbf{50.68}{\scriptsize \textcolor{red!70!black}{\,$\downarrow 4.1$}} & \textbf{44.26}{\scriptsize \textcolor{green!50!black}{\,$\uparrow 4.9$}} & \textbf{47.76}{\scriptsize \,0.0} & \textbf{30.59}{\scriptsize \textcolor{red!70!black}{\,$\downarrow 4.7$}} & \textbf{23.44}{\scriptsize \textcolor{green!50!black}{\,$\uparrow 9.4$}} & \textbf{27.52}{\scriptsize \textcolor{green!50!black}{\,$\uparrow 1.4$}} \\
 &  & easy & \textbf{45.21}{\scriptsize \textcolor{red!70!black}{\,$\downarrow 9.6$}} & \textbf{8.20}{\scriptsize \textcolor{green!50!black}{\,$\uparrow 4.9$}} & \textbf{28.36}{\scriptsize \textcolor{red!70!black}{\,$\downarrow 3.0$}} & \textbf{31.76}{\scriptsize \textcolor{green!50!black}{\,$\uparrow 2.4$}} & 0.00{\scriptsize \,0.0} & \textbf{18.12}{\scriptsize \textcolor{green!50!black}{\,$\uparrow 1.3$}} \\
 &  & hard & \textbf{17.81}{\scriptsize \textcolor{red!70!black}{\,$\downarrow 2.7$}} & \textbf{37.70}{\scriptsize \textcolor{red!70!black}{\,$\downarrow 3.3$}} & \textbf{26.87}{\scriptsize \textcolor{red!70!black}{\,$\downarrow 3.0$}} & \textbf{16.47}{\scriptsize \textcolor{green!50!black}{\,$\uparrow 5.9$}} & \textbf{18.75}{\scriptsize \textcolor{green!50!black}{\,$\uparrow 1.6$}} & \textbf{17.45}{\scriptsize \textcolor{green!50!black}{\,$\uparrow 4.0$}} \\
\bottomrule
\end{tabular}
}
\ACLorArxivorOther{
    \usebox{\zeroshotllamathirtyonebbox}
}{
    \usebox{\zeroshotllamathirtyonebbox}
}{
    \resizebox{\linewidth}{!}{\usebox{\zeroshotllamathirtyonebbox}}
}
\caption{\textbf{Results for \ours inference without training across ALFWorld and ScienceWorld for Llama~3.1~8B.} The \textbf{bold} values mark the best top-$k$ setting for each (\ours type, index) within a given dataset split. For dynamic retrieval rows, the superscript arrows report the per-cell difference against the corresponding static row with the same index and top-$k$: \textcolor{green!50!black}{$\uparrow$} indicates improvement, \textcolor{red!70!black}{$\downarrow$} indicates a decline, and black values denote ties.}
\label{tab:zero-shot-llama}
\end{table*}

%% file: tables/zero_shot_appendix/zeroshot_llama32_3b.tex
\begin{table*}[!h]
\centering
\small
\setlength{\tabcolsep}{5pt}
\newsavebox{\zeroshotllamathirtytwothreebbox}
\sbox{\zeroshotllamathirtytwothreebbox}{%
\begin{tabular}{l c c c c c c c c}
\toprule
\multirow{2}{*}{\makecell{\oursraw\\type}} & \multirow{2}{*}{Top-$K$} & \multirow{2}{*}{Index} & \multicolumn{3}{c}{ALFWorld} & \multicolumn{3}{c}{ScienceWorld} \\
\cmidrule(lr){4-6} \cmidrule(lr){7-9}
& & & Easy tasks & Hard tasks & All tasks & Easy tasks & Hard tasks & All tasks \\
\midrule
No RAG & 0 & -- & 1.37 & 0.00 & 0.75 & 3.53 & 0.00 & 2.01 \\
\midrule
\multirow{3}{*}{static} & \multirow[c]{6}{*}{1} & all & 8.22 & 3.28 & 5.97 & 15.29 & 10.94 & 13.42 \\
 &  & easy & 9.59 & \textbf{0.00} & 5.22 & 14.12 & \textbf{3.12} & \textbf{9.40} \\
 &  & hard & 4.11 & 3.28 & 3.73 & 8.24 & 9.38 & 8.72 \\
\addlinespace[0.4em]
\multirow{3}{*}{dynamic} &  & all & 10.96{\scriptsize \textcolor{green!50!black}{\,$\uparrow 2.7$}} & \textbf{3.28}{\scriptsize \,0.0} & 7.46{\scriptsize \textcolor{green!50!black}{\,$\uparrow 1.5$}} & 15.29{\scriptsize \,0.0} & 7.81{\scriptsize \textcolor{red!70!black}{\,$\downarrow 3.1$}} & 12.08{\scriptsize \textcolor{red!70!black}{\,$\downarrow 1.3$}} \\
 &  & easy & 10.96{\scriptsize \textcolor{green!50!black}{\,$\uparrow 1.4$}} & \textbf{1.64}{\scriptsize \textcolor{green!50!black}{\,$\uparrow 1.6$}} & 6.72{\scriptsize \textcolor{green!50!black}{\,$\uparrow 1.5$}} & 9.41{\scriptsize \textcolor{red!70!black}{\,$\downarrow 4.7$}} & \textbf{3.12}{\scriptsize \,0.0} & 6.71{\scriptsize \textcolor{red!70!black}{\,$\downarrow 2.7$}} \\
 &  & hard & \textbf{15.07}{\scriptsize \textcolor{green!50!black}{\,$\uparrow 11.0$}} & 0.00{\scriptsize \textcolor{red!70!black}{\,$\downarrow 3.3$}} & \textbf{8.21}{\scriptsize \textcolor{green!50!black}{\,$\uparrow 4.5$}} & \textbf{9.41}{\scriptsize \textcolor{green!50!black}{\,$\uparrow 1.2$}} & 4.69{\scriptsize \textcolor{red!70!black}{\,$\downarrow 4.7$}} & 7.38{\scriptsize \textcolor{red!70!black}{\,$\downarrow 1.3$}} \\
\midrule
\multirow{3}{*}{static} & \multirow[c]{4}{*}{2} & all & \textbf{15.07} & \textbf{9.84} & \textbf{12.69} & 9.41 & 14.06 & 11.41 \\
 &  & easy & \textbf{10.96} & \textbf{0.00} & \textbf{5.97} & 10.59 & 0.00 & 6.04 \\
 &  & hard & \textbf{9.59} & \textbf{6.56} & \textbf{8.21} & 5.88 & \textbf{15.62} & 10.07 \\
\addlinespace[0.4em]
\multirow{1}{*}{dynamic} &  & all & 6.85{\scriptsize \textcolor{red!70!black}{\,$\downarrow 8.2$}} & 1.64{\scriptsize \textcolor{red!70!black}{\,$\downarrow 8.2$}} & 4.48{\scriptsize \textcolor{red!70!black}{\,$\downarrow 8.2$}} & 14.12{\scriptsize \textcolor{green!50!black}{\,$\uparrow 4.7$}} & \textbf{17.19}{\scriptsize \textcolor{green!50!black}{\,$\uparrow 3.1$}} & 15.44{\scriptsize \textcolor{green!50!black}{\,$\uparrow 4.0$}} \\
\midrule
\multirow{3}{*}{static} & \multirow[c]{6}{*}{4} & all & 12.33 & 4.92 & 8.96 & \textbf{22.35} & \textbf{15.62} & \textbf{19.46} \\
 &  & easy & -- & -- & -- & \textbf{16.47} & 0.00 & \textbf{9.40} \\
 &  & hard & 6.85 & 3.28 & 5.22 & \textbf{9.41} & 12.50 & \textbf{10.74} \\
\addlinespace[0.4em]
\multirow{3}{*}{dynamic} &  & all & \textbf{12.33}{\scriptsize \,0.0} & \textbf{3.28}{\scriptsize \textcolor{red!70!black}{\,$\downarrow 1.6$}} & \textbf{8.21}{\scriptsize \textcolor{red!70!black}{\,$\downarrow 0.8$}} & \textbf{24.71}{\scriptsize \textcolor{green!50!black}{\,$\uparrow 2.4$}} & 7.81{\scriptsize \textcolor{red!70!black}{\,$\downarrow 7.8$}} & \textbf{17.45}{\scriptsize \textcolor{red!70!black}{\,$\downarrow 2.0$}} \\
 &  & easy & \textbf{21.92} & 0.00 & \textbf{11.94} & \textbf{22.35}{\scriptsize \textcolor{green!50!black}{\,$\uparrow 5.9$}} & 1.56{\scriptsize \textcolor{green!50!black}{\,$\uparrow 1.6$}} & \textbf{13.42}{\scriptsize \textcolor{green!50!black}{\,$\uparrow 4.0$}} \\
 &  & hard & 6.85{\scriptsize \,0.0} & \textbf{1.64}{\scriptsize \textcolor{red!70!black}{\,$\downarrow 1.6$}} & 4.48{\scriptsize \textcolor{red!70!black}{\,$\downarrow 0.7$}} & 7.06{\scriptsize \textcolor{red!70!black}{\,$\downarrow 2.4$}} & \textbf{10.94}{\scriptsize \textcolor{red!70!black}{\,$\downarrow 1.6$}} & \textbf{8.72}{\scriptsize \textcolor{red!70!black}{\,$\downarrow 2.0$}} \\
\bottomrule
\end{tabular}
}
\ACLorArxivorOther{
    \usebox{\zeroshotllamathirtytwothreebbox}
}{
    \usebox{\zeroshotllamathirtytwothreebbox}
}{
    \resizebox{\linewidth}{!}{\usebox{\zeroshotllamathirtytwothreebbox}}
}
\caption{\textbf{Results for \ours inference without training across ALFWorld and ScienceWorld for Llama~3.2~3B.} The \textbf{bold} values mark the best top-$k$ setting for each (\ours type, index) within a given dataset split. For dynamic retrieval rows, the superscript arrows report the per-cell difference against the corresponding static row with the same index and top-$k$: \textcolor{green!50!black}{$\uparrow$} indicates improvement, \textcolor{red!70!black}{$\downarrow$} indicates a decline, and black values denote ties.}
\label{tab:zero-shot-llama32-3b}
\end{table*}

%% file: figures/appendix/traj_format.tex
\begin{figure}[!htbp]
    \begin{tcolorbox}[title=chat JSON, colback=gray!5, boxrule=0.6pt,
  left=0pt, right=0pt, top=6pt, bottom=6pt,width=\textwidth]
\ttfamily\tiny
\begin{lstlisting}
        {
          "role": "assistant",
          "content": "look"
        },
        {
          "role": "user",
           "content": "You are in the middle of a room. Looking quickly around you, you see nothing."
        },
    \end{lstlisting}
\end{tcolorbox}
    \begin{tcolorbox}[title=agentic JSON, colback=gray!5, boxrule=0.6pt,
  left=0pt, right=0pt, top=6pt, bottom=6pt,width=\textwidth]
\ttfamily\tiny
\begin{lstlisting}
        {
        "role": "action",
         "content": "look"
        },
        {
         "role": "observation",
          "content": "You are in the middle of a room. Looking quickly around you, you see nothing."
        },
    \end{lstlisting}
\end{tcolorbox}
\begin{tcolorbox}[title=compact JSON, colback=gray!5, boxrule=0.6pt,
  left=0pt, right=0pt, top=6pt, bottom=6pt,width=\textwidth]
\ttfamily\tiny
\begin{lstlisting}
        {
         "action": "look"
        },
        {
         "observation": "You are in the middle of a room. Looking quickly around you, you see nothing."
        },
    \end{lstlisting}
\end{tcolorbox}
\begin{tcolorbox}[title=textual, colback=gray!5, boxrule=0.6pt,
  left=0pt, right=0pt, top=6pt, bottom=6pt,width=\textwidth]
\ttfamily\tiny
    Assistant:  look\\
    User:  You are in the middle of a room. Looking quickly around you, you see nothing.
    
\end{tcolorbox}
 \caption{Different trajectory formats: \textit{chat JSON}, \textit{agentic JSON}, \textit{compact JSON}, and \textit{textual}.}
 \label{fig:traj_format}
\end{figure}

%% file: tables/zero_shot_appendix/prompt_formatting.tex
\begin{table*}[!h]
\centering
\setlength{\tabcolsep}{3.5pt}
\renewcommand{\arraystretch}{1.12}
\resizebox{0.8\textwidth}{!}{
\begin{tabular}{l c c c c | c c c c}
\hline
& \multicolumn{4}{c}{ALFWorld} & \multicolumn{4}{c}{ScienceWorld} \\
\cline{2-5}\cline{6-9}
Format & \makecell{Ministral\\3-8B} & \makecell{Gemma\\3-4B} & \makecell{Qwen\\2.5-7B} & \makecell{Qwen\\2.5-7B-1M} & \makecell{Ministral\\3-8B} & \makecell{Gemma\\3-4B} & \makecell{Qwen\\2.5-7B} & \makecell{Qwen\\2.5-7B-1M} \\
\hline
Zero-shot & 4.5 & 0.8 & 29.9 & 5.2 & 10.4 & 2.0 & 2.6 & 3.4 \\
\hline
\multicolumn{9}{c}{\textit{Top-1 trajectories}} \\
\hline
Chat JSON & 39.3 & \textbf{12.7} & \textbf{67.2} & 61.9 & 26.2 & 3.4 & \textbf{12.1} & \textbf{11.4} \\
Agentic JSON & 39.6 & \textbf{12.7} & 61.9 & \textbf{64.2} & 27.5 & 0.7 & 4.0 & 8.7 \\
Compact JSON & 41.8 & 8.2 & 61.2 & 62.0 & 22.2 & 4.7 & 6.0 & 7.4 \\
Textual & \textbf{43.3} & 11.9 & 41.8 & 55.2 & \textbf{33.6} & \textbf{6.7} & 0.7 & 8.1 \\
\hline
\multicolumn{9}{c}{\textit{Top-2 trajectories}} \\
\hline
Chat JSON & 51.0 & 11.9 & \textbf{82.8} & 62.7 & 34.7 & 5.1 & \textbf{10.7} & \textbf{17.5} \\
Agentic JSON & 48.5 & 10.5 & 73.9 & 64.2 & \textbf{36.2} & 2.0 & 1.3 & 12.1 \\
Compact JSON & 49.3 & \textbf{14.9} & 79.1 & \textbf{66.4} & 30.9 & 5.4 & 9.4 & 16.8 \\
Textual & \textbf{52.2} & 12.7 & 14.9 & 45.5 & 32.9 & \textbf{16.1} & 1.3 & 5.4 \\
\hline
\multicolumn{9}{c}{\textit{Top-4 trajectories}} \\
\hline
Chat JSON & \textbf{64.2} & \textbf{20.9} & \textbf{83.6} & \textbf{81.3} & \textbf{34.5} & 8.1 & 7.4 & \textbf{23.5} \\
Agentic JSON & 56.0 & 19.4 & 81.3 & 74.6 & 30.9 & 3.4 & 0.7 & 14.1 \\
Compact JSON & 59.0 & 20.2 & 79.1 & 78.4 & 29.5 & 9.4 & \textbf{10.7} & 22.8 \\
Textual & 48.5 & 19.4 & 1.5 & 51.5 & 30.9 & \textbf{14.1} & 0.7 & 7.4 \\
\hline
\multicolumn{9}{c}{\textit{Top-5 trajectories}} \\
\hline
Chat JSON & \textbf{67.2} & \textbf{19.4} & 78.4 & 76.1 & 31.5 & 8.7 & \textbf{10.7} & \textbf{24.8} \\
Agentic JSON & 61.2 & 12.7 & \textbf{83.6} & 75.4 & 34.1 & 3.5 & 0.7 & 5.4 \\
Compact JSON & 59.7 & 14.2 & 70.9 & \textbf{81.3} & \textbf{38.3} & 7.4 & 8.7 & 24.2 \\
Textual & 44.0 & 18.7 & 0.0 & 75.4 & 29.5 & \textbf{10.7} & 1.3 & 3.4 \\
\hline
\end{tabular}%
}
\caption{Impact of trajectory formatting on \ours performance (all tasks, official valid unseen split)}
\label{tab:prompt_formatting}
\end{table*}

%% file: appendix/longer_training.tex
\section{Generalization Dynamics: Agents keep improving when trained longer.}
\label{app:longer-training}

During preliminary experiments, we found that longer fine-tuning can yield substantially stronger downstream agent performance. Across ALFWorld and ScienceWorld, and for both LoRA and \oursraw-LoRA, task success often continues to improve after validation loss has reached its minimum and started increasing, resulting in a weak correspondence between validation loss and rollout performance. This occurs both for unseen instances of task groups observed during training (e.g., train/easy $\rightarrow$ valid-unseen/easy) and for held-out task groups (e.g., train/easy $\rightarrow$ valid-unseen/hard), including task types not observed during fine-tuning.

Motivated by this observation, we train for up to 50 epochs and evaluate multiple checkpoints rather than selecting models solely by validation-loss minima. Out-of-distribution performance often continues improving well beyond 10 epochs and, in several settings, reaches its best values near the end of the 50-epoch runs despite rising validation loss. Below, we report these dynamics in detail.

\subsection{Experimental setting.}
We fine-tune Llama-3.1-8B-Instruct with LoRA adapters on \textbf{easy} tasks only (as defined in Section~\ref{sec:experiments:dataset-and-benchmarks}).
Training follows the same supervised setup as the main fine-tuning experiments (Appendix~\ref{app:implementation-details:finetuning}): trajectories are formatted as multi-turn chats and we compute the cross-entropy loss on assistant tokens only, with greedy decoding at inference time (temperature \(=0\); Table~\ref{tab:hyperparameters}). We train for up to \(50\) epochs and evaluate a set of checkpoints throughout training (epochs \(1\) to \(50\)), every 5 epochs.

\paragraph{Methods and evaluation.}
We report curves for (i) \textbf{LoRA (no retrieval)} and (ii) \textbf{\oursraw-LoRA}, which performs retrieval-augmented fine-tuning (i.e., the \ours memory block is included in the training context) and uses the same retrieval pipeline at inference time.
For each checkpoint, we report:
(a) the \textbf{validation loss} (blue line; left \(y\)-axis), computed on held-in validation trajectories, and
(b) \textbf{agent performance} (right \(y\)-axis) obtained by executing the agent in the environment, measured as success rate (green) and average episode score (orange).
We evaluate both \textbf{in-distribution} generalization to unseen instances of the \textbf{easy} task groups (labeled \emph{ind} / cross-scene) and \textbf{out-of-distribution} generalization to \textbf{hard} task groups not seen during training (labeled \emph{ood} / cross-task). For \oursraw-LoRA we use a \emph{matched} index: easy evaluations retrieve from an index built on easy-task training trajectories, and hard evaluations retrieve from a hard-task training index, consistent with the protocol described in Section~\ref{sec:experiments:finetuning}.

\subsection{Results on ALFWorld.}
\label{sec:finetuning:grokking}
\begin{figure}[ht!]
  \centering
  \begin{subfigure}[t]{0.45\textwidth}
    \centering
    \includegraphics[width=\linewidth]{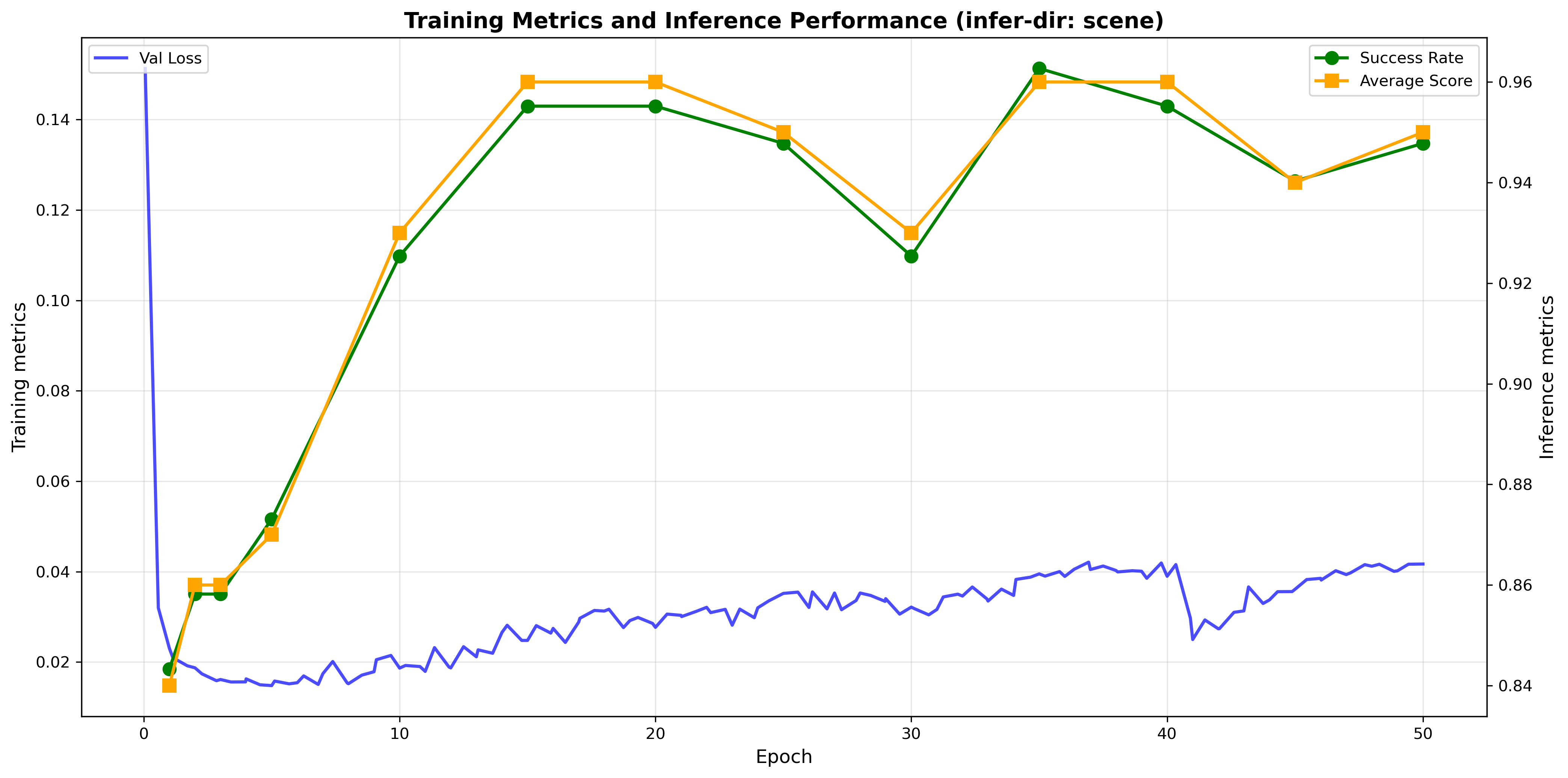}
    \caption{LoRA (no retrieval), ind}
    \label{fig:alf_cross_scene-lora}
  \end{subfigure}
  \hfill
  \begin{subfigure}[t]{0.45\textwidth}
    \centering
    \includegraphics[width=\linewidth]{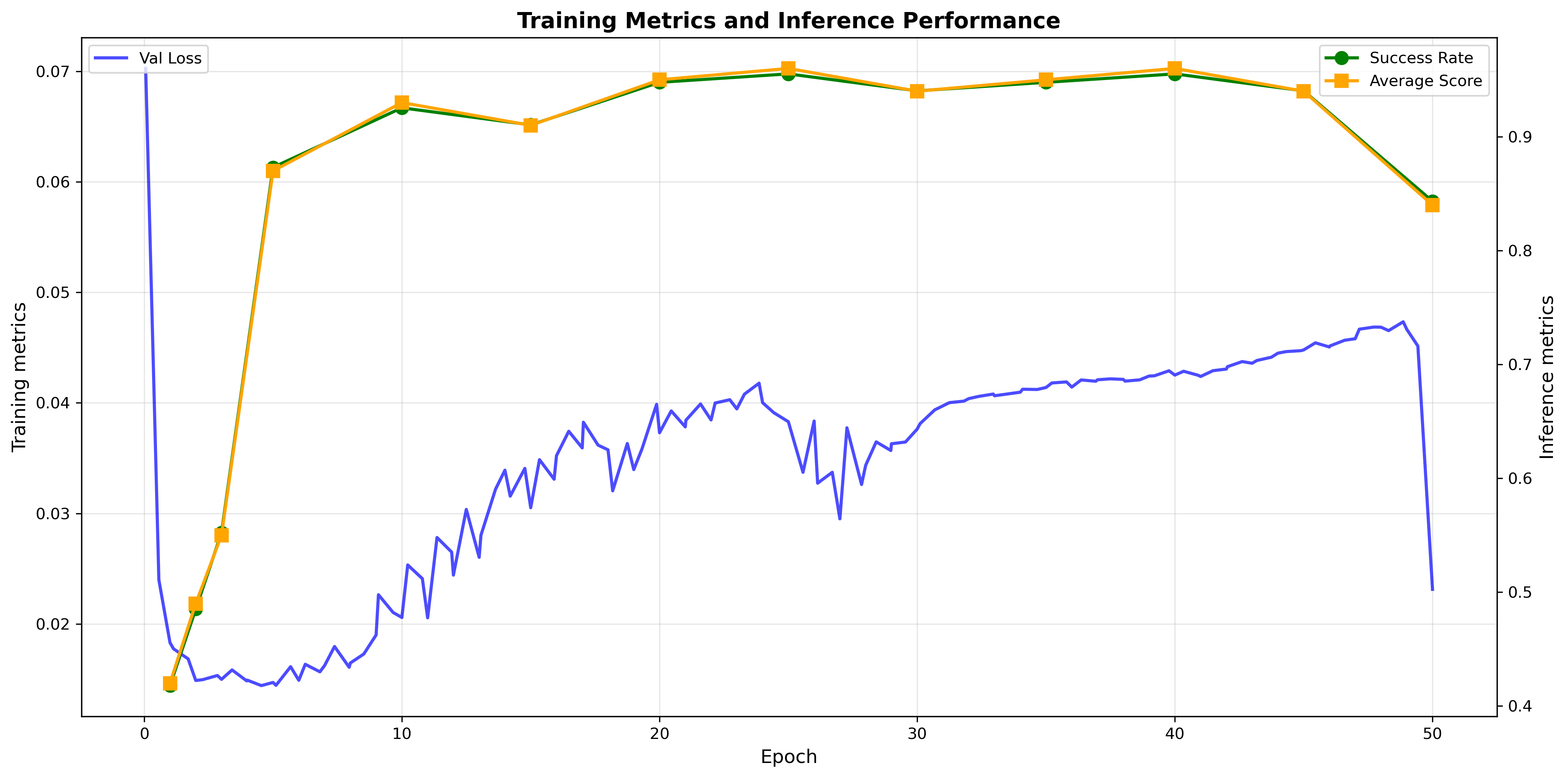}
    \caption{\oursraw-LoRA (matched index), ind}
    \label{fig:alf_cross_scene-lorag}
  \end{subfigure}
  \hfill
  \begin{subfigure}[t]{0.45\textwidth}
    \centering
    \includegraphics[width=\linewidth]{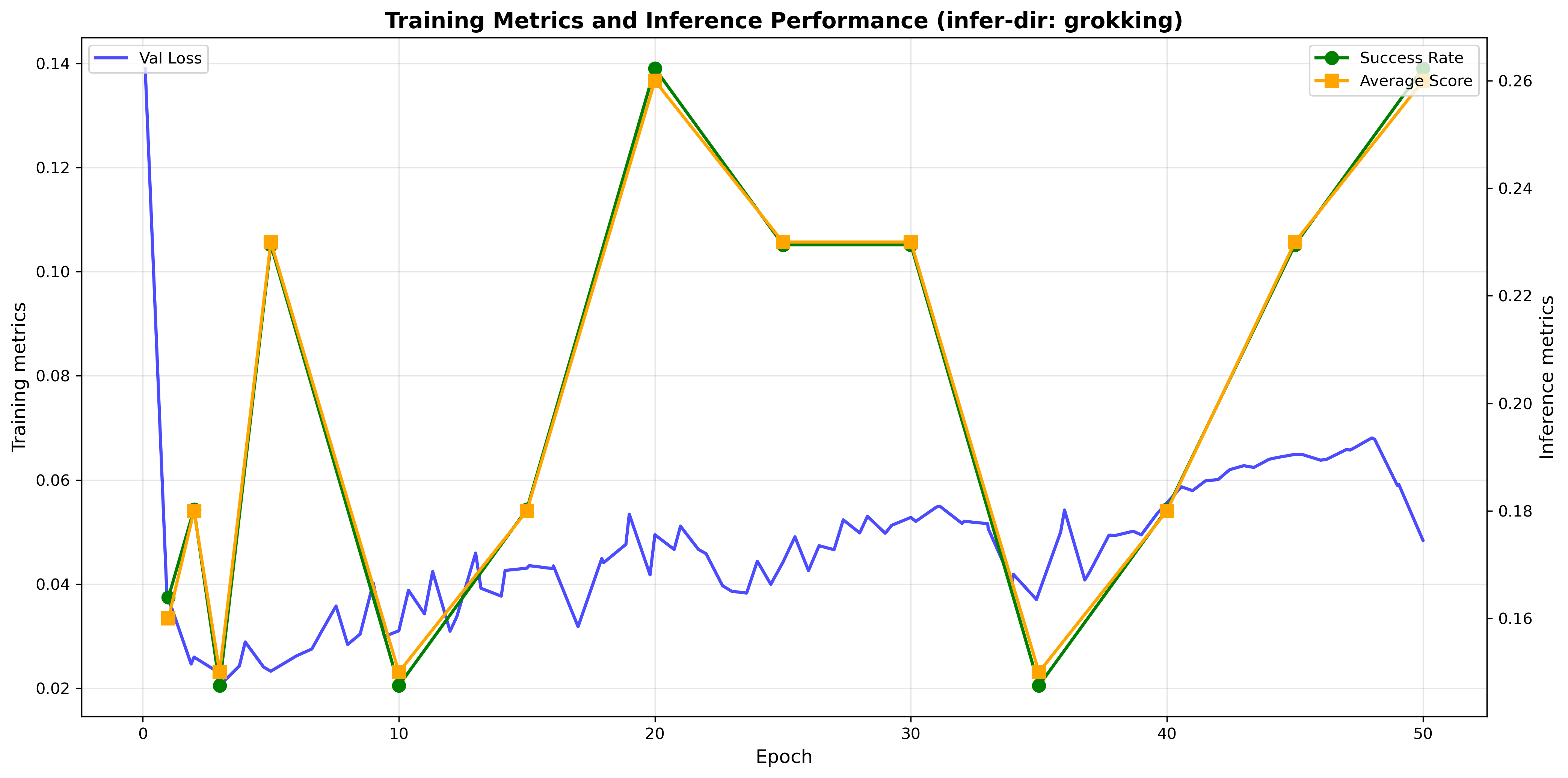}
    \caption{LoRA (no retrieval), ood}
    \label{fig:alf_cross_task-lora}
  \end{subfigure}
  \hfill
  \begin{subfigure}[t]{0.45\textwidth}
    \centering
    \includegraphics[width=\linewidth]{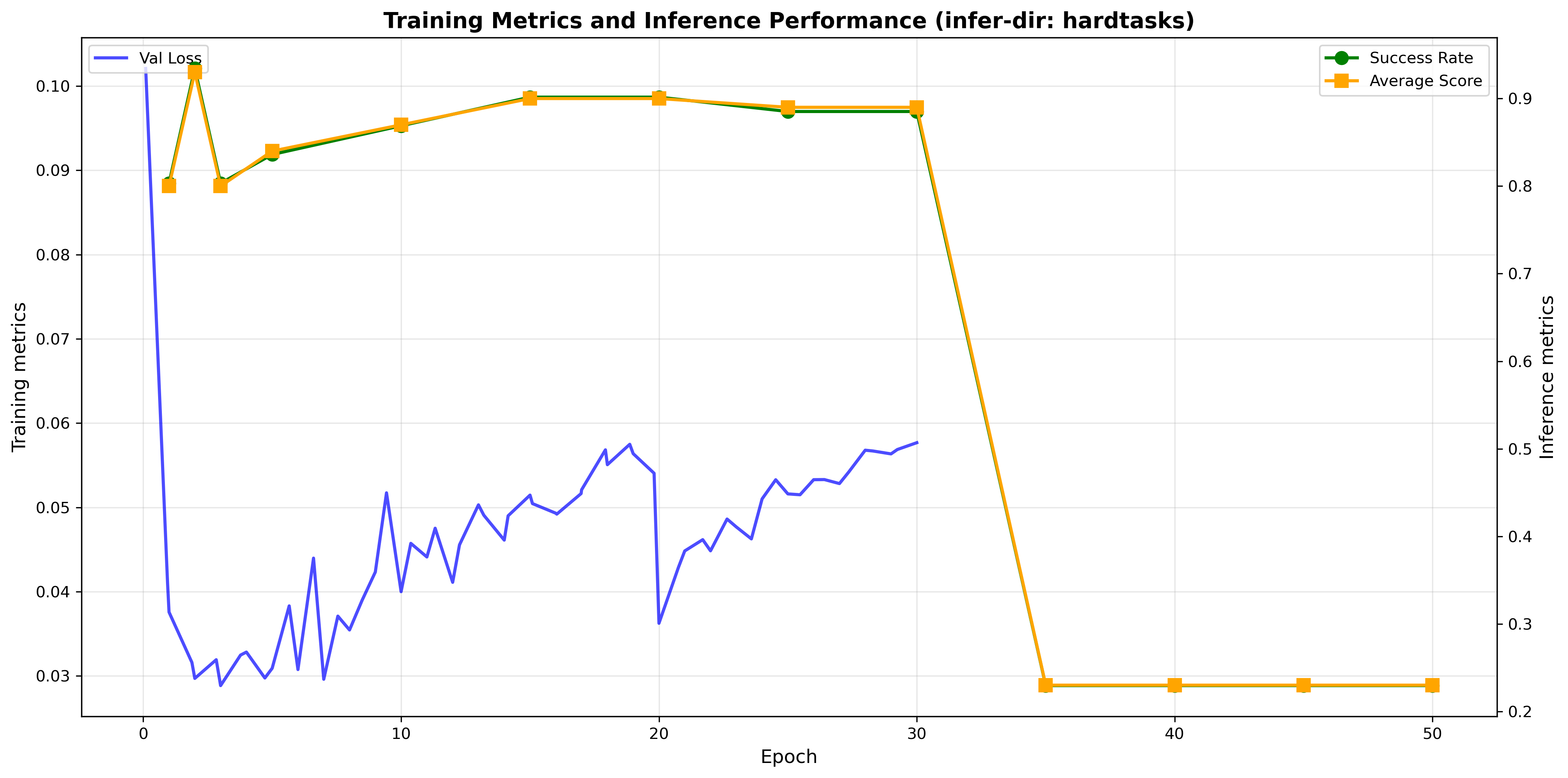}
    \caption{\oursraw-LoRA (matched index), ood}
    \label{fig:alf_cross_task-lorag}
  \end{subfigure}
  \caption{\textbf{Longer fine-tuning can improve generalization despite rising validation loss.} Comparison of validation loss and inference performance with respect to number of training epochs on ALFWorld for Llama-3.1-8B. Blue: validation cross-entropy (left axis; lower is better). Green/orange: rollout success rate and average episode score (right axis; higher is better) evaluated at multiple checkpoints during 50-epoch fine-tuning. Top: easy$\rightarrow$easy (in-distribution). Bottom: easy$\rightarrow$hard (out-of-distribution). For \oursraw-LoRA, retrieval uses a matched index for each evaluation split.}

\label{fig:alf_longer_training}
\end{figure}

Figure~\ref{fig:alf_longer_training} shows that validation loss typically bottoms out within the first few epochs and then increases steadily, while both in-distribution performance (Figures~\ref{fig:alf_cross_scene-lora} and~\ref{fig:alf_cross_scene-lorag}) and out-of-distribution performance (Figures~\ref{fig:alf_cross_task-lora} and~\ref{fig:alf_cross_task-lorag}) can continue improving much later into training.
This yields a weak (and sometimes negative) correspondence between validation loss and agent success, illustrating why early stopping based solely on validation loss can miss the best-performing checkpoints on unseen tasks.
We also observe that the extended-training regime can be non-monotonic, with occasional late-training instabilities for some settings (e.g., sharp drops for certain checkpoints), further motivating reporting multiple checkpoints rather than selecting by loss alone.

\subsection{Results on ScienceWorld.}
\begin{figure}[ht!]
  \centering
  \begin{subfigure}[t]{0.45\textwidth}
    \centering
    \includegraphics[width=\linewidth]{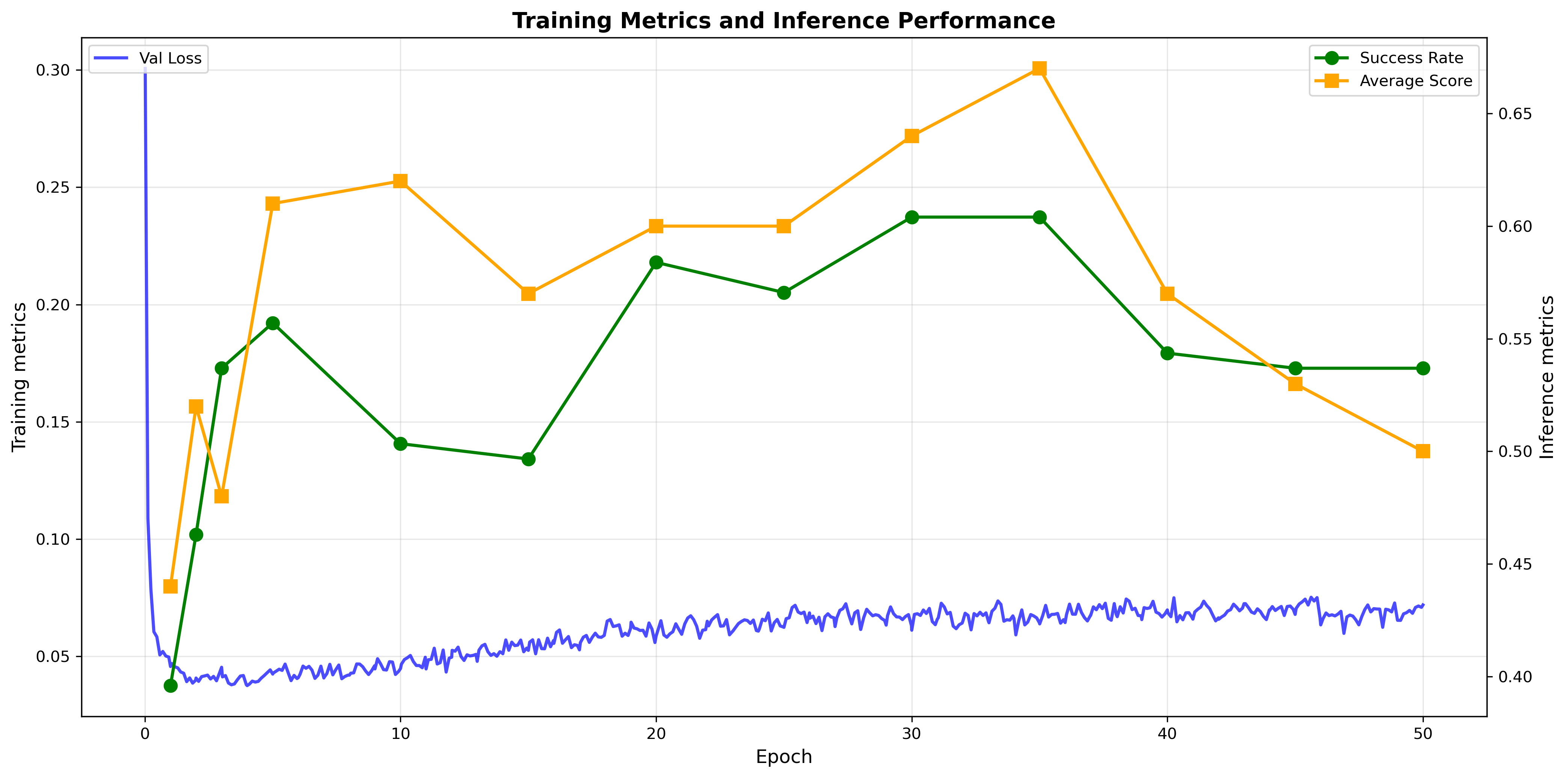}
    \caption{LoRA (no retrieval), ind}
    \label{fig:sci_cross_scene-lora}
  \end{subfigure}
  \hfill
  \begin{subfigure}[t]{0.45\textwidth}
    \centering
    \includegraphics[width=\linewidth]{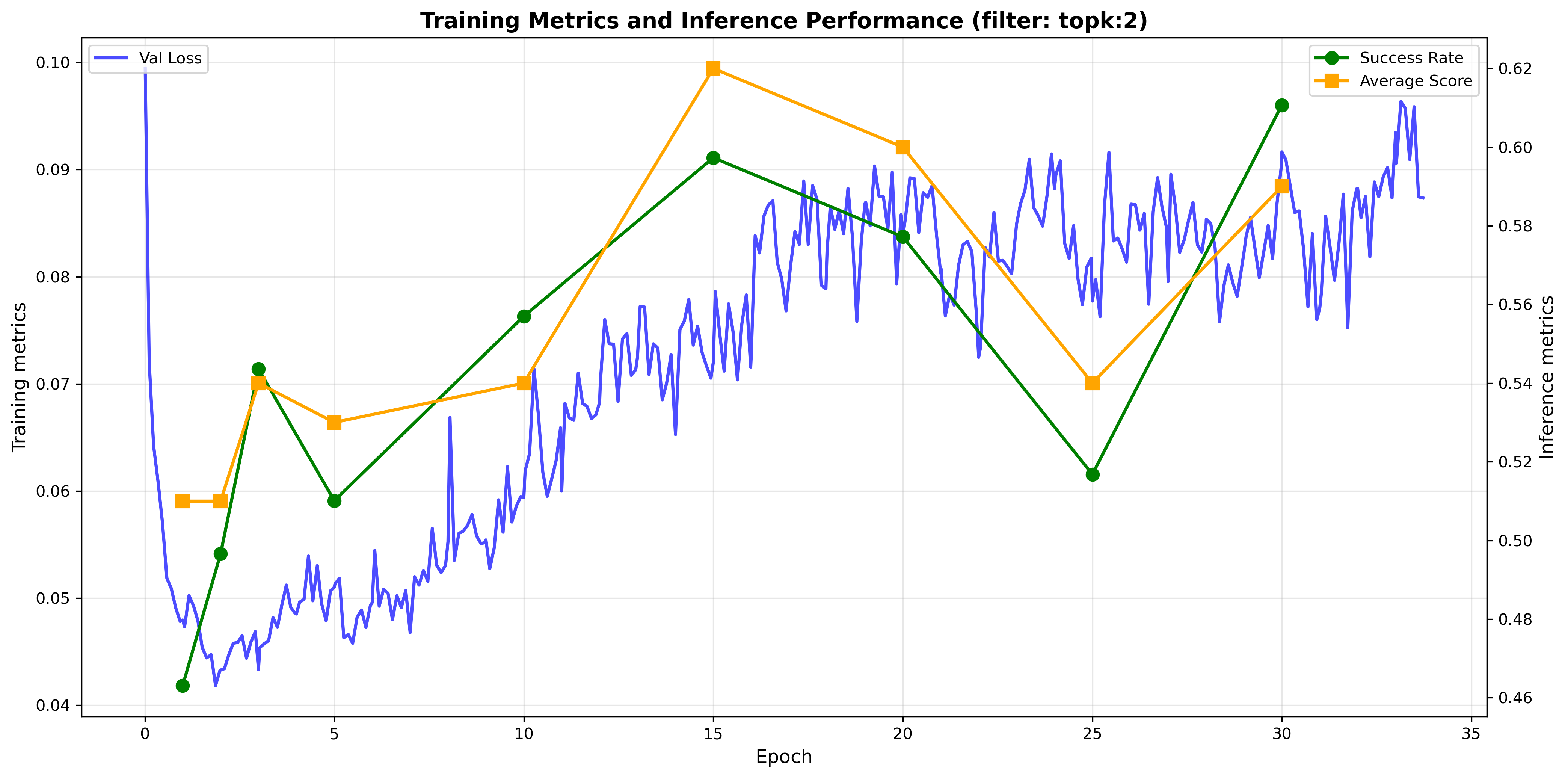}
    \caption{\oursraw-LoRA (matched index), ind}
    \label{fig:sci_cross_scene-lorag}
  \end{subfigure}
  \hfill
  \begin{subfigure}[t]{0.45\textwidth}
    \centering
    \includegraphics[width=\linewidth]{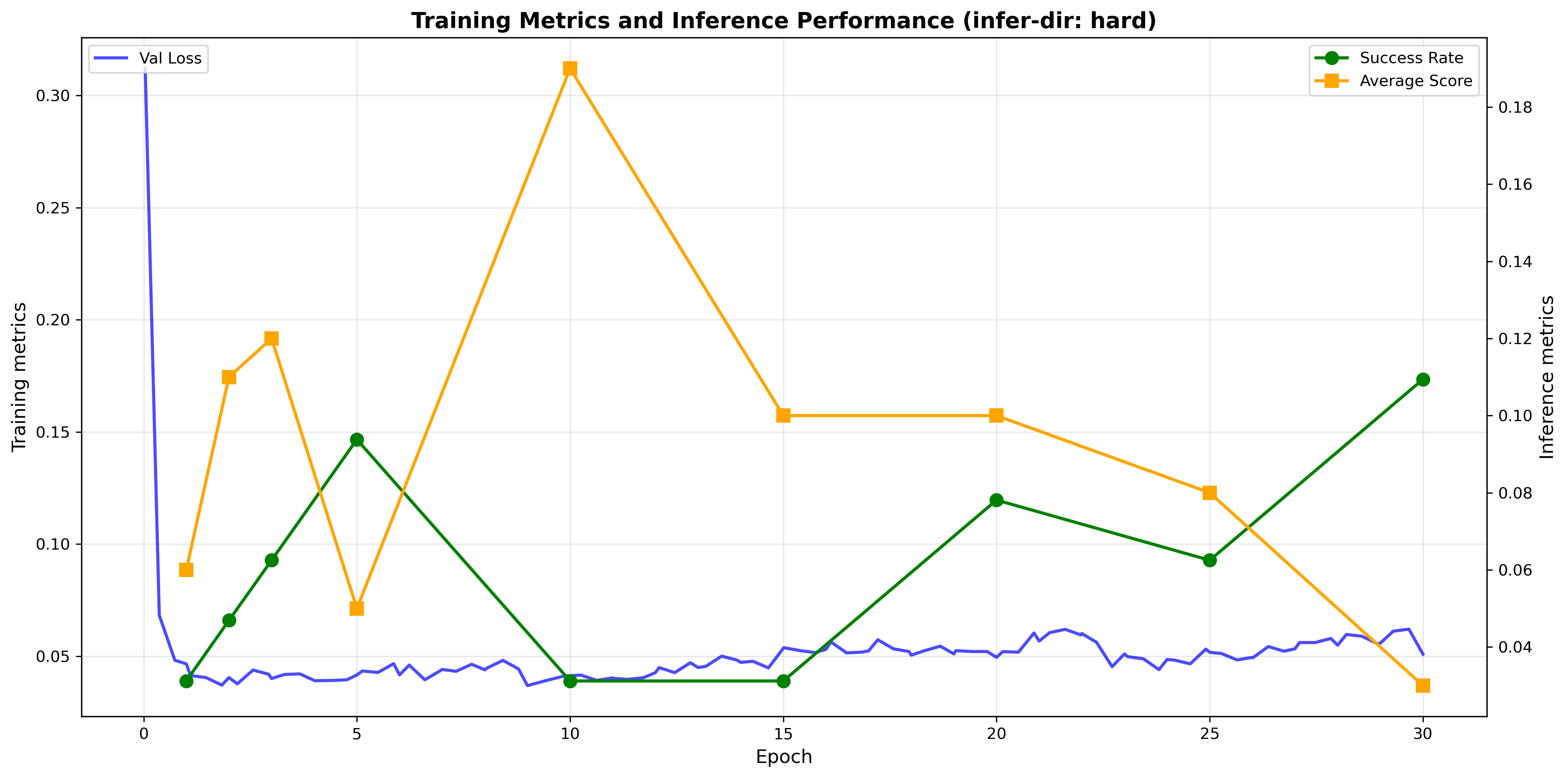}
    \caption{LoRA (no retrieval), ood}
    \label{fig:sci_cross_task-lora}
  \end{subfigure}
  \hfill
  \begin{subfigure}[t]{0.45\textwidth}
    \centering
    \includegraphics[width=\linewidth]{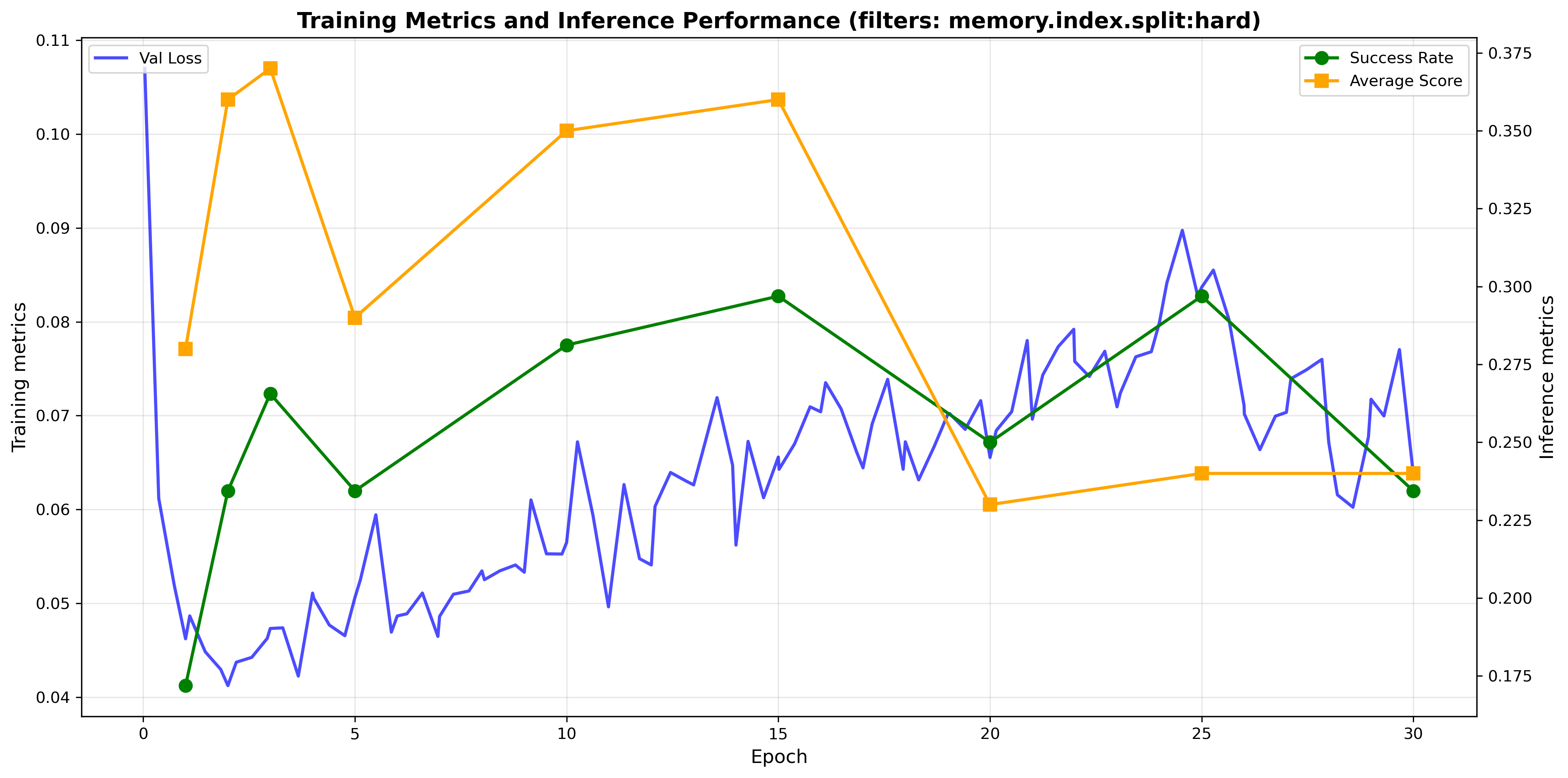}
    \caption{\oursraw-LoRA (matched index), ood}
    \label{fig:sci_cross_task-lorag}
  \end{subfigure}
  \caption{\textbf{Longer fine-tuning can improve generalization despite rising validation loss.} Comparison of validation loss and inference performance with respect to number of training epochs on ScienceWorld for Llama-3.1-8B. Blue: validation cross-entropy (left axis). Green/orange: rollout success rate and average episode score (right axis) across checkpoints during 50-epoch fine-tuning. Top: easy$\rightarrow$easy (in-distribution). Bottom: easy$\rightarrow$hard (out-of-distribution). For \oursraw-LoRA, retrieval uses a matched index for each evaluation split.}

\label{fig:sci_longer_training}
\end{figure}

We observe the same qualitative pattern in ScienceWorld (Figure~\ref{fig:sci_longer_training}). Despite validation loss increasing after its early minimum, success on both easy (Figures~\ref{fig:sci_cross_scene-lora} and~\ref{fig:sci_cross_scene-lorag}) and hard tasks (Figures~\ref{fig:sci_cross_task-lora} and~\ref{fig:sci_cross_task-lorag}) often improves at later epochs, with best checkpoints sometimes appearing near the end of training.
Overall, these curves support the main-text conclusion that longer fine-tuning can materially improve generalization, and that validation loss alone is an unreliable proxy for downstream agent success in this setting.

%% file: appendix/statistical_reliability.tex
\section{Statistical Reliability Analysis}
\label{app:statistical-reliability}

We report two complementary forms of uncertainty throughout the paper. When multiple training or inference seeds are available, we report variation across seeds. For paired method comparisons on the same evaluation instances, we additionally use task-level paired tests, which account for the fact that two methods are evaluated on identical tasks rather than on independent samples.

\paragraph{Paired task-level analysis.}
For each pair of methods, we convert every evaluated task instance into a binary success/failure outcome for both methods. We then compute (i) a paired bootstrap confidence interval over the difference in success rate, using \(10{,}000\) bootstrap resamples and a \(95\%\) interval, and (ii) a McNemar test over the discordant pairs, i.e., tasks solved by exactly one of the two methods. The bootstrap interval estimates the plausible range of the success-rate difference, while the McNemar test asks whether the number of method wins is reliably larger than the number of baseline wins.

The detailed paired analyses for inference-only and fine-tuning results are reported below.

\subsection{Statistical Reliability of Inference-Only ExpRAG}
\label{subapp:zero-shot-statistical-reliability}

Table~\ref{tab:zero-shot-statistical-reliability} reports paired task-level analyses for inference-only \ours against the strict zero-shot \textsc{No RAG} baseline on hard-task evaluations.

\begin{table}[p]
    \centering
    \small
    \setlength{\tabcolsep}{4pt}
    \begin{tabular}{lllccc}
    \toprule
    Backbone & Benchmark & Comparison & \(\Delta\) pp. & 95\% bootstrap CI & McNemar \(p\) \\
    \midrule
    \multicolumn{6}{l}{\textbf{Ministral 3-8B}} \\
    & ALFWorld-hard & \ours \(>\) \textsc{No RAG} & \(+47.5\) & \([+34.4,+60.7]\) & \(1.86{\times}10^{-9}\) \\
    & ScienceWorld-hard & \ours \(>\) \textsc{No RAG} & \(+15.6\) & \([+6.2,+25.0]\) & \(0.0032\) \\
    \midrule
    \multicolumn{6}{l}{\textbf{Gemma 3-4B}} \\
    & ALFWorld-hard & \ours \(>\) \textsc{No RAG} & \(+4.9\) & \([+0.0,+11.5]\) & \(0.125\) \\
    & ScienceWorld-hard & \ours \(>\) \textsc{No RAG} & \(+1.6\) & \([-4.7,+7.8]\) & \(0.5000\) \\
    \midrule
    \multicolumn{6}{l}{\textbf{Qwen 2.5-7B}} \\
    & ALFWorld-hard & \ours \(>\) \textsc{No RAG} & \(+45.9\) & \([+31.1,+60.7]\) & \(3.83{\times}10^{-7}\) \\
    & ScienceWorld-hard & \ours \(>\) \textsc{No RAG} & \(+4.7\) & \([-1.6,+12.5]\) & \(0.1875\) \\
    \midrule
    \multicolumn{6}{l}{\textbf{Qwen 2.5-7B-1M}} \\
    & ALFWorld-hard & \ours \(>\) \textsc{No RAG} & \(+59.0\) & \([+45.9,+72.1]\) & \(1.46{\times}10^{-11}\) \\
    & ScienceWorld-hard & \ours \(>\) \textsc{No RAG} & \(+12.5\) & \([+3.1,+23.4]\) & \(0.0193\) \\
    \bottomrule
    \end{tabular}%
    \caption{\textbf{Paired statistical reliability analysis for inference-only ExpRAG versus zero-shot.} Differences are in success-rate percentage points. Bootstrap intervals are paired task-level 95\% confidence intervals from 10{,}000 resamples; \(p\)-values are from McNemar tests over paired binary success/failure outcomes.}
    \label{tab:zero-shot-statistical-reliability}
\end{table}

\subsection{Statistical Reliability of Fine-Tuning Results}
\label{subapp:ft-statistical-results}

In Table~\ref{tab:ft_results}, significance markers are based on the paired McNemar test against the corresponding LoRA baseline: \(^{\dagger}p<0.1\), \(^{*}p<0.05\), \(^{**}p<0.01\), and \(^{***}p<0.001\). Bold entries in hard-task columns indicate improvements significant at \(p<0.05\). 

Table~\ref{tab:ft_statistical_reliability} reports the paired task-level analyses for the out-of-distribution hard-task comparisons in Table~\ref{tab:ft_results}. Overall, \oursraw-LoRA reliably improves over LoRA on ALFWorld-hard for all analyzed backbones and on ScienceWorld-hard for all analyzed backbones except Gemma~3-4B. The evidence is weaker for inference-only \ours over LoRA in several settings, and some gains over LoRA with inference-time \ours are not significant even when \oursraw-LoRA improves significantly over bare LoRA.

\begin{table}[p]
    \centering
    \small
    \setlength{\tabcolsep}{4pt}
    \resizebox{\linewidth}{!}{%
    \begin{tabular}{lllccc}
    \toprule
    Backbone & Benchmark & Comparison & \(\Delta\) pp. & 95\% bootstrap CI & McNemar \(p\) \\
    \midrule
    \multicolumn{6}{l}{\textbf{Ministral 3-8B}} \\
    & ALFWorld-hard & \oursraw-LoRA \(>\) LoRA & \(+54.1\) & \([+39.3,+68.9]\) & \(1.8{\times}10^{-8}\) \\
    & ALFWorld-hard & \oursraw-LoRA \(>\) LoRA (with \oursraw) & \(+21.3\) & \([+9.8,+34.4]\) & \(0.0012\) \\
    & ALFWorld-hard & LoRA (with \oursraw) \(>\) LoRA & \(+32.8\) & \([+18.0,+47.5]\) & \(9.0{\times}10^{-5}\) \\
    & ALFWorld-hard & \ours \(>\) LoRA & \(+13.1\) & \([-4.9,+31.1]\) & \(0.0977\) \\
    & ScienceWorld-hard & \oursraw-LoRA \(>\) LoRA & \(+26.6\) & \([+9.4,+42.2]\) & \(0.0023\) \\
    & ScienceWorld-hard & \oursraw-LoRA \(>\) LoRA (with \oursraw) & \(+14.1\) & \([+0.0,+28.1]\) & \(0.0466\) \\
    & ScienceWorld-hard & LoRA (with \oursraw) \(>\) LoRA & \(+12.5\) & \([+0.0,+26.6]\) & \(0.0577\) \\
    & ScienceWorld-hard & \ours \(>\) LoRA & \(+3.1\) & \([-9.4,+15.6]\) & \(0.4018\) \\
    \midrule
    \multicolumn{6}{l}{\textbf{Gemma 3-4B}} \\
    & ALFWorld-hard & \oursraw-LoRA \(>\) LoRA & \(+72.1\) & \([+60.7,+83.6]\) & \(5.68{\times}10^{-14}\) \\
    & ALFWorld-hard & \oursraw-LoRA \(>\) LoRA (with \oursraw) & \(+70.5\) & \([+59.0,+82.0]\) & \(1.14{\times}10^{-13}\) \\
    & ALFWorld-hard & LoRA (with \oursraw) \(>\) LoRA & \(+1.6\) & \([-3.3,+6.6]\) & \(0.5000\) \\
    & ALFWorld-hard & \ours \(>\) LoRA & \(+3.3\) & \([-3.3,+9.8]\) & \(0.3125\) \\
    & ScienceWorld-hard & \oursraw-LoRA \(>\) LoRA & \(+0.0\) & \([-7.8,+7.8]\) & \(0.6562\) \\
    & ScienceWorld-hard & \oursraw-LoRA \(>\) LoRA (with \oursraw) & \(-1.6\) & \([-9.4,+6.2]\) & \(0.7734\) \\
    & ScienceWorld-hard & LoRA (with \oursraw) \(>\) LoRA & \(+1.6\) & \([-4.7,+7.8]\) & \(0.5000\) \\
    & ScienceWorld-hard & \ours \(>\) LoRA & \(+0.0\) & \([-7.8,+7.8]\) & \(0.6562\) \\
    \midrule
    \multicolumn{6}{l}{\textbf{Qwen 2.5-7B}} \\
    & ALFWorld-hard & \oursraw-LoRA \(>\) LoRA & \(+68.9\) & \([+55.7,+80.3]\) & \(2.56{\times}10^{-12}\) \\
    & ALFWorld-hard & \oursraw-LoRA \(>\) LoRA (with \oursraw) & \(+19.7\) & \([+6.6,+32.8]\) & \(0.0059\) \\
    & ALFWorld-hard & LoRA (with \oursraw) \(>\) LoRA & \(+49.2\) & \([+32.8,+65.6]\) & \(1.41{\times}10^{-6}\) \\
    & ALFWorld-hard & \ours \(>\) LoRA & \(+49.2\) & \([+32.8,+65.6]\) & \(1.41{\times}10^{-6}\) \\
    & ScienceWorld-hard & \oursraw-LoRA \(>\) LoRA & \(+21.9\) & \([+7.8,+35.9]\) & \(0.0022\) \\
    & ScienceWorld-hard & \oursraw-LoRA \(>\) LoRA (with \oursraw) & \(+6.2\) & \([-7.8,+20.3]\) & \(0.2517\) \\
    & ScienceWorld-hard & LoRA (with \oursraw) \(>\) LoRA & \(+15.6\) & \([+3.1,+28.1]\) & \(0.0207\) \\
    & ScienceWorld-hard & \ours \(>\) LoRA & \(-1.6\) & \([-9.4,+6.2]\) & \(0.7734\) \\
    \midrule
    \multicolumn{6}{l}{\textbf{Qwen 2.5-7B-1M}} \\
    & ALFWorld-hard & \oursraw-LoRA \(>\) LoRA & \(+68.9\) & \([+54.1,+82.0]\) & \(6.56{\times}10^{-11}\) \\
    & ALFWorld-hard & \oursraw-LoRA \(>\) LoRA (with \oursraw) & \(+23.0\) & \([+13.1,+34.4]\) & \(6.10{\times}10^{-5}\) \\
    & ALFWorld-hard & LoRA (with \oursraw) \(>\) LoRA & \(+45.9\) & \([+27.9,+62.3]\) & \(7.55{\times}10^{-6}\) \\
    & ALFWorld-hard & \ours \(>\) LoRA & \(+39.3\) & \([+21.3,+55.7]\) & \(5.81{\times}10^{-5}\) \\
    & ScienceWorld-hard & \oursraw-LoRA \(>\) LoRA & \(+15.6\) & \([+1.6,+29.7]\) & \(0.0262\) \\
    & ScienceWorld-hard & \oursraw-LoRA \(>\) LoRA (with \oursraw) & \(+17.2\) & \([+4.7,+29.7]\) & \(0.0096\) \\
    & ScienceWorld-hard & LoRA (with \oursraw) \(>\) LoRA & \(-1.6\) & \([-12.5,+9.4]\) & \(0.7095\) \\
    & ScienceWorld-hard & \ours \(>\) LoRA & \(+1.6\) & \([-9.4,+12.5]\) & \(0.5000\) \\
    \bottomrule
    \end{tabular}%
    }
    \caption{\textbf{Paired statistical reliability analysis for hard-task fine-tuning results.} Differences are in success-rate percentage points. Bootstrap intervals are paired task-level \(95\%\) confidence intervals from \(10{,}000\) resamples; \(p\)-values are from McNemar tests over paired binary success/failure outcomes.}
    \label{tab:ft_statistical_reliability}
\end{table}

%% file: appendix/robustness.tex
\section{Robustness to Imperfect and Self-Generated Experience}
\label{app:imperfect-experience}

This appendix provides additional details for the experience-robustness
experiments in Sec.~\ref{sec:experiments:imperfect-experience}, including the construction of
each experience source, the Self-Experience evaluation protocol, and
the complete results for Ministral 3-8B.

Unless stated otherwise, these experiments reuse the same checkpoints
and inference configuration as Sec.~\ref{subsec:finetuning:results}, described in detail in Appendix~\ref{app:implementation-details:retrieval}. No additional fine-tuning is performed
for any robustness condition: only the experience available in the
retrieval index changes.

\subsection{Experience Index Construction}
\label{app:imperfect-experience:index}

Table~\ref{tab:imperfect-index-construction} defines the experience
sources used in the robustness experiments. Except for \textbf{Empty},
retrieval and inference are unchanged; conditions vary only in the
source, relevance, coverage, or origin of the available experience.

\begin{table}[t]
\centering
\small
\setlength{\tabcolsep}{2pt}
\begingroup
\renewcommand{\arraystretch}{1.08}
\begin{tabular}{
    >{\raggedright\arraybackslash}p{0.18\textwidth}
    >{\raggedright\arraybackslash}p{0.17\textwidth}
    >{\raggedright\arraybackslash}p{0.33\textwidth}
    >{\raggedright\arraybackslash}p{0.27\textwidth}}
\toprule
\textbf{Index condition} &
\textbf{Experience source} &
\textbf{Construction} &
\textbf{Purpose} \\
\midrule

\textbf{Matched Expert}
&
Scripted expert
&
Full expert index from the hard-task training split: 1,805 trajectories
for ALFWorld and 1,254 for ScienceWorld.
&
Reference condition with task-relevant expert experience.
\\

\textbf{Empty}
&
None
&
No trajectories are available for retrieval.
&
Measures performance when no external experience is available.
\\

\textbf{Mismatched Expert}
&
Scripted expert
&
Uses only trajectories from the easy-task training split while evaluating
on unseen hard tasks: 1,748 trajectories for ALFWorld and 2,335 for
ScienceWorld.
&
Tests robustness when experience is available but comes from different
task groups.
\\

\textbf{Low-Coverage Expert}
&
Scripted expert
&
Subsamples the matched hard-task index to five trajectories per hard task
type: 15 trajectories for ALFWorld and 65 for ScienceWorld.
&
Tests whether a large matched experience bank is necessary.
\\

\textbf{Matched Strong-LLM}
&
Qwen2.5-70B
&
Replaces scripted-expert trajectories with trajectories generated by
Qwen2.5-70B on matched hard tasks. These trajectories are generally more
suboptimal than those from the scripted expert.
&
Tests whether useful experience must come from a scripted expert or can
instead be seeded by a stronger model.
\\

\textbf{Self-Experience}
&
Evaluated agent
&
Uses the agent's own first-attempt trajectory on the hard evaluation
distribution, successful or failed, as experience for a second attempt.
Model parameters are unchanged between attempts.
&
Tests whether the agent can improve in-context from its own past
experience without test-time parameter updates.
\\

\textbf{Self-Experience + Source}
&
Agent + external source
&
Combines first-attempt experience with each non-empty external source:
Mismatched Expert, Low-Coverage Expert, Matched Expert, or Matched
Strong-LLM.
&
Tests whether self-generated experience remains useful when other
sources of experience are simultaneously available.
\\
\bottomrule
\end{tabular}
\endgroup
\caption{
Experience-index conditions used in the robustness experiments in
Sec.~\ref{sec:experiments:imperfect-experience}.
}
\label{tab:imperfect-index-construction}
\end{table}

\begin{table}[t]
\centering
\small
\setlength{\tabcolsep}{4pt}
\begin{tabular}{lrr}
\toprule
\textbf{Index condition} &
\textbf{Trajectories} &
\textbf{Success rate (\%)} \\
\midrule
Matched Expert                 & 1,805 & 62.2 (1,123/1,805) \\
Mismatched Expert\textsuperscript{a} & 1,748 & 97.5 (1,704/1,748) \\
Low-Coverage Expert           & 15    & 86.7 (13/15) \\
Matched Strong-LLM            & 1,799 & 68.6 (1,234/1,799) \\
\bottomrule
\end{tabular}
\caption{
Composition of the external ALFWorld experience indices. Success rate is reported as a percentage, followed by successful trajectories over total trajectories. \textbf{Notes:} \textsuperscript{a} Calculated from the \texttt{train/easy} split, although evaluation is on unseen hard tasks.
}
\label{tab:imperfect-index-stats-alfworld}
\end{table}

\begin{table}[t]
\centering
\small
\setlength{\tabcolsep}{3pt}
\begin{tabular}{lrrr}
\toprule
\textbf{Index condition} &
\textbf{Trajectories} &
\textbf{Success rate (\%)} &
\textbf{Score $=-1$ (\%)} \\
\midrule
Matched Expert                              & 1,254 & 99.0 (1,241/1,254) & 0.0 (0/1,254) \\
Mismatched Expert\textsuperscript{a}        & 2,335 & 93.7 (2,188/2,335) & 0.2 (5/2,335) \\
Low-Coverage Expert                         & 65    & 98.5 (64/65)        & 0.0 (0/65) \\
Matched Strong-LLM\textsuperscript{b}       & 165   & 73.3 (121/165)      & 2.4 (4/165) \\
Self-Experience\textsuperscript{c}          & 64    & 3.1 (2/64)           & 45.3 (29/64) \\
\bottomrule
\end{tabular}
\caption{ Composition of the ScienceWorld experience indices. Success rate and the proportion with score $=-1$ are reported as percentages, followed by count over total trajectories. \textbf{Notes:} \textsuperscript{a} Calculated from the \texttt{train/easy} split, although evaluation is on unseen hard tasks. \textsuperscript{b} The Qwen index is subsampled to 15 trajectories per hard task type (165 total). \textsuperscript{c} Calculated from zero-shot trajectories in \texttt{valid\_unseen/hard}.
}
\label{tab:imperfect-index-stats-scienceworld}
\end{table}

\subsection{Self-Experience Protocol}
\label{app:self-experience}

\textbf{Self-Experience} uses the zero-shot trajectory produced by the
agent's first attempt on each hard evaluation instance as memory for a
second attempt. The trajectory is stored regardless of whether the
first attempt succeeds or fails. We then perform the second attempt
with these zero-shot trajectories available through the same \ours
retrieval mechanism used elsewhere in the paper. Crucially, model
parameters remain fixed between the two attempts; adaptation occurs
exclusively through the retrieved context.

The first-attempt trajectories use exactly the same representation as
the other experience sources. Following Appendix~\ref{app:implementation-details:memory-format},
successful and unsuccessful trajectories are identified as such in the
memory block, allowing the policy to distinguish demonstrations of
successful behavior from failed experience.

On ALFWorld hard tasks, all first attempts fail. Thus, the
\textbf{Self-Experience}-only condition is especially strict: the second
attempt has access to self-generated experience, but none of that
experience contains a successful solution.

\subsection{Full Results}
\label{app:imperfect-experience:results}

Tables~\ref{tab:imperfect-results-alfworld},
\ref{tab:imperfect-results-scienceworld}, and
\ref{tab:imperfect-results-scienceworld-score} report results under each
experience condition. As in Fig.~\ref{fig:robustness_stress_tests_unified},
we evaluate three policies: inference-only \oursraw, LoRA with
\ours introduced only at inference time, and retrieval-aware
\oursraw-LoRA. All methods within a row receive experience from the
same source; thus, differences reflect how the policy uses the
available experience rather than differences in retrieval configuration.

\begin{table}[!h]
\centering
\scriptsize
\setlength{\tabcolsep}{3pt}
\resizebox{0.95\textwidth}{!}{
\begin{tabular}{llccc}
\toprule
\textbf{Method} &
\textbf{Experience source} &
\textbf{Mean (\%)} &
\textbf{Within-game 95\% CI} &
\textbf{Retention (\%)} \\
\midrule

\multirow{10}{*}{\oursraw}
& Matched Expert                         & 44.3 & [37.8, 50.8] & 100.0 [100.0, 100.0] \\
& Empty                                  & 0.0 & [-2.2, 2.2] & 0.0 [0.0, 0.0] \\
& Mismatched Expert                      & 8.7 & [4.6, 13.1] & 19.8 [10.5, 31.2] \\
& Low-Coverage Expert                    & 36.1 & [29.1, 43.2] & 81.5 [64.1, 102.7] \\
& Matched Strong-LLM                     & 44.3 & [37.4, 51.1] & 100.0 [82.7, 120.8] \\
& Self-Experience                        & 0.0 & [-2.2, 2.2] & 0.0 [0.0, 0.0] \\
& Self-Experience + Mismatched Expert    & 19.7 & [14.2, 25.3] & 44.4 [31.3, 59.7] \\
& Self-Experience + Low-Coverage Expert  & 23.0 & [16.9, 29.4] & 51.9 [38.2, 68.2] \\
& Self-Experience + Matched Expert       & 34.4 & [28.1, 40.9] & 77.8 [63.4, 94.0] \\
& Self-Experience + Matched Strong-LLM   & 55.7 & [48.7, 62.8] & 125.9 [106.0, 150.8] \\
\midrule
\multirow{10}{*}{LoRA + \oursraw}
& Matched Expert                         & 68.3 & [61.8, 74.5] & 100.0 [100.0, 100.0] \\
& Empty                                  & 31.1 & [25.3, 36.9] & 45.6 [35.2, 56.8] \\
& Mismatched Expert                      & 29.5 & [24.0, 35.1] & 43.2 [33.6, 53.7] \\
& Low-Coverage Expert                    & 50.8 & [44.6, 57.0] & 74.4 [63.4, 86.2] \\
& Matched Strong-LLM                     & 65.6 & [59.7, 71.3] & 96.0 [83.5, 109.9] \\
& Self-Experience                        & 31.1 & [25.2, 37.0] & 45.6 [34.9, 57.3] \\
& Self-Experience + Mismatched Expert    & 39.3 & [33.0, 45.5] & 57.6 [46.4, 69.7] \\
& Self-Experience + Low-Coverage Expert  & 57.4 & [50.8, 63.9] & 84.0 [71.5, 98.3] \\
& Self-Experience + Matched Expert       & 67.2 & [60.4, 73.9] & 98.4 [88.7, 108.9] \\
& Self-Experience + Matched Strong-LLM   & 72.1 & [66.5, 77.7] & 105.6 [93.2, 119.3] \\
\midrule
\multirow{10}{*}{\oursraw-LoRA}
& Matched Expert                         & 91.8 & [87.5, 95.9] & 100.0 [100.0, 100.0] \\
& Empty                                  & 30.6 & [24.3, 37.3] & 33.3 [25.9, 41.1] \\
& Mismatched Expert                      & 41.5 & [35.4, 47.5] & 45.2 [37.4, 53.0] \\
& Low-Coverage Expert                    & 83.6 & [77.8, 89.2] & 91.1 [84.1, 98.2] \\
& Matched Strong-LLM                     & 90.2 & [85.9, 94.3] & 98.2 [92.5, 104.3] \\
& Self-Experience                        & 60.7 & [53.7, 67.5] & 66.1 [57.6, 74.5] \\
& Self-Experience + Mismatched Expert    & 41.0 & [34.6, 47.3] & 44.6 [36.8, 52.8] \\
& Self-Experience + Low-Coverage Expert  & 70.5 & [64.7, 76.2] & 76.8 [69.5, 84.0] \\
& Self-Experience + Matched Expert       & 93.4 & [89.3, 97.2] & 101.8 [98.8, 105.0] \\
& Self-Experience + Matched Strong-LLM   & 90.2 & [85.9, 94.2] & 98.2 [93.1, 103.7] \\

\bottomrule
\end{tabular}
}
\caption{
Success rate (\%) on ALFWorld unseen hard tasks. Within-game CIs are
95\% normalized bootstrap CIs. Retention is relative to the Matched
Expert mean for the same method; brackets report its joint paired
bootstrap CI. All robustness conditions use the same trained
checkpoints; only the available experience changes.
}
\label{tab:imperfect-results-alfworld}
\end{table}

\begin{table}[!h]
\centering
\setlength{\tabcolsep}{3pt}
\resizebox{0.95\textwidth}{!}{
\begin{tabular}{llccc}
\toprule
\textbf{Method} &
\textbf{Experience source} &
\textbf{Mean (\%)} &
\textbf{Within-game 95\% CI} &
\textbf{Retention (\%)} \\
\midrule

\multirow{10}{*}{\oursraw}
& Matched Expert                         & 20.3 & [15.8, 25.1] & 100.0 [100.0, 100.0] \\
& Empty                                  & 4.7 & [1.4, 8.1] & 23.1 [9.4, 41.7] \\
& Mismatched Expert                      & 15.6 & [11.9, 19.6] & 76.9 [52.2, 111.8] \\
& Low-Coverage Expert                    & 22.9 & [18.4, 27.7] & 112.8 [84.8, 151.5] \\
& Matched Strong-LLM (subsampled)        & 26.6 & [21.9, 31.4] & 130.8 [97.7, 177.4] \\
& Self-Experience                        & 7.8 & [4.6, 11.1] & 38.5 [21.1, 60.7] \\
& Self-Experience + Mismatched Expert    & 7.8 & [4.6, 11.1] & 38.5 [21.1, 60.7] \\
& Self-Experience + Low-Coverage Expert  & 16.7 & [12.6, 21.1] & 82.1 [57.9, 114.3] \\
& Self-Experience + Matched Expert       & 17.2 & [12.8, 21.8] & 84.6 [63.6, 110.3] \\
& Self-Experience + Matched Strong-LLM   & 18.2 & [14.2, 22.5] & 89.7 [64.9, 123.3] \\
\midrule
\multirow{10}{*}{LoRA + \oursraw}
& Matched Expert                         & 31.2 & [25.2, 37.3] & 100.0 [100.0, 100.0] \\
& Empty                                  & 14.6 & [10.7, 18.7] & 46.7 [30.9, 66.7] \\
& Mismatched Expert                      & 12.5 & [8.6, 16.6] & 40.0 [26.3, 56.1] \\
& Low-Coverage Expert                    & 22.9 & [17.5, 28.4] & 73.3 [54.7, 95.8] \\
& Matched Strong-LLM (subsampled)        & 33.9 & [28.4, 39.2] & 108.3 [84.0, 139.6] \\
& Self-Experience                        & 6.2 & [3.4, 9.1] & 20.0 [9.8, 32.7] \\
& Self-Experience + Mismatched Expert    & 9.9 & [6.7, 13.2] & 31.7 [18.6, 48.1] \\
& Self-Experience + Low-Coverage Expert  & 12.5 & [8.2, 17.0] & 40.0 [25.0, 59.3] \\
& Self-Experience + Matched Expert       & 21.9 & [17.1, 26.8] & 70.0 [51.5, 93.3] \\
& Self-Experience + Matched Strong-LLM   & 18.8 & [14.6, 23.2] & 60.0 [42.9, 81.2] \\
\midrule
\multirow{10}{*}{\oursraw-LoRA}
& Matched Expert                         & 47.9 & [41.0, 54.8] & 100.0 [100.0, 100.0] \\
& Empty                                  & 3.1 & [0.1, 6.3] & 6.5 [2.1, 12.4] \\
& Mismatched Expert                      & 9.4 & [6.0, 13.0] & 19.6 [11.3, 29.6] \\
& Low-Coverage Expert                    & 54.7 & [47.9, 61.4] & 114.1 [96.2, 136.2] \\
& Matched Strong-LLM (subsampled)        & 38.0 & [32.2, 44.1] & 79.3 [65.3, 96.1] \\
& Self-Experience                        & 4.7 & [1.9, 7.5] & 9.8 [4.0, 17.3] \\
& Self-Experience + Mismatched Expert    & 7.3 & [4.6, 10.1] & 15.2 [8.0, 24.4] \\
& Self-Experience + Low-Coverage Expert  & 14.1 & [9.5, 19.0] & 29.3 [19.6, 40.9] \\
& Self-Experience + Matched Expert       & 29.7 & [23.5, 36.1] & 62.0 [50.0, 75.3] \\
& Self-Experience + Matched Strong-LLM   & 25.5 & [20.8, 30.6] & 53.3 [40.6, 68.2] \\

\bottomrule
\end{tabular}
}
\caption{
Success rate (\%) on ScienceWorld unseen hard tasks. Within-game CIs are
95\% normalized bootstrap CIs. Retention is relative to the Matched
Expert mean for the same method; brackets report its joint paired
bootstrap CI. The matched strong-LLM bank is reported as Qwen Teacher
(Subsampled) in the analysis output. Experimental settings otherwise
match Table~\ref{tab:imperfect-results-alfworld}.
}
\label{tab:imperfect-results-scienceworld}
\end{table}

\begin{table}[!h]
\centering
\scriptsize
\setlength{\tabcolsep}{3pt}
\resizebox{0.95\textwidth}{!}{
\begin{tabular}{llccc}
\toprule
\textbf{Method} &
\textbf{Experience source} &
\textbf{Mean score} &
\textbf{Within-game 95\% CI} &
\textbf{Retention (\%)} \\
\midrule

\multirow{10}{*}{\oursraw}
& Matched Expert                         & 0.399 & [0.352, 0.447] & 100.0 [100.0, 100.0] \\
& Empty                                  & 0.146 & [0.110, 0.183] & 36.5 [27.5, 47.2] \\
& Mismatched Expert                      & 0.288 & [0.253, 0.326] & 72.2 [58.5, 88.2] \\
& Low-Coverage Expert                    & 0.384 & [0.338, 0.432] & 96.1 [81.2, 113.5] \\
& Matched Strong-LLM (subsampled)        & 0.421 & [0.374, 0.466] & 105.3 [89.4, 123.8] \\
& Self-Experience                        & 0.204 & [0.169, 0.239] & 51.0 [40.2, 63.3] \\
& Self-Experience + Mismatched Expert    & 0.179 & [0.146, 0.214] & 44.9 [34.5, 57.1] \\
& Self-Experience + Low-Coverage Expert  & 0.295 & [0.252, 0.340] & 73.9 [60.7, 89.5] \\
& Self-Experience + Matched Expert       & 0.311 & [0.266, 0.358] & 77.9 [65.8, 90.7] \\
& Self-Experience + Matched Strong-LLM   & 0.331 & [0.289, 0.374] & 82.8 [70.1, 97.7] \\
\midrule
\multirow{10}{*}{LoRA + \oursraw}
& Matched Expert                         & 0.490 & [0.438, 0.542] & 100.0 [100.0, 100.0] \\
& Empty                                  & 0.228 & [0.187, 0.269] & 46.4 [36.9, 56.8] \\
& Mismatched Expert                      & 0.252 & [0.214, 0.291] & 51.4 [41.7, 61.9] \\
& Low-Coverage Expert                    & 0.407 & [0.357, 0.458] & 83.1 [72.1, 95.3] \\
& Matched Strong-LLM (subsampled)        & 0.520 & [0.475, 0.564] & 106.0 [92.8, 121.3] \\
& Self-Experience                        & 0.215 & [0.186, 0.243] & 43.8 [35.8, 52.3] \\
& Self-Experience + Mismatched Expert    & 0.242 & [0.211, 0.274] & 49.4 [40.2, 59.6] \\
& Self-Experience + Low-Coverage Expert  & 0.277 & [0.234, 0.321] & 56.4 [46.1, 67.8] \\
& Self-Experience + Matched Expert       & 0.377 & [0.332, 0.424] & 77.0 [66.4, 88.5] \\
& Self-Experience + Matched Strong-LLM   & 0.337 & [0.298, 0.377] & 68.8 [57.8, 81.0] \\
\midrule
\multirow{10}{*}{\oursraw-LoRA}
& Matched Expert                         & 0.662 & [0.609, 0.713] & 100.0 [100.0, 100.0] \\
& Empty                                  & 0.112 & [0.075, 0.150] & 16.9 [12.4, 21.8] \\
& Mismatched Expert                      & 0.197 & [0.158, 0.237] & 29.7 [23.2, 37.1] \\
& Low-Coverage Expert                    & 0.679 & [0.628, 0.731] & 102.6 [93.4, 112.8] \\
& Matched Strong-LLM (subsampled)        & 0.532 & [0.480, 0.586] & 80.4 [71.8, 89.7] \\
& Self-Experience                        & 0.181 & [0.147, 0.215] & 27.3 [21.4, 34.2] \\
& Self-Experience + Mismatched Expert    & 0.195 & [0.162, 0.228] & 29.4 [22.9, 36.7] \\
& Self-Experience + Low-Coverage Expert  & 0.285 & [0.239, 0.334] & 43.1 [35.7, 51.4] \\
& Self-Experience + Matched Expert       & 0.442 & [0.387, 0.498] & 66.8 [58.2, 75.8] \\
& Self-Experience + Matched Strong-LLM   & 0.392 & [0.346, 0.440] & 59.2 [50.7, 68.7] \\

\bottomrule
\end{tabular}
}
\caption{
Transformed score (0--1) on ScienceWorld unseen hard tasks. Before
averaging, we replace the failure score \(-1\) with \(0\), so failures
contribute zero rather than a negative value and do not create an
asymmetric penalty relative to the nonnegative task scores. Within-game
CIs are 95\% normalized bootstrap CIs. Retention is relative to the
Matched Expert mean for the same method; brackets report its joint
paired bootstrap CI. Experimental settings otherwise match
Table~\ref{tab:imperfect-results-scienceworld}.
}
\label{tab:imperfect-results-scienceworld-score}
\end{table}

\clearpage

\subsection{Index Stress Test in other backbones}
\label{app:imperfect-experience:cross-backbone}

To complement the complete Ministral 3-8B stress test above,
Table~\ref{tab:cross-backbone-index-stress} reports initial
cross-backbone experiment that evaluates only two conditions:
\textbf{Empty} and \textbf{Mismatched Expert}. Each cell gives the success rate with the matched hard-task expert index followed by the success rate under the stressed index. This experiment uses the original fine-tuning evaluation setup; its absolute values should therefore be interpreted independently of the updated Ministral results above.

\input{tables/robustness_cross_backbone}

\subsection{Statistical Reliability}
\label{app:imperfect-experience:statistics}

Figure~\ref{fig:robustness_stress_tests_unified} and
Tables~\ref{tab:imperfect-results-alfworld},
\ref{tab:imperfect-results-scienceworld}, and
\ref{tab:imperfect-results-scienceworld-score}
report 95\% within-game normalized bootstrap CIs, using \texttt{game\_id +
seed} as the resampling unit. This normalization removes shared game
difficulty and is intended for visual comparisons among the aligned
conditions, rather than as standalone absolute uncertainty. The paired
task-level testing procedure used for the main fine-tuning comparisons
is described separately in Appendix~\ref{app:statistical-reliability}.

%% file: tables/robustness_cross_backbone.tex
\begin{table}[t]
\centering
\small
\setlength{\tabcolsep}{3pt}
\renewcommand{\arraystretch}{0.94}
\begin{tabular}{llcc}
\toprule
\multirow{3}{*}{\shortstack[c]{\textbf{Index}\\\textbf{condition}}} &
\multirow{3}{*}{\textbf{Method}} &
\multicolumn{1}{c}{\textbf{ALFWorld}} &
\multicolumn{1}{c}{\textbf{ScienceWorld}} \\
& & \textbf{Hard tasks} & \textbf{Hard tasks} \\
& & \textbf{(oo-d)} & \textbf{(oo-d)} \\
\midrule
\multirow{3}{*}{\shortstack[l]{\textbf{Empty}\\\textit{Gemma 3-4B}}}
& \ours & $4.9 \rightarrow 0.0$ & $6.3 \rightarrow 4.7$ \\
& LoRA + \oursraw & $3.3 \rightarrow 1.6$ & $7.8 \rightarrow 6.3$ \\
& \oursraw-LoRA & $73.8 \rightarrow 4.9$ & $4.7 \rightarrow 0.0$ \\
\cmidrule(l){2-4}
\multirow{3}{*}{\shortstack[l]{\textbf{Mismatched Expert}\\\textit{Gemma 3-4B}}}
& \ours & $4.9 \rightarrow 4.9$ & $6.3 \rightarrow 1.6$ \\
& LoRA + \oursraw & $3.3 \rightarrow 4.9$ & $7.8 \rightarrow 7.8$ \\
& \oursraw-LoRA & $73.8 \rightarrow 9.8$ & $4.7 \rightarrow 3.1$ \\
\midrule
\multirow{3}{*}{\shortstack[l]{\textbf{Empty}\\\textit{Qwen 2.5-7B}}}
& \ours & $68.9 \rightarrow 22.9$ & $6.3 \rightarrow 3.1$ \\
& LoRA + \oursraw & $70.5 \rightarrow 21.3$ & $23.4 \rightarrow 7.8$ \\
& \oursraw-LoRA & $90.2 \rightarrow 34.4$ & $29.7 \rightarrow 4.7$ \\
\cmidrule(l){2-4}
\multirow{3}{*}{\shortstack[l]{\textbf{Mismatched Expert}\\\textit{Qwen 2.5-7B}}}
& \ours & $68.9 \rightarrow 54.1$ & $6.3 \rightarrow 0.0$ \\
& LoRA + \oursraw & $70.5 \rightarrow 24.6$ & $23.4 \rightarrow 17.2$ \\
& \oursraw-LoRA & $90.2 \rightarrow 50.8$ & $29.7 \rightarrow 17.2$ \\
\midrule
\multirow{3}{*}{\shortstack[l]{\textbf{Empty}\\\textit{Qwen 2.5-7B-1M}}}
& \ours & $54.1 \rightarrow 3.3$ & $7.8 \rightarrow 3.1$ \\
& LoRA + \oursraw & $68.9 \rightarrow 23.0$ & $12.5 \rightarrow 12.5$ \\
& \oursraw-LoRA & $91.8 \rightarrow 36.1$ & $29.7 \rightarrow 4.7$ \\
\cmidrule(l){2-4}
\multirow{3}{*}{\shortstack[l]{\textbf{Mismatched Expert}\\\textit{Qwen 2.5-7B-1M}}}
& \ours & $54.1 \rightarrow 29.5$ & $7.8 \rightarrow 0.0$ \\
& LoRA + \oursraw & $68.9 \rightarrow 21.3$ & $12.5 \rightarrow 6.3$ \\
& \oursraw-LoRA & $91.8 \rightarrow 60.7$ & $29.7 \rightarrow 3.1$ \\
\midrule
\multirow{3}{*}{\shortstack[l]{\textbf{Empty}\\\textit{Llama 3.1-8B}}}
& \ours & $34.4 \rightarrow 3.3$ & $20.3 \rightarrow 9.4$ \\
& LoRA + \oursraw & $44.3 \rightarrow 21.3$ & $21.9 \rightarrow 7.8$ \\
& \oursraw-LoRA & $80.3 \rightarrow 14.8$ & $21.9 \rightarrow 3.1$ \\
\cmidrule(l){2-4}
\multirow{3}{*}{\shortstack[l]{\textbf{Mismatched Expert}\\\textit{Llama 3.1-8B}}}
& \ours & $34.4 \rightarrow 1.6$ & $20.3 \rightarrow 4.7$ \\
& LoRA + \oursraw & $44.3 \rightarrow 21.3$ & $21.9 \rightarrow 4.7$ \\
& \oursraw-LoRA & $80.3 \rightarrow 34.4$ & $21.9 \rightarrow 1.6$ \\
\midrule
\multirow{3}{*}{\shortstack[l]{\textbf{Empty}\\\textit{Llama 3.2-1B}}}
& \ours & $0.0 \rightarrow 0.0$ & $0.0 \rightarrow 0.0$ \\
& LoRA + \oursraw & $8.2 \rightarrow 0.0$ & $4.7 \rightarrow 4.7$ \\
& \oursraw-LoRA & $34.4 \rightarrow 1.6$ & $10.9 \rightarrow 1.6$ \\
\cmidrule(l){2-4}
\multirow{3}{*}{\shortstack[l]{\textbf{Mismatched Expert}\\\textit{Llama 3.2-1B}}}
& \ours & $0.0 \rightarrow 0.0$ & $0.0 \rightarrow 0.0$ \\
& LoRA + \oursraw & $4.3 \rightarrow 21.3$ & $4.7 \rightarrow 0.0$ \\
& \oursraw-LoRA & $80.3 \rightarrow 34.4$ & $10.9 \rightarrow 4.7$ \\
\bottomrule
\end{tabular}
\caption{\textbf{Index stress test for other backbones.} Each cell reports the success rate (\%) as $a \rightarrow b$, where $a$ uses the matched hard-task expert index and $b$ uses either \textbf{Empty} (no trajectories) or \textbf{Mismatched Expert} (easy-task training trajectories).}
\label{tab:cross-backbone-index-stress}
\end{table}

%% file: appendix/prompts.tex
\section{Prompts}
\label{app:prompts}

This appendix details the exact prompting interface used in our experiments. For each environment, we use a \emph{minimal} system prompt that specifies: (i) the task setting, (ii) the action interface/grammar, and (iii) the response format (one action per turn). 

\paragraph{Restricting turn-level action-space hints.} We provide a static list of action \emph{templates} valid for the whole environment (Figures~\ref{fig:alfworld-system-prompt} and~\ref{fig:scienceworld-system-prompt}), rather than per-step instantiated valid-action candidates, even though the environment can provide those candidates. This design removes turn-level action-space hints, requiring the agent to infer plausible next actions from past observations and actions alone. Importantly, this diverges from contemporaneous work that exposes these turn-level candidate actions during rollout, arguably introducing some ground-truth information leakage. In early explorations in our zero-shot setting, we found this stricter setup to reduce backbone performance by approximately 15\%.

Figure~\ref{fig:alfworld-system-prompt} shows the full ALFWorld system prompt. It includes the complete command inventory and formatting constraints (including one-command output), plus guidance for handling invalid actions (e.g., ``Nothing happens''). Figure~\ref{fig:scienceworld-system-prompt} shows the analogous ScienceWorld prompt, with its environment-specific action grammar and explicit command-format constraints (e.g., \texttt{pick up <OBJ>} and no extra prefix text).

Finally, Figure~\ref{json-example} illustrates how interactions are serialized as chat data. Consistent with the method in Section~\ref{sec:method}, each trajectory is encoded as a \emph{multi-turn chat}: environment observations/task context are mapped to user turns, and action strings are mapped to assistant turns. This contrasts with stepwise formatting used by contemporaneous work, where each \((h_t, a_t)\) pair is treated as an independent sample with history re-encoding at every step. We use multi-turn serialization to preserve conversational structure and training/inference consistency with chat models while keeping supervision on assistant action tokens.

As a final note, when using retrieval, we append the retrieved trajectories to the system prompt as a memory block, as described in Section~\ref{app:implementation-details:memory-format}.

\onecolumn
\begin{figure}[!h]
\centering
\begin{tcolorbox}[title=Example of (partial) JSON trajectory (ALFWorld), colback=gray!5, colframe=black, boxrule=0.6pt,
  left=0pt, right=0pt, top=6pt, bottom=6pt,width=\textwidth]
\ttfamily\tiny
\begin{lstlisting}
{ 
    "role": "user",
    "content": "-= Welcome to TextWorld, ALFRED! =-\n\nYou are in the middle of a room. 
                Looking quickly around you, you see a bathtubbasin 1, a countertop 2, 
                a countertop 1, a drawer 4, a drawer 3, a drawer 2, a drawer 1, a garbagecan 1,
                a handtowelholder 1, a sinkbasin 1, a toilet 1, a toiletpaperhanger 1,
                a towelholder 2,  and a towelholder 1.\n\nYour task is to: put a candle in drawer."
},
{
    "role": "assistant",
    "content": "look"
},
{
    "role": "user",
    "content": "You are in the middle of a room. Looking quickly around you, you see nothing."
},
{
    "role": "assistant",
    "content": "go to toilet 1"
},
{
    "role": "user",
    "content": "You arrive at toilet 1. On the toilet 1, you see a spraybottle 3, a spraybottle 2, 
               and a spraybottle 1."
},
{
    "role": "assistant",
    "content": "go to drawer 1"
}
....
\end{lstlisting}
\end{tcolorbox}
\caption{Example of a partial JSON trajectory in ALFWorld.}
\label{json-example}
\end{figure}

\begin{figure}[!h]
\centering
\begin{tcolorbox}[title=System prompt for ALFWorld,breakable, colback=gray!5, colframe=black, boxrule=0.6pt,
  left=6pt, right=6pt, top=6pt, bottom=6pt]
\ttfamily\small
Interact with a household to solve a task. Here is the list of commands:
\\[0.4\baselineskip]
Available commands:\\
look:                             look around your current location\\
inventory:                        check your current inventory\\
go to (receptacle):               move to a receptacle\\
open (receptacle):                open a receptacle\\
close (receptacle):               close a receptacle\\
take (object) from (receptacle):  take an object from a receptacle\\
move (object) to (receptacle):    place an object in or on a receptacle\\
examine (something):              examine a receptacle or an object\\
use (object):                     use an object\\
heat (object) with (receptacle):  heat an object using a receptacle\\
clean (object) with (receptacle): clean an object using a receptacle\\
cool (object) with (receptacle):  cool an object using a receptacle\\
slice (object) with (object):     slice an object using a sharp object
\\[0.6\baselineskip]
You can only perform these commands. The text command you generate must be strictly compliant with the command grammar.\\
Only provide one single command in your response. This is an example command: go to shelf 1\\
Each time you provide a command, the household will return the outcome of it, and you will have to provide another command.\\
Note that you cannot carry two items at the same time so when you take an item, you must remember it and possibly move it to a receptacle later.
\end{tcolorbox}
\caption{System prompt used for ALFWorld experiments.}
\label{fig:alfworld-system-prompt}
\end{figure}

\begin{figure}[!h]
\centering
\begin{tcolorbox}[title=System prompt for ScienceWorld, colback=gray!5, colframe=black, boxrule=0.6pt,
  left=6pt, right=6pt, top=6pt, bottom=6pt]
\ttfamily\small
Interact with the environment to solve a scientific experiment.\\
In the environment, there are several rooms: kitchen, foundry, workshop, bathroom, outside, living room, bedroom, greenhouse, art studio, hallway.
\\[0.6\baselineskip]
All containers in the environment have already been opened, you can directly get items from the containers.
\\[0.6\baselineskip]
The available actions are:
\\[0.4\baselineskip]
open OBJ: open a container\\
close OBJ: close a container\\
activate OBJ: activate a device\\
deactivate OBJ: deactivate a device\\
connect OBJ to OBJ: connect electrical components\\
disconnect OBJ: disconnect electrical components\\
use OBJ [on OBJ]: use a device/item\\
look around: describe the current room\\
examine OBJ: describe an object in detail\\
look at OBJ: describe a container's contents\\
read OBJ: read a note or book\\
move OBJ to OBJ: move an object to a container\\
pick up OBJ: move an object to the inventory\\
pour OBJ into OBJ: pour a liquid into a container\\
mix OBJ: chemically mix a container\\
teleport to LOC: teleport to a specific room\\
focus on OBJ: signal intent on a task object\\
wait: task no action for 10 steps\\
wait1: task no action for a step
\\[0.6\baselineskip]
For the pick up action: the command is `pick up <OBJ>`.\\
Do not prefix the command with text.\\
You can only perform these commands. The text command you generate must be strictly compliant with the command grammar. Only provide one single command in your response.
\end{tcolorbox}
\caption{System prompt used for ScienceWorld experiments.}
\label{fig:scienceworld-system-prompt}
\end{figure}

%% file: appendix/efficiency.tex
\section{Efficiency Cost of \oursraw}
\label{app:efficiency}

Table~\ref{tab:efficiency} reports how the value of top-$K$ affects execution time and context size for \oursraw. We note that the average number of prompt tokens scales almost linearly with $k$. The average number of steps required to perform the task is decreasing with higher top-$k$, meaning that the agent is able to solve the tasks faster. Total processing time is therefore growing much slower. We stick with top-$2$ as the best efficiency-effectiveness tradeoff for the follow-up experiments. 

\begin{table}[!htbp]
    \centering
    \begin{tabular}{c|cccc}
    \toprule
    & prompt tokens  & steps/sample& total time (sec) & time/step \\
    \midrule
    No RAG & 489.6 & 48 & 2769.59 & 0.430194 \\
    \oursraw~top1 & 1851.9 & 34.4 & 2719.94 & 0.590649 \\
    \oursraw~top2 & 3096.7 & 29.8 & 2927.81 & 0.734339 \\
    \oursraw~top4 & 5597.7 & 25.7 & 3556.04 & 1.03403 \\
    \oursraw~top5 & 6866.4 & 25 & 3657.74 & 1.09284 \\
    \bottomrule
    \end{tabular}
    \caption{Model: Ministral 3-8B, dataset: ALFWorld. Efficiency of different \ours variants compared to zero-shot: average number of prompt tokens processed at each step, average number of steps per episode, total generation time for 134 episodes, and average time per step.}
    \label{tab:efficiency}
\end{table}